\documentclass[10pt]{article}
\usepackage[utf8]{inputenc}
\usepackage[a4paper, left=2cm, right=2cm, top=2cm]{geometry}
\usepackage{amsmath,amssymb,amsthm,mathrsfs,nicefrac}
\usepackage{mathtools}
\usepackage{latexsym,amscd,amsbsy,dsfont,amsfonts}
\usepackage{graphicx,wrapfig}
\usepackage{authblk,textcomp}
\usepackage[dvipsnames]{xcolor}
\usepackage{cancel}
\usepackage{bm}
\usepackage{bbm}
\usepackage[cal=euler]{mathalfa}
\usepackage{libertine}
\usepackage[libertine,smallerops]{newtxmath}
\usepackage[T1]{fontenc}
\usepackage[font=small]{caption}
\usepackage{tikz}
\usetikzlibrary{arrows}
\usetikzlibrary{decorations}
\usepackage{subfigure}
\usepackage{booktabs}
\usepackage{doi}
\usepackage{float}
\usepackage{enumitem}
\usepackage{physics}
\usepackage{hyperref}
\hypersetup{pdfauthor={EPFL-SMILS},pdftitle={sgf_dynamics},%
            colorlinks, linktocpage=true, pdfstartpage=1, pdfstartview=FitV,%
    breaklinks=true, pdfpagemode=UseNone, pageanchor=true, pdfpagemode=UseOutlines,%
    plainpages=false, bookmarksnumbered, bookmarksopen=true, bookmarksopenlevel=1,%
    hypertexnames=true, pdfhighlight=/O,%
    urlcolor=teal, linkcolor=Blue, citecolor=NavyBlue}

\def\w{\bm{w}}

\def\x{\bm{x}}
\def\X{\bm{X}}
\def\Prob{\mathbb{P}}
\def\E{\mathbb{E}}

\def\time{\tau}

\def\svscalar{w}
\def\sv{\bm{\svscalar}}
\def\svbrow{\bm{\eta}}
\def\svI{\sv_{0}}

\def\svode{\sv^{\text{ode}}}

\def\Var{\text{Var}}

\def\tI{t_{0}}

\def\ProbFI{\Prob_{\lr}(\sv,t|\svI,\tI)}
\def\Prop{\Prob_{\Delta \time} (\sv_{k+1}, \time_{k+1}|\sv_k, \time_k)}

\def\O{\mathcal{O}}

\def\Jac{\pmb{\mathscr{J}}}
\def\Jacf{\Jac_{\! \! \f}}
\def\U{\bm{U}}

\def\Hess{\bm{{\cal H}}}
\def\HessL{\Hess_{\lossL}}

\def\dsvdtD{\frac{\Delta\sv_k}{\Delta \time}}

\def\Om{\bm{\Omega}}

\def\Omke{\Om_{k}^{\eps}}
\def\OmkeC{\Om^{\eps} (\time, \sv(\time))}

\def\Omeps{\Om^{\eps}}

\def\Dsvt{\mathcal{D}\left[\sv (\cdot) \right]}

\def\Dzt{\bar{\mathcal{D}}\left[ \z (\cdot) \right]}

\def\Lagr{\mathcal{L}}

\def\LagrfE{\Lagr^{\eps}\left(\time,\sv(\time), \dot{\sv}(\time)\right)}
\def\LagrfEode{\Lagr^{\eps}\left(\time,\svode(\time), \dot{\sv}^{\text{ode}}(\time)\right)}

\def\PsiB{\bm{\Psi}}

\def\C{\pmb{\mathscr{C}}}

\def\S{\mathcal{S}}

\def\Svte{\S^{\eps}\left[\sv (\cdot) \right]}

\def\Svteode{\S^{\eps}\left[\svode (\cdot) \right]}

\def\ID{\mathbb{I}}
\def\T{\mathcal{T}}
\def\N{\mathcal{N}}
\def\Ind{\mathbbm{1}}

\def\etab{\bm{\eta}}

\def\z{\pmb{z}}

\def\zero{\bm{0}}

\def\y{\bm{y}}
\def\R{\mathbb{R}}
\def\NN{\mathbb{N}}
\def\bb{\bm{\beta}}
\def\bbh{\hat{\bb}}
\def\bbt{\tilde{\bb}}
\def\beps{\bm{\epsilon}}

\def\f{\bm{f}}
\def\G{\bm{G}}
\def\brow{\bm{\mathscr{b}}}

\def\sub{\mathcal{A}}
\def\subc{{\sub^c}}

\def\Gk{\G_{k}}
\def\GkT{\G_{k}^\top}
\def\fk{\f_{k}}
\def\eps{\epsilon}

\def\grad{\bm{\nabla}}
\def\lr{\gamma}

\def\loss{l}
\def\lossL{\mathscr{L}}

\def\i{\nu}

\newcommand{\rnum}[1]{\uppercase\expandafter{\romannumeral #1\relax}}

\def\IDn{\ID_{n}}
\def\betas{\bbh_\sub}
\def\betasc{\bbh_{\subc}}
\def\betasode{\bbh_{\sub}^{\text{GF}}}
\def\betassde{\bbh_{\sub}^{\text{SGF}}}
\def\bbhz{\bbh^0}
\def\bbhzs{\bbhz_\sub}

\def\Etest{\mathcal{E}_{\mathrm{test}}}
\def\Etestode{\Etest^{\text{GF}}(t)}
\def\Etestsde{\Etest^{\text{SGF}}(t)}

\def\Etrain{\mathcal{E}_{\mathrm{train}}}
\def\Etrainode{\Etrain^{\text{GF}}(t)}

\def\Usvd{\bm{U}} 
\def\usvd{\bm{u}}
\def\Lsvd{\bm{\Lambda}} 
\def\Vsvd{\bm{V}}
\def\vsvd{\bm{v}} 
\def\Lsvdr{\Lsvd_r}  
\def\rorth{{r^{\perp}}}

\def\e{\text{{\bf e}}}
\def\xib{\bm{\xi}}
\def\EUn{\E_{k_\i \sim {\cal U} \{1 , n \}} }
\def\EUnq{\E_{q_\i \sim {\cal U} \{1 , n \}} }
\def\xibAll{\xib_{k_\i} \bigl(\bbh ;\X,\y; \sub \bigl)}
\def\bigO{{\cal O}}

\def\betat{\bbt_\sub}
\def\bbtcoord{\tilde{\beta}}

\def\xnew{\x^{\text{new}}}
\def\ynew{y^{\text{new}}}
\def\epsnew{\epsilon^{\text{new}}}

\def\avec{\bm{a}}
\def\Unifd{\text{Unif}}
\newcommand{\Unif}[1]{\Unifd \left( #1 \right) }
\newcommand{\sphere}[2]{\mathcal{S}^{#1 - 1} \left( #2 \right)}
\newcommand{\orthg}[1]{O \left( #1 \right)}

\def\Wishnp{\nu_{(n \times p)}}
\def\Wishpn{\nu_{(p \times n)}}
\def\MPalphaorig{\nu_{\alpha}}
\def\MPalphaoriginv{\nu_{\nicefrac{1}{\alpha}}}
\def\MPalpha{\rho_{\alpha}}
\def\MPalphainv{\rho_{\nicefrac{1}{\alpha}}}
\def\MPconverge{\xrightarrow{n, p \to \infty,\; p / n \to \alpha}}
\def\MPconvergeshort{\underset{p / n \to \alpha\;}{\xrightarrow{n, p \to \infty}}  }
\def\limasymp{ \lim_{\substack{n, d, p \to \infty \\ d /n \to \psi,\: p / n \to \alpha}} }
\def\aplus{{\alpha_+}}
\def\aminus{{\alpha_-}}

\def\D{\bm{D}}
\DeclareMathOperator{\diag}{diag}

\usepackage[toc,page]{appendix} 
\appto\appendix{\counterwithin{equation}{section}}

\title{Stochastic Gradient Flow Dynamics of Test Risk and its Exact Solution for Weak Features \vspace*{0.8em}}

\author{Rodrigo Veiga}
\author{Anastasia Remizova}
\author{Nicolas Macris}
\affil{\small Ecole Polytechnique Fédérale de Lausanne (EPFL).
Lab for Statistical Mechanics of Inference in Large Systems (SMILS). \newline CH-1015 Lausanne, Switzerland.}

\date{}

\begin{document}

\maketitle

\vspace{-2.5ex}
\begin{abstract}
We investigate the test risk of a continuous-time stochastic gradient flow dynamics in learning theory. Using a path integral formulation we provide, in the regime of small learning rate, a general formula for computing the difference between test risk curves of pure gradient and stochastic gradient flows. We apply the general theory to a simple model of weak features, which displays the double descent phenomenon, and explicitly compute the corrections brought about by the added stochastic term in the dynamics, as a function of time and model parameters. The analytical results are compared to simulations of discrete-time stochastic gradient descent and show good agreement.

\end{abstract}

%%%%%%%%%%%%%%%%%%%%%%%%%%%%%%
%%%%%%%%%%%%%%%%%%%
\section{Introduction}\label{sec:intro}
%%%%%%%%%%%%%%%%%%%
In supervised learning of neural networks and regression models, understanding the dynamics of optimization algorithms, and in particular stochastic gradient descent (SGD), is of utmost importance. However, despite much progress in a number of directions, this still remains a highly challenging theoretical problem. A fruitful approach that allows making analytical progress consists of suitably approximating SGD by a continuous time approximation, henceforth called stochastic gradient flow (SGF). In this contribution, we build up on this approach, to develop a general formalism characterizing the dynamics of the stochastic process, and apply it to the investigation of the test risk (or generalization error) as a function of time. 

As is well known, the classical bias-variance trade-off has been challenged in a number of models displaying the double descent phenomenon~\cite{belkin_2019, belkin_2020, belkin_2021}. Analytical derivations of double descent curves have been achieved for relatively simple models, but are limited to the use of least squares estimators (no dynamics) and pure gradient flow (GF) approximations of gradient descent (GD). The present work goes one step further by investigating the effects of stochasticity on the double descent curve. 

Our {\bf main contributions} are summarized as follows:

\begin{description}[wide = 2pt]
    \item[C1]  We consider a general Itô stochastic differential equation (SDE) and represent the Markovian transition probability as a path integral, Eq.~\eqref{eq:0funct_integral}. A general `explicit' formula for the transition probability, Eq.~\eqref{gaussian-final},  is derived in the limit of a small learning rate by using a Laplace approximation. This constitutes one of the main results of this paper: suppose that $\svode(t)$ is the deterministic solution of pure GF (with no stochastic term), then by representing the large deviation of the Itô diffusion of SGF as $\svode(t) + \sqrt{\lr} \z(t)$ we provide a general formula for the covariance $\C(t) \equiv \E[\z(t)\z(t)^\top]$. The formula is confronted against a simple rigorously solvable SDE and we find exact agreement.
    \item[C2] We use the simplest relevant SGF approximation of SGD in a learning theory setting, which is known to come with a first-order guarantee with respect to small learning rates~\cite{li_2019}. We use our general theory to express the covariance $\C(t)$ in terms of the data matrix, training cost and its Hessian, and the deterministic GF trajectory.
    \item[C3]  This general theoretical framework is applied to a particular random weak features model, displaying the double descent phenomenon, first discussed by~\cite{breiman_1983} and revisited in~\cite{belkin_2020}. We explicitly compute the corrections brought about by the added stochastic term in the {\it whole} learning dynamics. For i.i.d. Gaussian data, and in a suitable large-size asymptotic regime, the formula for the test risk of SGF is given explicitly by an expression involving only single and double integrals over the Marchenko-Pastur distribution, Eqs.~\eqref{assympt-1} and~\eqref{assympt-2}. We check our analytical predictions against numerical simulations of SGD and find good agreement.     
\end{description}

%%%%%%%%%%%%%%%%%%%
\paragraph{Relation to previous work --} 
Approximating discrete time stochastic processes by continuous time dynamics has been studied for more than three decades (see~\cite{kushner_2003} for a comprehensive monograph). In particular, many works have attempted to approximate SGD by considering continuous time Ornstein-Uhlenbeck processes~\cite{mandt_2016,mandt_2017,jastrzkbski_2018, fan_2018, wang_2020} assuming constant covariance in the continuous time limit; however this is not necessarily a good approximation~\cite{ali_2020}. In~\cite{mandt_2015,li_2015} the authors heuristically proposed `better' SDEs with a non-trivial covariance. This framework is a stochastic version of the method of modified equations in the analysis of classical finite difference methods~\cite{warming_1974}. Further developments are found in~\cite{li_2017, li_2019} where rigorous approximation results, in the sense of convergence in probability, are proved. 

The case of regularized least squares in high-dimensions is discussed in~\cite{paquette_2022homogenization,paquette_2022} where the dynamics of the so-called \emph{homogenized} SGD~\cite{paquette_2021} is characterized with rigorous guarantees in terms of Volterra integral equations. We expect the solution of these equations to agree with ours when applied to the weak features model in the limit of a small learning rate. Nevertheless, the saddle point calculation of the path integral employed here circumvents the need to solve Volterra integral equations, which is itself a non-trivial task. 

Other interesting formulations of SGF were also analyzed in~\cite{mignacco_2020,mignacco_2021} through Dynamical Mean-Field Theory (DMFT) in the contexts of high-dimensional Gaussian mixture classification and phase retrieval. Rigorous advances along this line of work were recently obtained~\cite{gerbelot_2024}. Furthermore, the role of stochasticity in escaping flat directions near initialization is discussed in~\cite{arnaboldi_2023} in the context of two-layer neural networks. 

The path integral formulation used here remains at a heuristic level. We point out that for more restricted SDEs (with constant covariance and small noise) there exist rigorous formulations belonging to the Freidlin-Wentzell large deviation theory (see~\cite{freidlin2012random} and~\cite{DBLP:books/sp/DemboZ98}, Chapter 5.6). Extending the Freidlin-Wentzell large deviation theory with state-dependent covariances is, to the best of our knowledge, an open mathematical problem.

The complete time dependence of training and generalization errors of GF has been calculated for the random features and Gaussian covariate models in~\cite{bodin_2021, bodin_2022} using advanced random matrix machinery. However, the corresponding curves for SGF have not been investigated within these models. Here, as our main focus is on SGF dynamics, we limit ourselves to the simplest possible weak features model which requires minimal random matrix theory, and postpone the harder calculation for other models to later work. 

An interesting study of SGF, and notably its implicit bias, has been initiated for diagonal networks in a recent series of papers~\cite{pesme_2021, pesme_2023}, where the benefit of the stochasticity for large training times has been uncovered, provided that the weights are properly initialized. Additionally, we mention the \emph{decoupling approximation} near local minima proposed in~\cite{mori_2022}, making the noise contribution additive near these minima and considerably simplifying the stochastic dynamics. These perspectives could also constitute an interesting test lab for the present approach, with which one could analyze the effects of stochasticity over the whole dynamics.

%%%%%%%%%%%%%%%%%%%
\paragraph{Outline --}
%%%%%%%%%%%%%%%%%%%
The simplest relevant SGF approximation of SGD is briefly described in Section~\ref{SGD-SGF-approx-review}. The general path integral is formulated in Section~\ref{path-integral-form}, where we also derive the Laplace approximation, test the theory against a simple solvable model, and apply it to a general learning theory setting. In Section~\ref{excatly} we compute the whole time evolution of the test risk under SGF for the weak features model, and compare the theoretical predictions to numerical simulations of SGD\@.

%%%%%%%%%%%%%%%%%%%
\paragraph{Reproducibility --}
%%%%%%%%%%%%%%%%%%%
Code is available in this \href{https://github.com/rodsveiga/sgf\_dyn}{repository}.

%%%%%%%%%%%%%%%%%%%
\section{Stochastic modeling of SGD}\label{SGD-SGF-approx-review}
%%%%%%%%%%%%%%%%%%%
Consider a data set composed of $n$ pairs $(\x^k , y^k )_{k\in[n]} \in \R^{d+1} \sim \Prob ( \x, y) $ i.i.d.~and a model parametrized by $\bbh\in\R^d$. Starting with $\bbh^0$, the usual full-batch GD update is $\bbh^{\i+1}  = \bbh^\i -  \lr \grad   \lossL\bigl(\bbh^\i ;\X,\y\bigl)$, where $\lossL\bigl(\bb;\X,\y\bigl) =  \frac{1}{n} \sum_{k=1}^n \loss \bigl( \bb ; \x^k , y^k  \bigl) $ is the total training loss, with $\X = [ \x_1 | \dots | \x_n ]^\top \in \R^{n\times d}$ being the data matrix, $\y = [ y_1, \dots, y_n  ]^\top \in \R^n $ the target vector and $\loss \bigl( \bb ; \x^k , y^k  \bigl)$ a penalization function. The parameter $\lr > 0$ is the \emph{learning rate} and $\i=0,1,2,\dots$. The GD rule can be thought of as a discretization of the continuous ordinary differential equation (ODE): 
\begin{equation}\label{sgf-ode}
   \dv{\bbh(t)}{t}  =  -  \grad \lossL\bigl(\bbh(t) ;\X,\y\bigl) \;,
\end{equation}
with $\bbh(0) = \bbh^0$, known as \emph{gradient flow} (GF).

Stochastic gradient descent (SGD), on the other hand, does not make use of the whole training set at each optimization step. In its \emph{single batch} form, it can be written as $\bbh^{\i+1}  = \bbh^\i - \lr \grad \loss \bigl( \bbh^\i ; \x^{k_\i} , y^{k_\i}  \bigl) $ with $k_\i$ sampled uniformly in $\{1,n\}$. In general, one could perform \emph{batch} SGD by sampling a set $\mathcal{B}^\i$ of pairs, and replace the gradient above by $\left(\nicefrac{1}{\abs{\mathcal{B}^\i}} \right)\sum_{k \in \mathcal{B}^\i}  \loss \bigl( \bbh^\i ; \x^{k} , y^{k} ) $. Conceptually, the general path-integral approach we introduce in Section~\ref{path-integral-form} does not change with the batch size. Therefore, for the sake of simplicity in the exposition, we adopt single batch SGD in the mathematical formulation. The expressions can be straightforwardly extended to the batch case by replacing the single sample gradient by the average gradient over the batch.

Summing and subtracting the total training loss gradient $\lr \grad\lossL\bigl(\bbh^\i;\X,\y\bigl)$ on the SGD update makes the effective noise explicit:
\begin{equation}
      \label{eq:sgd2}
          \bbh^{\i+1}
     = 
    \underbrace{\bbh^\i 
    -  \lr \grad\lossL\bigl(\bbh^\i;\X,\y\bigl)}_{\text{Full-bath GD}}  
     -  \lr \underbrace{\xib_{k_\i} \bigl(\bbh(t) ;\X,\y\bigl) }_{\text{Stochastic perturbation}} \;,
\end{equation}
where the stochastic perturbation is by definition:
\begin{equation}
    \label{eq:xi}
   \xib_{k_\i} \bigl(\bbh(t) ;\X,\y\bigl)  \equiv  \grad \loss \bigl( \bbh(t) ; \x^{k_\i} , y^{k_\i}  \bigl) - \grad \lossL\bigl(\bbh(t);\X,\y\bigl) \;.
\end{equation}
Since $k_\i$ is sampled uniformly in $\{1,n\}$, the stochastic perturbation is a zero mean vector, $ \E_{k_\i \sim {\cal U} \{ 1 , n \} } \left[ \xib_{k_\i} \bigl(\bbh(t) ;\X,\y\bigl)  \right]  = \bm{0}$. In the spirit of \emph{gradient flow}, Eq.~\eqref{eq:sgd2} can be thought of as a discretization of a stochastic differential equation (SDE), which we will call \emph{stochastic gradient flow} (SGF), in the continuous-time limit. Here we choose the simplest representation proposed in~\cite{mandt_2015} and~\cite{li_2015}:
\begin{equation}
    \label{eq:sdeg}
    \dd \bbh(t) = - \grad \lossL\bigl(\bbh(t) ;\X,\y\bigl) \dd t  
    + \sqrt{ \lr \;  \bm{\Sigma} \bigl(\bbh(t) ;\X,\y\bigl) } \; \dd \brow(t) \;,
\end{equation}
where $\dd \brow(t) = \brow(t + \dd t)-\brow(t) \in \R^n$ is the forward Itô increment associated to a standard $n$-dimensional Wiener process with $\brow(0)=\zero$, $\E [\brow(\time)]=\zero$, $\E [\brow(t)\brow(t^\prime)^\top] = \ID_n \min(t, t^\prime)$, with $\ID_n$ as the $n\times n$ identity matrix. The matrix $\bm{\Sigma}$ is matched to the covariance matrix of the stochastic perturbation in~\eqref{eq:xi}:
\begin{equation}
     \label{eq:diff_gen}
    \bm{\Sigma} \bigl(\bbh(t) ;\X,\y\bigl) \equiv  \E_{k_\i  \sim {\cal U} \{ 1 , n \}} \left[ \xib_{k_\i} \bigl(\bbh(t) ;\X,\y\bigl)  \xib_{k_\i} \bigl(\bbh(t) ;\X,\y\bigl)^\top   \right] \;.
\end{equation}
 This is a $d\times d$ positive semi-definite matrix and its `square-root' is defined as the $d\times n$ matrix such that $\sqrt{\bm{\Sigma}} \sqrt{\bm{\Sigma}}^\top= \bm{\Sigma}$. 

Solving the SGF~\eqref{eq:sdeg} can be far from trivial, since a priori
both the \emph{drift vector} $\grad \lossL\bigl(\bbh(t) ;\X,\y\bigl) $ and the \emph{diffusion matrix} $\bm{\Sigma} \bigl(\bbh(t) ;\X,\y\bigl)$ depend on the stochastic process $\bbh(t)$ and on the data $(\bm{X}, \bm{y})$. We provide in the next section a general method to compute the fluctuations around the deterministic trajectory when the learning rate is small.

%%%%%%%%%%%%%%%%%%%
\section{General path integral formulation}\label{path-integral-form}
%%%%%%%%%%%%%%%%%%%
Inspired by Eq.~\eqref{eq:sdeg}, we consider the following general SDE for a process $\sv (\time) \in \R^d $, sampled between $\tI$ and $t  > \tI $:
\begin{equation}
    \label{eq:sdege}
    \dd \sv (\time) = \f \left(\time,\sv(\time)\right) \dd \time + \sqrt{\gamma} \; \G \left(\time,\sv(\time)\right)  \dd \svbrow(\time) \;,   
\end{equation}
where $\dd \svbrow(\time) = \svbrow(\time+ \dd\time)-\svbrow(\time) \in \R^n$ is the forward Itô increment of a standard $n$-dimensional Wiener process (as defined previously). We assume this SDE has a unique solution. Standard conditions ensuring existence and uniqueness are Lipshitzness and linear growth of $\f$ and $\G$ w.r.t $\sv$ uniformly in $\time$ (see, e.g., \cite{evans2012introduction}, Section 5.B.3). The explicit expressions of the drift vector $\f \left(\time,\sv(\time)\right) \in \R^d$ and the diffusion matrix $\G \left(\time,\sv(\time)\right) \in \R^{d\times n} $ in the learning theory context will depend on the model (data distribution, architecture and loss function). However, the method discussed here is not restricted to machine learning models: it can be viewed as a general approximation scheme to solve SDEs for small $\gamma$. To stress that fact, we generically named the stochastic process as $\sv (\time) $ in this section, which in the learning theory context is the estimator $\bbh(\time)$ itself or some parametrization of it. The general path integral formulation is outlined in this section; the complete detailed derivation is contained in Appendix~\ref{app:cov}.

Let $\Prob_{\gamma} (\sv,t | \sv_{\text{0}},\tI )$ be the transition probability of the process associated to Eq.~\eqref{eq:sdege} to go from $\svI$ at time $\tI$ to $\sv$ at time $t$. If $\gamma=0$ the SDE~\eqref{eq:sdege} becomes a first-order ODE:
\begin{equation}\label{first-order-ODE}
\dv{\sv(\time)}{\time} = \f (\time, \sv(\time)) \;.
\end{equation}
The solution of this ODE with initial condition $\svode(\tI)  = \sv_{0}$ (we assume a unique global solution here) is denoted by $\svode (\time) $. In such case: $\Prob_{\gamma=0} (\sv , t  | \sv_{\text{0}}, t ) = \delta\left(\sv - \svode (t) \right) $. Our goal is to study the fluctuations around the deterministic trajectories $\svode (\time)$ when $\lr$ is small.

The continuous-time Itô SDE~\eqref{eq:sdege} must be understood according to its discrete-time companion process $\sv_k = \sv(k \Delta \time)$:
\begin{equation}
\label{eq:sde_discrD}
   \sv_{k+1} = \sv_{k} + \f(k,\sv_k) \Delta \time + \sqrt{\gamma}\; \G (k,\sv_k) \Delta \etab_k \;,
\end{equation}
where the time interval $[\tI, t]$ has been discretized into $N$ slices of length $\Delta \time \equiv \frac{t - t_0}{N}   $  with $ \time_k = t_0 + k \Delta \time$ for  $ k=0, \dots,N$ and $\time_N = t$. The discrete quantities $\Delta\etab_k$ are sampled independently at each step $k$ from a Gaussian with $\E[\Delta\etab_k] =\zero$, $\E[\Delta\etab_k(\Delta\etab_k)^\top] = \Delta \time \: \ID_n$. In general, the covariance matrix $\G (k,\sv_k) \G (k,\sv_k)^\top$ is only positive semi-definite, and in particular when $d>n$ it certainly has zero eigenvalues. Thus the process might become singular in sub-manifolds of $\R^d$. This is cured by introducing a regularization parameter $\epsilon >0$ and replacing Eq.~\eqref{eq:sde_discrD} by 
\begin{equation}
    \label{eq:dWdistr}
    \Delta \sv_k \equiv \sv_{k+1} - \sv_k  \sim  \N\left( \fk \Delta\time, \gamma\Delta\time\left(  \Gk \GkT + \eps\ID_d  \right)\right) \;,
\end{equation} 
with $\fk\equiv\f(k,\sv_k)$ and $\Gk\equiv \G (k,\sv_k)$.

In order to compute the continuous time-limiting transition probability and associated expected values, the correct prescription is to take the limit $\epsilon\to 0$ {\it after} $N\to +\infty$. 

The one-step propagator $\Prop$ associated with Eq.~\eqref{eq:sde_discrD} is known from Eq.~\eqref{eq:dWdistr}. Then from the Chapman-Kolmogorov equation, $\ProbFI$ equals:
\begin{equation}
    \ProbFI =     \lim_{\epsilon\to 0}\lim_{N\to\infty} \int \prod_{k'=1}^{N-1} \dd \sv_{k'} \prod_{k=0}^{N-1} \Prop \;,
\end{equation}
which yields
\begin{equation}
\label{discretized-path-integral}
\ProbFI  =  
    \lim_{\epsilon \to 0}\lim_{N\to\infty} 
    \int \prod_{k'=1}^{N-1} \frac{\dd \sv_{k'}}{\left( 2 \pi \gamma \Delta\time \right)^{\nicefrac{d}{2}} \left( \det  \Omke \right)^{\nicefrac{1}{2}}}\\
     \exp\  \left[ -  \frac{\Delta\time}{2 \gamma } \sum_{k=0}^{N-1} \left(\dsvdtD  -  \fk \right)^\top
     \left( \Omke \right)^{-1} \left(\dsvdtD  - \fk  \right)   \right] \;,
\end{equation}
where $\Omke \equiv \Gk   \Gk^\top  + \eps \ID_d$ is the regularized covariance.
 
In the continuous-time limit, the expression above becomes a {\it path integral} over all possible trajectories $\sv(\time)\in \mathbb{R}^d$ connecting $\sv_0$ and $\sv$ from time $\tI$ to $t$:
\begin{equation}
\label{eq:0funct_integral}
    \ProbFI =  \lim_{\eps \to 0 }  \int^{\sv(t) = \sv}_{\sv(\tI) = \svI} \Dsvt  \exp\left( - \frac{1}{\gamma } \Svte \right)  \;.
\end{equation}
The \emph{action} functional is defined as 
\begin{equation}\label{action}
    \Svte \equiv  \int_{\tI}^{t} \dd\time   \;  \LagrfE \;,
\end{equation}
with the {\it Lagrangian}:
\begin{equation}
      \label{eq:lagrangian}
        \LagrfE =
    \frac{1}{2} \left( \dv{\sv(\time)}{\time}  - \f \left(\time,\sv(\time)\right)\right)^\top  \OmkeC^{-1}
    \left( \dv{\sv(\time)}{\time}   - \f \left(\time,\sv(\time)\right)\right)  \;,
\end{equation}
with $\dot{\sv}(\time)= \dv{\sv(\time)}{\time}$, $\OmkeC \equiv \G \left(\time,\sv(\time)\right) \G \left(\time,\sv(\time)\right)^\top + \eps  \ID_d$ and $\Dsvt$ as defined in Eq.~\eqref{eq:app_path_measure} of the Appendix~\ref{app:cov}.

The path measure and action functional are to be understood under the Itô discretization~\eqref{discretized-path-integral}. If the SDE had been formulated with Stratonovich or other intermediate Brownian increments, the path integral would have to be understood accordingly~\cite{pirey_2023}.

\begin{figure}[ht]
\vskip 0.1in
\begin{center}
\centerline{\includegraphics[width=0.55\columnwidth]{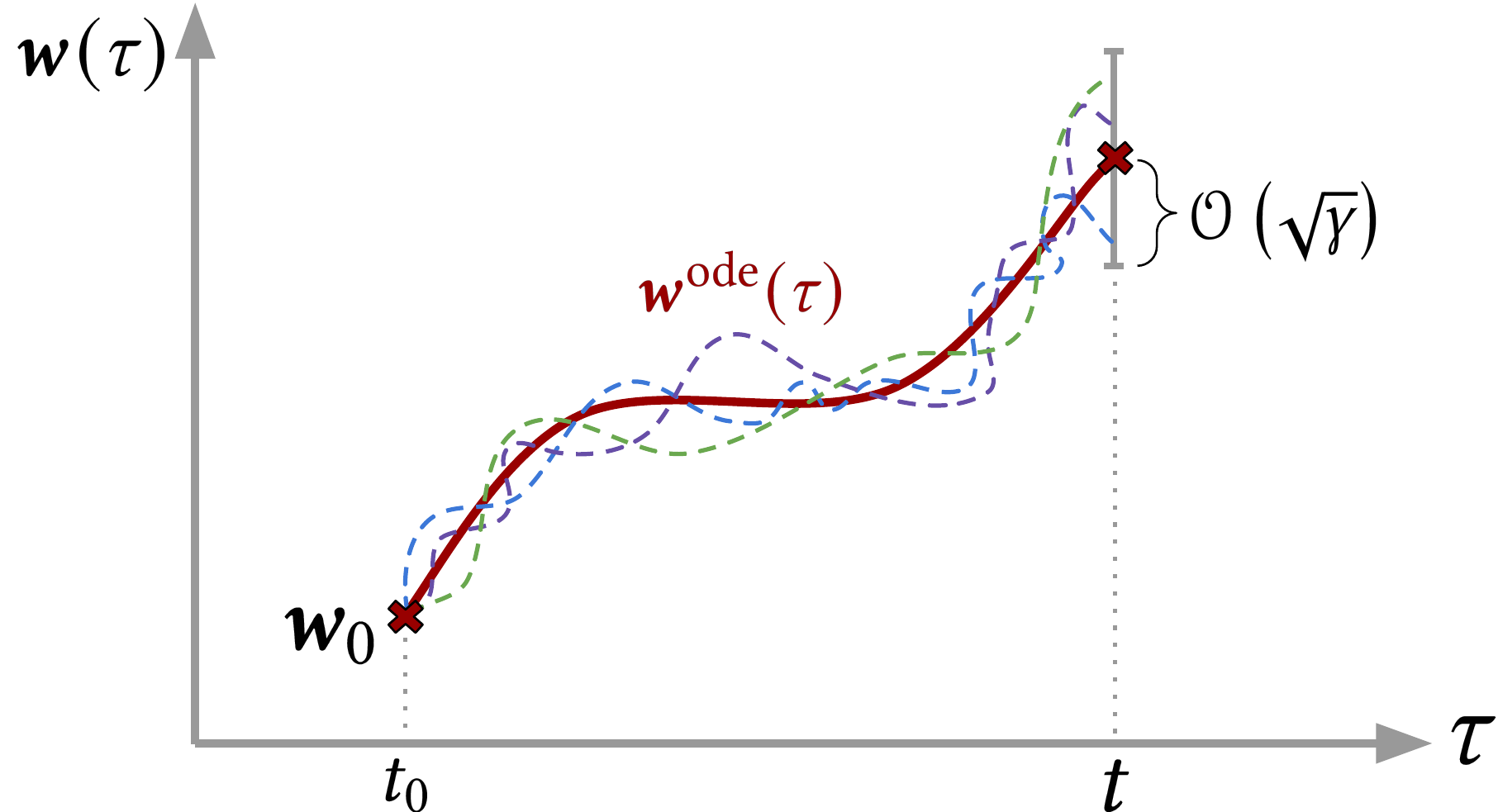}}
\caption{The solution $\svode(\time)$ of the first order ODE $\dot{\sv}(\time) = \f (\time, \sv(\time))$, $\sv(0)= \sv_0$ yields a minimum vanishing action. The dominant contributions to the path integral for small $\gamma$ are fluctuations of order $\O (\sqrt\gamma)$ around this trajectory.}
\label{fig:traj}
\end{center}
\vskip -0.2in
\end{figure}

%%%%%%%%%%%%%%%%%%%
\paragraph{Laplace approximation --} 
%%%%%%%%%%%%%%%%%%%
For small enough  $\gamma$, the path integral~\eqref{eq:0funct_integral} is dominated by
trajectories minimizing the action $\Svte$. This is the path given by Hamilton's action principle (e.g., like in classical analytical mechanics)
$\delta \Svte = 0$
for variations over paths satisfying the {\it two boundary conditions} $\sv(\tI)= \sv_0$, $\sv(t)=\sv$. 
A path is a stationary point of the action if and only if it satisfies the Euler-Lagrange equations:
\begin{equation}
    \label{eq:EL}
   \left( \dv{\time} \pdv{\dot{\svscalar}_j} - \pdv{\svscalar_j} \right) \LagrfE  = 0
\end{equation}
for $j=1,\dots,d$, supplemented by the two boundary conditions. Replacing~\eqref{eq:lagrangian} into the Euler-Lagrange equations~\eqref{eq:EL} renders a complicated set of second-order differential equations. However, it is not very difficult to prove that (see Appendix~\ref{app:highprob}): 
 \begin{itemize}
    \item If the terminal point $\w$ at time $t$ is {\it on} the trajectory $\svode (\cdot)$ i.e., it is on the unique solution of the first-order initial value problem $\dot{\sv}(\time)=\bm{f}(\time, \sv(\time))$, with $\sv(0)=\sv_0$, then the solution of the Euler-Lagrange equation \eqref{eq:EL} is precisely
    $\svode(\time)$ and $\LagrfE =0$.
    \item If the terminal point $\w$ at time $t$ is {\it not on} the trajectory $\svode(\cdot)$, then the solution of the Euler-Lagrange equation cannot satisfy exactly the first order ODE i.e., $\dot{\sv}(\time) \neq \f (\time, \sv(\time))$ and $\LagrfE >0$.
\end{itemize}
In particular, we conclude that for small $\gamma$'s:
\begin{itemize}
\item
When the terminal condition is away from the trajectory $\svode(\cdot)$ by an amount $\vert\sv(t) - \svode (t)\vert =  \Omega_\gamma(1)$, then $ \ProbFI  = \O (e^{-C/\gamma})$ is an exponentially small probability. 
\item
When the terminal condition is close enough to the trajectory $\sv(\cdot)$ in the sense $\vert\sv(t) - \svode (t)\vert =  \O (\sqrt\gamma)$, then the $\ProbFI $ is given by the fluctuations around the stationary point of the action functional. This picture is schematically illustrated in Fig.~\ref{fig:traj}. 
\end{itemize}

To find the probability distribution of these fluctuations, we set
$\sv(\time) = \svode (\time) + \sqrt{\lr} \; \z (\time)$
with $\z(0) = \zero$\footnote{At time $t$ we have $\sv = \svode(t) + \sqrt\gamma \z(t)$ and because $\sv$ is `random' we should not constrain the terminal condition $\z(t)$.} and expand the action functional~\eqref{action} to second order in $\sqrt\gamma\z(\cdot)$. The zero-th order term vanishes because $\svode(\time)$ satisfies the first order ODE, and the first order term also vanishes because it satisfies the Euler-Lagrange equation. Thus we find:
\begin{equation}
    \label{gaussian-fluctuations}
   \Svte \approx \frac{\gamma}{2} \int_{\tI}^{t} \dd\time   \left( \dv{\z (\time)}{\time} - \Jacf(\time,\svode(\time) ) \z(\time) \right)^\top     
          \Om^\eps ( \time,  \svode (\time)   ) 
          \left( \dv{\z (\time)}{\time} - \Jacf(\time,\svode(\time)) \z(\time) \right) \;,
\end{equation}
where $\Jacf$ is the Jacobian matrix of $\bm{f}$ along the trajectory $\svode(\cdot)$ (see Eq.~\eqref{eq:app_jacdef} for an explicit definition). Eq.~\eqref{gaussian-fluctuations} gives us the Gaussian probability distributions of fluctuations $\z(\cdot)$ around the trajectory $\svode(\cdot)$ (see Fig.~\ref{fig:traj}). This has zero mean $\E[\z(\cdot)] = \zero$ and covariance $\C (\cdot) \equiv \E [\z(\cdot) \z(\cdot)^\top]$ given by
\begin{subequations}
    \label{general-final-cov}
    \begin{equation}
        \C(t) =  \int_{\tI}^{t} \dd \time  \;  \U^\text{ode} (t, \time) \; \Om^0 (\time, \svode(\time)) \;  \U^\text{ode} (t, \time)^\top \;,
    \end{equation}
where
    \begin{equation}
        \U^\text{ode} (t, \time)  \equiv  \T\left\{ e^{- \int_t^\time \dd s \Jacf (s, \svode(s)) } \right\}  \;,
    \end{equation}
\end{subequations}
with $\T \{\cdots\}$ denoting the time-ordered exponential (see Appendix~\ref{app:time_ord}). The complete derivation of the covariance is found in Appendix~\ref{app:cov}.

Using Wick's theorem, this covariance allows us to compute any expectation that reduces to multinomials of $\z(t_1), \z(t_2), \dots, \z(t_m)$. In particular, we will see below that the generalization error involves the simplest such polynomial $\z(t)^\top \z(t)$ at the terminal time.

Finally, within our approximate theory, we have 
\begin{align}\label{gaussian-final}
    \mathbb{P}_\gamma(\w, t | \sv_0, \tI)
    \approx 
    \frac{e^{-\frac{1}{2}(\sv - \svode(t))^\top \C(t)^{-1} (\sv - \svode(t))}}{(2\pi)^{d/2}(\det \C(t))^{1/2}}
\end{align}
We note that for $d>n$ the $d\times d$ covariance matrix has zero eigenvalues and this expression is somewhat formal. This means that the probability distribution has its support in an $n$-dimensional subspace of $\R^d$. In the context of machine learning, $d>n$ represents the over-parameterized regime.

%%%%%%%%%%%%%%%%%%%
\subsection{An exactly solvable example}\label{excatly}
%%%%%%%%%%%%%%%%%%%
In this short section, we verify the above result for a simple SDE which is exactly solvable. Consider the one-dimensional stochastic process $\{w(\time), \time\geq t_0 \}$ defined by 
\begin{align}\label{simple-SDE}
    \dd w(\time) = h(\time) (y - w(\time)) \dd \time + \sqrt{\gamma}\sigma(\time) \dd\eta(\time)
\end{align}
with initial condition $w(\tI)=w_0$. The functions $h$ and $\sigma$ are real valued continuous for $\time\in [0,1)$, $y \in \R $,  and $\eta$ standard one dimensional Brownian motion. One checks explicitly that the solution is:
\begin{align}
    w(\time) = \phi(\time) w_0 + (1-\phi(\time)) y + \sqrt{\gamma}\phi(\time) \int_{t_0}^\time  \dd \eta(s) \frac{\sigma(s)}{\phi(s)}
\end{align}
where $\phi(\time) =  \exp(- \int_{t_0}^\time \dd s  \; h(s))$. This equation implies that the process is Gaussian with transition probability $\mathbb{P}_\gamma(w, t|w_0, t_0)$ with mean and variance:
\begin{subequations}
    \begin{align}
        \E [w(t)]  &=  y + (w_0 -y) e^{-\int_{t_0}^t \dd s \;  h(s)}  \;,  \label{mean-onedim} \\ 
        \Var [w(t)]   &=   \gamma \int_{t_0}^t \dd u\; \sigma(u)^2 e^{-2\int_{u}^t \dd s  \; h(s)} \label{var-onedim} \;.
    \end{align}
\end{subequations}

It is an exercise to check that~\eqref{mean-onedim} coincides with $w^{\text{ode}}(t)$, the solution of the first order ODE $ \dv{w(t)}{t} = h(t) (y - w(t))$ and~\eqref{var-onedim} coincides with~\eqref{general-final-cov}. We conclude that our general theory is exact for the simple linear SDE~\eqref{simple-SDE} .

In~\cite{hildebrandt_2020} the authors discuss in detail a {\it pinning phenomenon} which is of relevance to us. Suppose $h$ and $\sigma$ satisfy (i) $h\geq 0$; (ii)
$\lim_{t\to +\infty}\int_{t_0}^\infty \dd s\; h(s)  = +\infty$; (iii) $\int_{t_0}^\infty \dd s  \sigma(s)^2 $ is finite. Then we have for $t\to + \infty$, $\E[w(t)] \to y$ and $\Var[w(t)] \to 0$ (the later point can be shown by noticing that $\Var [w(t)]\leq \gamma \int_{t_0}^{+\infty} \dd u\; \sigma(u)^2 e^{-2\int_{u}^t \dd s \; h(s)}$ and applying the dominated convergence theorem to the right-hand side; for more details, we refer to~\cite{hildebrandt_2020}). These diffusion processes become `pinned' onto $y$ as $t\to +\infty$. For a simple picture, one may keep in mind two continuous functions, such that for $r\to +\infty$, on one hand, $h$ does not tend to zero fast enough and on the other hand, $\sigma(r)$ tends to zero fast enough.  
Roughly speaking for large times the process fluctuates around the gradient flow path for a `loss function' $\frac{1}{2} h(\time)(y-w)^2$ and for large times as the fluctuations become smaller it tends to the minimum. 

Note that if $\sigma(s)^2$ is not integrable, pinning does not necessarily occur. For example, suppose $\sigma(s) = b$ and $h(s)=a>0$ two constants. Then an explicit computation shows $\mathbb{E}[w(t)] \to y$ and $\Var [w(t)] \to \gamma b^2/2a$.

%%%%%%%%%%%%%%%%%%%
\subsection{Test risk of SGF}\label{test-sgf}
%%%%%%%%%%%%%%%%%%%
We now come back to the setting of learning theory. Comparing equations~\eqref{eq:sdeg} and~\eqref{eq:diff_gen}, we have the correspondence
$\sv(t) \to \bbh(t)$, 
$\f(t, \sv(t))\to - \grad \lossL\bigl(\bbh(t) ;\X,\y\bigl)$, $G(t, \sv(t)) \to \sqrt{\bm{\Sigma} \bigl(\bbh(t) ;\X,\y\bigl) }$ and $\dd\bm{\eta}(t)\to\dd \brow(t)$. Therefore, we approximate the solution of~\eqref{eq:sdeg} by $\bbh(t) =  \bbh^{\text{GF}}  + \sqrt\gamma \z(t)$ with $\bbh^{\text{GF}}$ the solution of 
\eqref{sgf-ode} and the Gaussian process $\z(t)$ has covariance (setting now $t_0=0$):
\begin{subequations}
    \label{cov-ML}
    \begin{equation}
        \C(t) = \int_0^t 
        \dd u  \;  \U^\text{GF} (t, u) \; \bm{\Sigma} (u, \bbh^\text{GF}(u)) \;  \U^\text{GF} (t, u)^\top \label{cov-ML-1}  \;,
    \end{equation}
where now
    \begin{equation}
        \U^\text{GF} (t, u)   =   \T\left\{ e^{ - \int_u^t \dd r \HessL ( \bbh^{\text{GF}}(r)) } \right\} \label{cov-ML-2} \;,
    \end{equation}
\end{subequations}
with $\HessL = \grad \otimes \grad \lossL$ the $d\times d$ Hessian matrix of the loss function (see also~\ref{sec:app1_ml0} for a slightly more general setting). 

From these formulas, we may expect a set of qualitative behaviors for the generalization error, which we now discuss. We stress, however, that these are by no means necessarily universal and their validity should be checked for each model. For $d< n$, below the interpolation threshold, $\bm{\Sigma}$ is generically full rank, so the covariance is also expected to be full rank.
In other words, we expect the stochastic process to fluctuate in parts of $\R^d$ of `full measure' even for large times. The difference between the generalization errors of GF and SGF should be non-trivial. On the other hand, above the interpolation threshold, $\bm{\Sigma}(\time, \bbh^{\text{GF}}(\time))$ certainly has $d-n$ zero eigenvalues (and possibly more). Therefore, we expect the process to fluctuate in a `submanifold' of $\R^d$. Such submanifold can be quite complicated as it will in general depend on time and on the GF trajectory. For large times we furthermore expect that $\bm{\Sigma}$ tends to zero altogether. Indeed, for linear networks this matrix is proportional to the loss (see next section), which tends to zero above the interpolation threshold; and more generally if the number of data points is much less than the number of parameters, the data points are seen many times and there should not be a major difference between a full batch and a mini-batch. Consequently, for large times we expect that the diffusion is pinned on the minima of the loss, and that there should be no essential difference between generalization error of GF and SGF\@. These considerations are corroborated in the next section by an explicit analysis of a simple model.

%%%%%%%%%%%%%%%%%%%
\section{Application to a weak features model}\label{sec:simple-model-regression}
%%%%%%%%%%%%%%%%%%%
\subsection{Setting}

We consider a special regression model with random projections, first studied by~\cite{breiman_1983} in the underparametrized regime, and later extended by~\cite{belkin_2020} to the overparametrized regime. Despite its simplicity, the test risk displays the double descent shape, calculated explicitly in~\cite{belkin_2020} for the least-squares estimator.\footnote{Here the least-squares estimator equals the GF estimator for infinite times.} The model is a suitable laboratory to apply our formalism and compute the whole time-dependence of the test risk for SGF, and compare it with the GF solution. 

Again, the data set is composed of $n$ pairs $(\x^k , y^k) \in \R^{d+1}$, $k\in [n]\equiv\{1, \dots, n\}$ sampled i.i.d. from $\Prob ( \x, y)$. The distribution generating the data features is Gaussian, $\Prob (\x) =  {\cal N} ( \bm{0}, \ID_d )$, while $ \Prob (y | \x) $ is modeled by a linear function of $\bb \in \R^d$, $\Vert\bb\Vert = 1$,  $ y^k = \bb^\top \x^k + \mu \epsilon^k$ with $k\in [n]$. The constant $\mu >0 $ controls the strength of the additive Gaussian noise $\epsilon^k \sim {\cal N} ( 0,1 )$. The feature matrix is denoted $\X = [ \x^1 | \dots  | \x^n ]^\top \in \R^{n\times d}$.

The weights  $\bbh  \in \R^{d} $ are learned using only a subset $ \sub  \subseteq [d] \equiv \{ 1,\dots,d  \}$ of $p\equiv \abs{\sub}$ components. For a vector $\bm{v} \in \R^d$, we denote $\bm{v}_\sub \equiv \left[ v_j : j \in \sub \right]^\top \in \R^p $ its subvector of entries from $\sub$; besides, we denote $\subc \equiv [d] \backslash \sub$ to be complement of $\sub$. Then, the training loss is defined as
\begin{equation}
    \lossL\bigl(\bbh;\X,\y; \sub\bigl)  = \frac{1}{n} \sum^n_{k = 1} \loss \bigl( \bbh ; \x^k , y^k;\sub  \bigl)    = \frac{1}{n} \sum^n_{k = 1} \frac{1}{2} \left( y^k  - \betas^\top \x^{k}_\sub   \right)^2 \;.
\end{equation}
Only components $\betas \in \R^p$ are learned, while $\betasc \in \R^{d - p}$ are set to zero. Besides, we denote  $\X_\sub = \left[ \x^1_\sub | \dots  | \x^n_\sub \right]^\top \in \R^{n\times p}$ as a matrix of features used for regression. 

The SGD process associated with training of the regression leads to the SDE modeling of Eq.~\eqref{eq:sdeg}. From~\eqref{eq:diff_gen} we approximate the diffusion matrix with (see Appendix~\ref{app:sde_modeling}):
\begin{equation}
    \label{eq:diff_gen_belkin}
    \bm{\Sigma} \bigl(\bbh(t) ;\X,\y;\sub\bigl) \approx \frac{2}{n} \lossL\bigl(\bbh(t);\X,\y;\sub\bigl) \X^\top_\sub  \X_\sub,
\end{equation}
which results in the SGF:
\begin{equation}
     \label{eq:belkin_sde}
        \dd \betas = \frac{1}{n}  \X_{\sub}^\top  ( \y  - \X_{\sub} \betas)    \dd t
        + \frac{\sqrt{\lr}}{n} \norm{\y  - \X_{\sub} \betas} \;  \X^\top_\sub  \dd \brow \;. 
\end{equation}
For $\gamma = 0$ this becomes the respective pure GF equation. Setting $\tI = 0$ and assuming the initial condition  $\betas(0) = \bbhzs$, we obtain in Appendix~\ref{app:belkin:gf} the exact GF trajectory:
\begin{equation}
    \label{eq:belkin_ode_sol}
       \betasode(t)  =    
       e^{ - \frac{1}{n} \X^\top_{\sub} \X_{\sub} t}  \bbhzs +  \X^\dagger_{\sub} \left( \IDn  - e^{-\frac{1}{n}  \X_{\sub} \X^\top_{\sub} t }  \right) \bm{y} \;,
\end{equation}
where $\dagger$ denotes the Moore-Penrose inverse. Note that this tends to the usual least-squares estimator for $t\to +\infty$. Translating to the general setting of Section~\ref{path-integral-form}, the solution of SGF will have the form $\betassde(t) = \betasode(t) +  \sqrt{\lr} \z (t)$, with $\z(0)=\bm{0}$  and  $\z(t)$ being a zero mean Gaussian process with the covariance matrix 
$\C(t)$.

%%%%%%%%%%%%%%%%%%%
\subsection{Test risk of GF and SGF}\label{sec:test-GF-SGF}
%%%%%%%%%%%%%%%%%%%
We are particularly interested in the test risk provided by GF and SGF and how they compare. Let $\sub$ be a uniformly random subset of $[d]$ of cardinality $p$. If $(\x, y)$ is a new sample from the distribution $\Prob ( \x, y)$, the test risk on it is defined as:
\begin{align} 
    \label{eq:etestsde_def}
    \Etestsde  &= \frac{1}{2}
    \E_{\sub; \X, \y ; \brow;  \x,  y }  \left[ \left( y -  \x^\top_{\sub}  \betassde(t)\right)^2 \right] 
    \nonumber \\ 
     &=  \frac{1}{2}
    \E_{\sub; \X, \y ; \brow }  \left[ \left\| \bb_\sub - \betassde(t) \right\|^2 + \left\| \bb_{\subc} \right\|^2\right] + \mu^2 \;.  
\end{align}
As $ \Etestsde\big\vert_{\gamma = 0} =  \Etestode$ and, for any fixed train data and choice of features, $\E_{\brow} \left[   \z (t)  \right]  = \zero$, we get
\begin{align} \label{sum-of-two-terms}
    \Etestsde  & = \Etestode  
    + \frac{\gamma}{2} \E_{\sub; \X, \y; \brow} \left[  \z^\top (t)  \z (t)   \right] 
    \nonumber \\ &
    =
     \Etestode +  \frac{\gamma}{2} \E_{\sub; \X, \y } \left[ \Tr{\C (t) } \right] \;.
\end{align}
Therefore, the inclusion of the perturbation has an additive effect on the test risk proportional to the learning rate $\lr$.

Let $\Lsvdr \in \R^{r \times r}$ with  $r = \min(n, p)$, be a diagonal matrix consisting of the $r$ highest singular values of $\X_\sub$; with probability $1$ they are all strictly positive. Besides, all other singular values of $\X_\sub$ are zero. From the GF solution, the test risk can be expressed as (see Appendix~\ref{app:belkin:gf_test} for a detailed derivation and Appendix~\ref{app:add_numerics} for plots of the GF test risk):
\begin{equation}
        \label{eq:etest_ode}
      2 \: \Etestode
      =   
      \frac{ \| \bb  - \bbhz \|^2 }{ d}    
      \left[ \max(0, p - n)
     +  \E_{ \Lsvdr} \Tr     e^{- 2  \frac{\Lsvdr^2}{n} t }
     \right]
     +  \left( \left(1 - \frac{p}{d} \right)   \| \bb  \|^2 +  \mu^2\right) 
     \left[ 1 +
     \E_{\Lsvdr} \Tr       \Lsvdr^{-2} \big( \ID_r  - e^{-  \frac{\Lsvdr^2}{n} t } \big)^2\right] \;.
\end{equation}
When $t \to \infty$, this expression becomes exactly the one from~\cite{belkin_2020} (deduced directly form least-squares and can also be computed for finite $n,p,d$). 

For the second term in~\eqref{sum-of-two-terms} giving the correction to the test error coming from the stochastic term we get 
\begin{equation}
\label{SGF-finite-size}
    \begin{aligned}
   &  \frac{\gamma}{2} \E_{\sub; \X, \y } \left[ \Tr{\C (t) } \right] =  \\
   & = \frac{\gamma}{2}  \frac{\| \bb  - \bbhz \|^2  }{d}   \int_{0}^{t} \dd \tau  \: \E_{\Lsvdr} \left[ \Tr{  \frac{\Lsvdr^2}{n}     e^{ - 2\frac{\Lsvdr^2}{n}       (t - \tau) }  }
    \Tr{ \frac{\Lsvdr^2}{n} 
      e^{-2\frac{\Lsvdr^2}{n}       \tau }  }   \right]     \\
          & +   \frac{\gamma}{2n}  \left( \left( 1 - \frac{p}{d} \right)   \| \bb  \|^2 +  \mu^2 \right)  \left(  \max \left(0, \frac{ n - p}{2} \right)     \E_{\Lsvdr}  \left[  \Tr{\ID_r - e^{-2\frac{\Lsvdr^2}{n}  t}}  \right] + \int_{0}^{t} \dd \tau \;  
    \E_{\Lsvdr} \left[   \Tr{ \frac{\Lsvdr^2}{n}     e^{ - 2\frac{\Lsvdr^2}{n}       (t - \tau) }  }      \Tr{ 
    e^{-2\frac{\Lsvdr^2}{n}  \tau }  }  \right]    
    \right)      \;.
    \end{aligned}
\end{equation}

In the remaining part of this paragraph, we discuss the {\it asymptotic regime} $n, d, p \to \infty$ with $p/n \to \alpha$, $d/n\to \psi$ with $\alpha$ and $\psi$ fixed positive numerical constants. Note that $p\leq d$ hence  $\alpha \leq \psi$. At the same time, in order to obtain a well-defined limit for \eqref{SGF-finite-size} one should rescale the learning rate in SGF as follows $\gamma = \gamma'/d$ with $\gamma'$ fixed. Indeed, in~\eqref{eq:sdege} the order of magnitude of the diffusion term is $\E \norm{\sqrt{\gamma}  \; \dd\svbrow(\time)}^2 = \gamma n   \dd \time$ and the scaling ensures that it is of the same order $\dd\tau$ as the drift. This sort of scaling has been discussed in~\cite{veiga2022,ben_arous_2022}. We observe that the diffusion contribution vanishes if $\gamma$ goes to zero faster than $1/d$. 

With our normalizations the empirical distribution 
$n^{-1}\#\{\text{eigenvalues of } \bm{\Lambda}^2_r /n \in [a,b]\}$
tends weakly to the (a.c part of) Marchenko-Pastur law:
\begin{equation}
    \nonumber
     \MPalpha([a,b]) = \int_a^b
 \dd \sigma\frac{\sqrt{(\alpha_+ - \sigma)(\sigma - \alpha_-)}}{2\pi\alpha\sigma} \Ind(\sigma\in [\alpha_-, \alpha_+]) \;,
\end{equation}
 with $\alpha_{\pm}=(1\pm \sqrt\alpha)^2$; 
 from which we can express the expectations of traces in \eqref{eq:etest_ode} in the limit $n, d, p\to\infty$ (noting that $ \Vert\bb - \bbhz\Vert^2 \to 2$ for two random vectors on the high dimensional unit sphere): 
\begin{equation}
\label{assympt-1}
   \lim\Etestode =
      \max\left(0, \frac{\alpha -1}{\psi}\right)
     +
    \frac{\alpha}{2 \psi}
\int \rho_\alpha(\dd\sigma)  e^{-2\sigma t}
+ \frac{1}{2}
\left(\left(1-\frac{\alpha}{\psi}\right) + \mu^2\right)\biggl[1+ \alpha\int \rho_\alpha(\dd\sigma)
\frac{(1- e^{-\sigma t})^2}{\sigma}
\biggr] \;. 
\end{equation}

Using the concentration of expressions $n^{-1}\Tr{\varphi(\bm{\Lambda}_r^2/n)}$ for any `reasonable' function $\varphi$, we can compute the limit for $n,p,d\to \infty$ using again the Marchenko-Pastur distribution. With the scaling $\gamma=\gamma'/d$, we get the well-defined limit (again note $\Vert\bb - \bbhz\Vert^2 \to 2$ for two random vectors on the high dimensional unit sphere): 
\begin{equation}
  \label{assympt-2}
      \lim \frac{\gamma}{2}
        \E_{\sub; \X, \y }  \left[ \Tr{\C (t) } \right] =  
           \frac{\gamma' \alpha}{2 \psi} \left[ \frac{2 \alpha}{\psi} F_1(\alpha, t) 
        + \biggl( 1 - \frac{\alpha}{\psi}  + \mu^2\biggr)  
        \left( \alpha \: F_2(\alpha, t) +  \max(1-\alpha, 0)  \int \: \rho_\alpha(\dd\sigma) \: \frac{1 -  e^{-2\sigma t}}{2}    \right) \right]   \;,
\end{equation}
where
\begin{subequations}
    \begin{align}
         F_1(\alpha, t)  &= \int\int\rho_\alpha(\dd\sigma_1)\rho_\alpha(\dd \sigma_2) \sigma_1\sigma_2 K(t,\sigma_1, \sigma_2)\;, \nonumber \\ 
        F_2(\alpha, t)  &=  \frac{1}{2}\int\int\rho_\alpha(\dd\sigma_1)\rho_\alpha(\dd \sigma_2) (\sigma_1+\sigma_2) K(t,\sigma_1, \sigma_2)  \nonumber \;,
    \end{align}
\end{subequations}
and 
\begin{equation}
    \nonumber
    K(t, \sigma_1,\sigma_2) = \frac{e^{-2\sigma_1 t} - e^{-2\sigma_2 t}}{2(\sigma_2 - \sigma_1)} \;.
\end{equation}

%%%%%%%%%%%%%%%%%%%
\subsection{Discussion and comparison with simulations}\label{sec:discussion}
%%%%%%%%%%%%%%%%%%%
Eqs.~\eqref{assympt-1} and~\eqref{assympt-2} give the complete time dependence of the test risk (or average generalization error) under SGF dynamics for small learning rate. To interpret these formulas correctly, one should recall that $\alpha\in [0, \psi]$ and $\psi\geq 0$.

The dominant contribution is given by~\eqref{assympt-1}. The $t\to +\infty$ limit of this expression agrees with the well-known double descent curve obtained by the least-squares estimator. Such finite time dependencies have been obtained and discussed in detail in~\cite{bodin_2021, bodin_2022} for the random features and Gaussian covariate models.

Here we concentrate on the correction term which is our main interest, i.e.,
$\Etestsde - \Etestode$ or Eq.~\eqref{assympt-2}. It can be computed for $t\to +\infty$ after the large size limit $n,p,d\to  \infty$ and we find (see Appendix \ref{app:sgf_testrk} for details):
\begin{align}\label{theory-pred-1}
\frac{\gamma'}{4} \frac{\alpha}{\psi} \left( \left(1-\frac{\alpha}{\psi} \right) + \mu^2 \right) \max(1-\alpha, 0) \; \; , \; \;  \alpha\leq \psi \;.
\end{align}

\begin{figure}[ht]
    \vskip 0.05in
    \begin{center}
\centerline{\includegraphics[width=0.55\columnwidth]{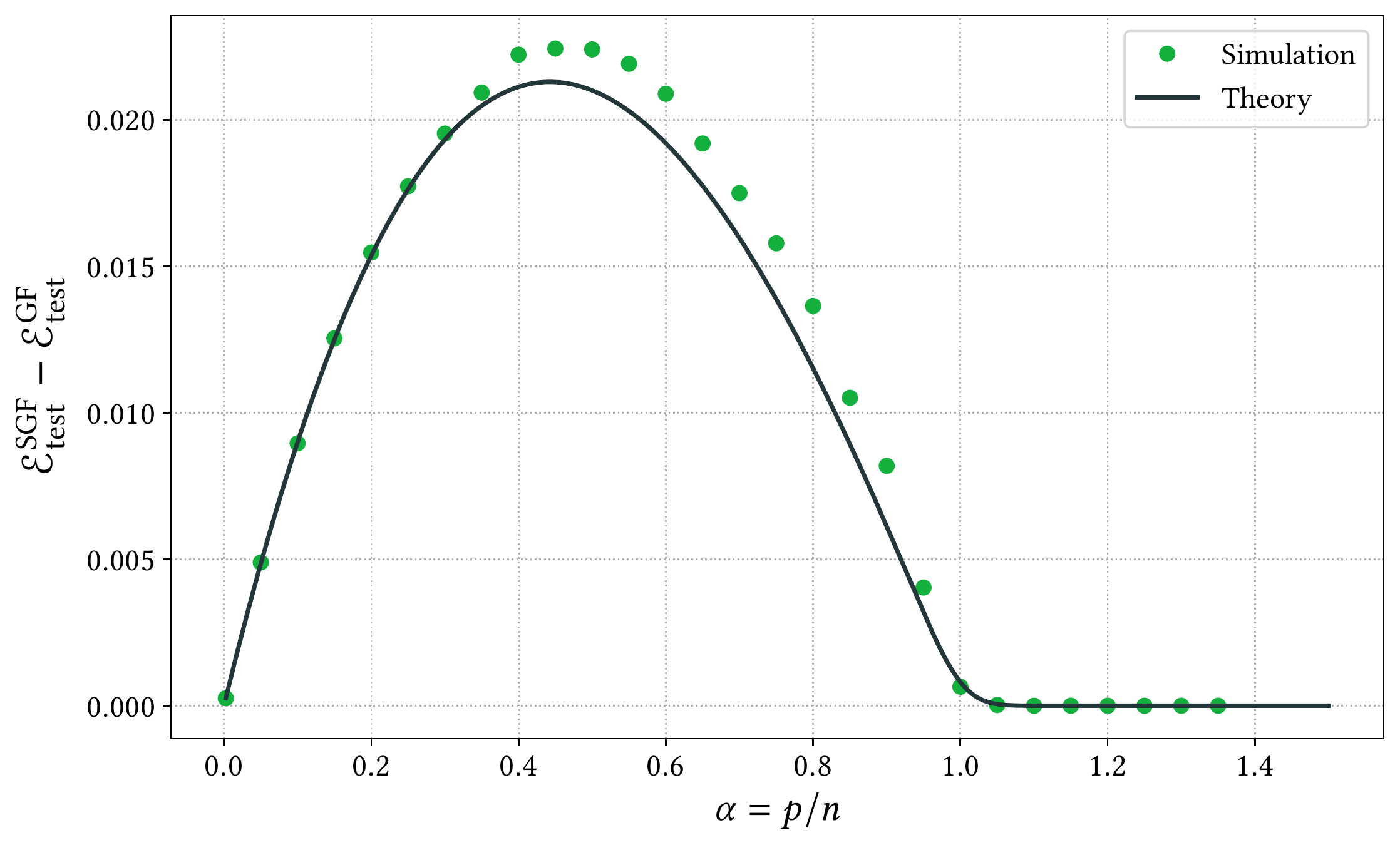}}
    \caption{Continuous curve: the difference between SGF and GF test risks for $t\to + \infty$ according to Eq.~\eqref{theory-pred-1}. Dots: difference of SGD and GD test risks obtained from numerical simulations (averaged over $1000$ different random subsets $\sub$). Simulation parameters: $d=1000$, $n=400$, such that $\psi = 2.5$, and $\lr= 10^{-3}$ (hence $\gamma'=1)$. Vectors $\bm{\beta}$, $\bm{\beta}_0$ are taken at random on the unit $1000$-dimensional sphere and here $\norm{\bm{\beta}-\bm{\beta}_0}^2 \approx 2.11$.}
    \label{fig:cov_gausmodel}
    \end{center}
    \vskip -0.2in
\end{figure}

For $\alpha \leq 1$ this is a cubic polynomial (in $\alpha$) with roots at $\alpha_1=0$, $\alpha_2=1$ and $\alpha_3=(1+\mu^2)\psi$. The last root is always `unphysical' since $\alpha_3 > \psi$. The root  $\alpha_2=1$ (the interpolation threshold) is present only for $\psi\geq 1$. For infinitely large times the correction to GF brought about by the stochastic term in SGF is present only in the underparametrized regime $\alpha<1$. In the overparametrized regime $\alpha > 1$ (hence also $\psi>1$) there is no such correction for infinite time. This result is compared with numerical simulations of the discrete-time stochastic gradient descent in Fig.~\ref{fig:cov_gausmodel}. The agreement is satisfying given that (i) the continuous-time dynamics is a heuristic approximation of stochastic gradient descent and, (ii) the analysis of the stochastic process uses a Laplace approximation of the path integral.  

\begin{figure}[ht]
    \vskip 0.05in
    \begin{center}
    \centerline{\includegraphics[width=0.55\columnwidth]{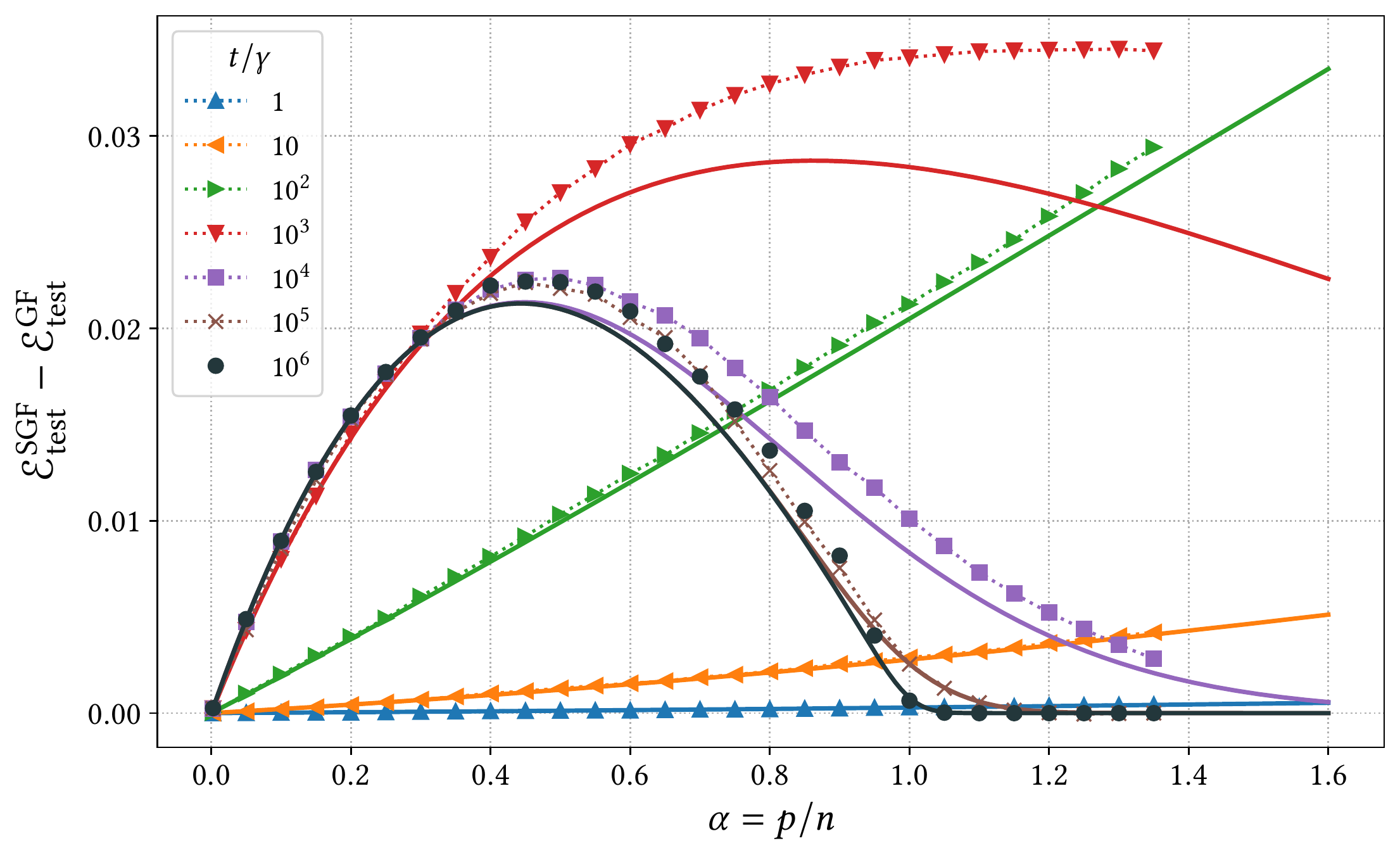}}
    \caption{Continuous curves: difference between test risk of SGF and GF for various times $t\in [10^{-3}, 10^3]$ according to~\eqref{assympt-2}. Dots: difference of test risk of SGD and GD obtained from numerical simulations with the same parameters as in Fig~ \ref{fig:cov_gausmodel}. Asymptotic analytical theory underestimates SGD quantities for intermediate times $1\lessapprox t\lessapprox 10$ but works well for small and large times.}
    \label{fig:cov_fintime}
    \end{center}
    \vskip -0.2in
\end{figure}

\begin{figure}[ht]
    \vskip 0.05in
    \begin{center}
\centerline{\includegraphics[width=0.55\columnwidth]{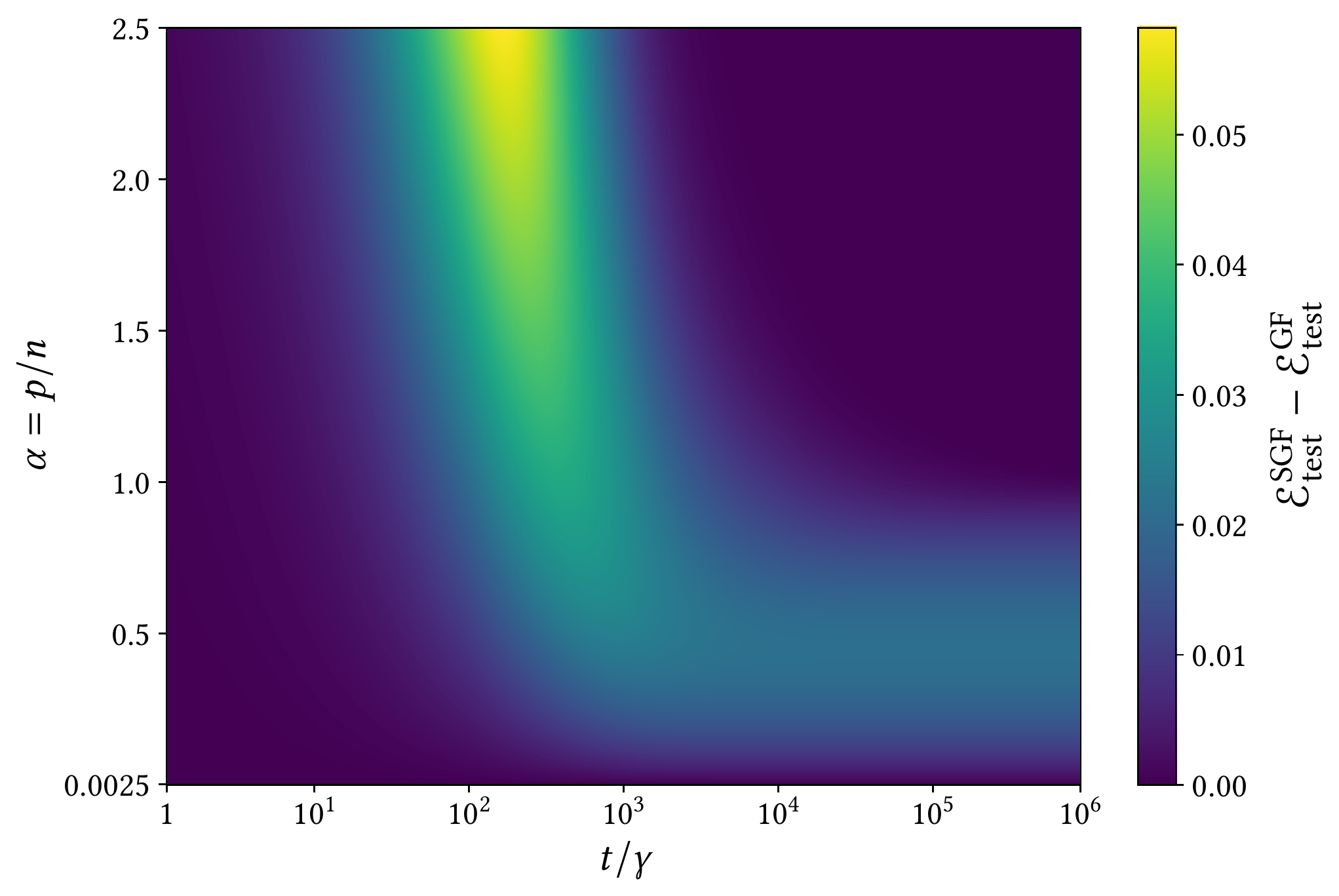}}
    \caption{Theoretical prediction for the difference between test risk of SGF and GF in the $(t, \alpha)$ plane. Here we fix $\psi =2.5$.} 
    \label{heatmap-GF-testrisk}
    \end{center}
    \vskip -0.2in
\end{figure}

Figs.~\ref{fig:cov_fintime} and~\ref{heatmap-GF-testrisk} show various aspects of the finite time behavior of $\Etestsde - \Etestode$. In Fig.~\ref{fig:cov_fintime} we observe that (for $\psi = 2.5$) the maximum difference between the test risk curves occurs above the interpolation threshold for `early' times and this maximum moves below the interpolation threshold at `later' times. For intermediate times, however, our analytical theory underestimates the difference in test risks (see Appendix~\ref{app:add_numerics} for more details).

In Fig.~\ref{heatmap-GF-testrisk} the difference in test risks of GF and SGF shows a knee structure. This bending occurs roughly at times corresponding to the onset of double descent.

%%%%%%%%%%%%%%%%%%%
\section{Conclusion}
%%%%%%%%%%%%%%%%%%%
The general framework developed here gives concrete means to compute the whole-time evolution of the generalization error under stochastic gradient flow dynamics. In particular, we provide general formulas, 
Eqs.~\eqref{cov-ML}, for the covariance of the fluctuations around the deterministic pure gradient flow.

In this work the formalism is applied only to a situation where the dynamics is very simple.
It will be interesting to investigate to what extent one can apply it beyond the weak features model for more complicated landscapes with many minima and saddles. Random features with non-linear activation, diagonal linear networks, generalized linear or multi-index models might still be tractable. More broadly, it is of interest to further develop the theory extracting general properties from the covariance formula by leveraging statistical analysis of the data matrix and of the Hessian.  Recent activity in the literature connects the eigenvalues of the Hessian to the optimization landscape, see e.g.,~\cite{gurari_2018,sagun_2018,xie_2022,sabanayagam_2023}.

There exist more sophisticated SGFs yielding better approximations of SGD~\cite{li_2019}. The analysis is then more complicated as there are new correction terms in the path integral. In the same vein, it would be desirable to investigate the use of alternative processes, other than Brownian motion,  to model the SGD fluctuations.

We plan to come back to these issues in forthcoming works.

%%%%%%%%%%%%%%%%%%%%%%%%%
\section*{Acknowledgements}
%%%%%%%%%%%%%%%%%%%%%%%%%
We acknowledge funding from EPFL and from the Swiss National Science Foundation grant number 200021-204119. 

%%%%%%%%%%%%%%%%%%%%%%%%%%%%%%%
\newpage
\appendix
\section*{\centerline{\huge{-- Appendices --}}\vspace*{0.5cm}}\label{sec:app}
\section{General Path Integral Formulation}
\label{app:cov}

In this appendix, we provide the details on the general path integral framework and on the derivation of the covariance matrix. For easy readability and to make this section as self-contained as possible, some equations from the main part are rewritten here. 

Consider the following general stochastic differential equation (SDE) for a process $\sv (\time) \in \R^d $, sampled between $\tI$ and $t  > \tI $:
\begin{equation}
    \label{eq:app_sdege}
    \dd \sv (\time) = \f \left(\time,\sv(\time)\right) \dd \time + \sqrt{\gamma} \; \G \left(\time,\sv(\time)\right)  \dd \svbrow(\time) \;,   
\end{equation}
where $\f \left(\time,\sv(\time)\right) \in \R^d$ and $\G \left(\time,\sv(\time)\right) \in \R^{d\times n} $ are the drift vector and the diffusion matrix, respectively; while $ \svbrow(\time) $ is a standard n-dimensional Wiener process, with $\dd \svbrow(\time) = \svbrow(\time+ \dd \time)-\svbrow(t) \in \R^n$ as its forward increment. The initial and terminal conditions are $\sv(\tI) = \svI$ and $\sv(t) = \sv$, respectively. We assume that this SDE has a unique solution. Standard conditions ensuring existence and uniqueness are Lipshitzness and linear growth of $\f$ and $\G$ w.r.t $\sv$ uniformly in $\time$ (see, e.g., \cite{evans2012introduction}, Section 5.B.3). 

The continuous-time equation Eq.~\eqref{eq:app_sdege} must be understood according to its discrete-time companion process $\sv_k = \sv(k \Delta \time)$:
\begin{equation}
\label{eq:app_sde_discrD}
   \sv_{k+1} = \sv_{k} + \f(k,\Tilde{\sv}_k) \Delta \time + \sqrt{\gamma}\; \G (k,\Tilde{\sv}_k) \Delta \etab_k \;,
\end{equation}
where the time interval $[\tI, t]$ has been discretized into $N$ slices of length $\Delta \time \equiv \frac{t - t_0}{N}   $  with $ \time_k = t_0 + k \Delta \time$ for  $ k=0, \dots,N$ and $\time_N = t$.  The drift and the diffusion functions are evaluated at the point
\begin{equation}
\label{eq:app_discrD}
    \Tilde{\sv}_k = \w_k + \alpha \Delta\sv_k  \;,
    \end{equation}
with $0\le\alpha\le 1$ and $\Delta\sv_k \equiv \sv_{k+1}-\sv_k$. The choice of $\alpha$ fully determines the discretization scheme and $\alpha\ne0$ introduces additional contributions to the drift on the path-integral~\cite{pirey_2023}. Such contributions are also related to different SDEs as studied in~\cite{li_2019}. In the present work, we adopt the simplest Itô discretization, $\alpha=0$,  $\Tilde{\sv}_k = \w_k $. The discrete quantities $\Delta\etab_k$ are sampled independently at each step $k$ from a Gaussian with $\E[\Delta\etab_k] =\zero$, $\E[\Delta\etab_k(\Delta\etab_k)^\top] = \Delta \time  \; \ID_n$. 

The matrix $\G (k,\sv_k) \G (k,\sv_k)^\top$ is positive semi-definite, and in particular when $d>n$ it certainly has zero eigenvalues\footnote{In the learning theory context, $d>n$ means overpametrization.}, which can make the process singular in sub-manifolds of $\R^d$. A regularization parameter $\eps > 0 $ is then introduced, allowing us to write: 
\begin{equation}
    \label{eq:app_dWdistr}
    \Delta \sv_k  \equiv  \sv_{k+1}  -  \sv_k  \sim \N\left( \fk \Delta \time, \gamma\Delta \time \left(  \Gk \GkT + \eps\ID_d  \right)\right) \;,
\end{equation} 
with $\fk\equiv\f(k,\sv_k)$ and $\Gk\equiv \G (k,\sv_k)$.

Let $\ProbFI$ be the probability of the process to be at $\sv$ at time $t$ given it starts from $\svI$ at time $\tI$. And $\Prop$ the one-step propagator associated with Eq.~\eqref{eq:app_sdege}. Since the process is Markovian, the Chapman–Kolmogorov equation holds over the intermediate time windows $[t_k, t_{k+1}]$:
\begin{equation}
    \label{eq:app_chapkolm}
    \ProbFI =  \lim_{\epsilon\to 0}\lim_{N\to\infty} \int \prod_{k'=1}^{N-1} \dd \sv_{k'} \prod_{k=0}^{N-1} \Prop  \;.
\end{equation}
The transition probability over each intermediate time window $[\time_k, \time_{k+1}]$ is known from Eq.\eqref{eq:app_dWdistr}. Replacing in Eq.\eqref{eq:app_chapkolm}, we have:
\begin{equation}
    \ProbFI  =     \lim_{\epsilon \to 0}\lim_{N\to\infty} 
    \int \prod_{k'=1}^{N-1} \frac{\dd \sv_{k'}}{\left( 2 \pi \gamma \Delta \time  \right)^{\nicefrac{d}{2}} \left(  \det \Omke \right)^{\nicefrac{1}{2}}}      \exp \left[ -  \frac{\Delta \time}{2 \gamma } \sum_{k=0}^{N-1} \left( \dsvdtD -  \fk  \right)^\top
 \left(  \Omke \right)^{-1}  \left(\dsvdtD  - \fk \right)   \right]   \;,
\end{equation}
where $\Omke \equiv \Gk   \Gk^\top  + \eps \ID_d$ is the regularized covariance.
 
In the continuous-time limit, the expression above becomes a {\it path integral} over all possible trajectories $\sv(\time)\in \R^d$ for the process to be in $\sv$ at time $t$ given the process starts in $\svI$ at time $\tI$:
\begin{equation}
    \label{eq:app_pathint}
    \ProbFI = \lim_{\eps \to 0 }   \int^{\sv(t) = \sv}_{\sv(\tI) = \svI} \Dsvt  \exp\left( - \frac{1}{\gamma } \Svte \right)  \;.
\end{equation}
The \emph{action} functional is defined as 
\begin{equation}
    \Svte \equiv  \int_{\tI}^{t} \dd \time   \;  \LagrfE  \;,
\end{equation}
where the continuous-time Lagrangian function is given by
\begin{equation}
    \LagrfE = \frac{1}{2} \left(  \dv{\sv(\time)}{\time} - \f \left(\time,\sv(\time)\right)\right)^\top  \OmkeC^{-1}  
    \left( \dv{\sv(\time)}{\time} - \f \left(\time,\sv(\time)\right)\right) 
\end{equation}
with $\OmkeC \equiv \G \left(\time,\sv(\time)\right) \G \left(\time,\sv(\time)\right)^\top + \eps \; \ID_d$.
The path measure is expressed as
\begin{equation}
    \label{eq:app_path_measure}
    \Dsvt  \equiv  \lim_{N\to\infty}  \prod_{k'=1}^{N-1} \frac{\dd \sv_{k'}}{\left( 2 \pi \gamma \Delta \time  \right)^{\nicefrac{d}{2}} \left(\det  \Omke \right)^{\nicefrac{1}{2}}}   \;,
\end{equation}
and the integrals must be understood under the Itô discretization. For example
\begin{equation}
    \int_{\tI}^{t} \dd \time  \left(  \dv{\sv(\time)}{\time} - \f \left(\time,\sv(\time)\right)\right)
    =
    \lim_{N\to\infty}   \sum_{k=0}^{N-1}  \Delta \time \left( \dsvdtD  - \f  \left(k,\sv_k \right)   \right)\;.
\end{equation}

We observe that, by construction, $ \G (\time, \sv(\time)) \G (\time, \sv(\time))^\top$ is positive semi-definite, then the pseudo-inverse $\OmkeC^{-1}$ is well-defined and also positive semi-definite. Therefore, for any point $\time$ in the time-interval $[\tI, t]$:
\begin{equation}
    \LagrfE   \ge  0 \;,
\end{equation}
which allows us to write:
\begin{equation}
    \label{eq:app_act_g0}
    \Svte   \ge   0 \;.
\end{equation}

Realizing that the action is non-negative will play an important role on identifying the high-probability paths in the limit $\gamma\to0$.

%%%%%%%%%%%%%%%%%%%
\subsection{Laplace approximation}
%%%%%%%%%%%%%%%%%%%
If $\gamma$ is small enough, the path integral in Eq.~\eqref{eq:app_pathint} is dominated by the paths minimizing the action $\Svte$. This is equivalent to the variational statement of classical mechanics known as Hamilton's action principle~\cite{fetter_2023}, which requires the vanishing of the first-order variation of the action functional,
\begin{equation}
  \delta  \Svte = \delta \int_{\tI}^{t} \dd\time   \;  \LagrfE \;,
\end{equation}
satisfying the {\it two boundary conditions} $\sv(\tI)= \sv_0$, $\sv(t)=\sv$. The path is a stationary point of the action if and only if it satisfies the Euler-Lagrange equations
\begin{equation}
    \label{eq:aa_EL}
   \left( \dv{\time} \pdv{\dot{\svscalar}_j} - \pdv{\svscalar_j} \right) \LagrfE  = 0 \;,
\end{equation}
for $j=1,\dots,d$.

Writing $\LagrfE$ explicitly as sum over the components, 
\begin{equation}
    \LagrfE = \frac{1}{2} \sum_{l,m=1}^{d} \left( \dv{\svscalar_l (\time)}{\time}  - f_l (\time,\sv(\time)) \right) \left(\OmkeC^{-1}\right)_{lm}  \left( \dv{\svscalar_m (\time)}{\time} - f_m (\time,\sv(\time)) \right) \;.
\end{equation}
and the Jacobian matrix of $\f$ along the trajectory $\sv(\time)$:
\begin{equation}
    \label{eq:app_jacdef}
    \Jacf (\time,\sv(\time))   \equiv \begin{bmatrix}
        \grad f_1 (\time,\sv(\time))^\top \\
        \vdots \\
        \grad f_d (\time,\x(\time))^\top  
    \end{bmatrix} = 
        \begin{bmatrix}
            \pdv{f_1 (\time,\sv(\time))}{\svscalar_1} &  \pdv{f_1 (\time,\sv(\time))}{\svscalar_2} & \dots  &  \pdv{f_1 (\time,\sv(\time))}{\svscalar_d}\\
        \pdv{f_2 (\time,\sv(\time))}{\svscalar_1} & \pdv{f_2 (\time,\sv(\time))}{\svscalar_2} & \dots  &  \pdv{f_2 (\time,\sv(\time))}{\svscalar_d}\\
        \vdots & \vdots & \ddots & \vdots \\
        \pdv{f_d (\time,\sv(\time))}{\svscalar_1} &  \pdv{f_d (t,\sv(\time))}{\svscalar_2} & \dots  &  \pdv{f_d (\time,\sv(\time))}{\svscalar_d} 
    \end{bmatrix} \in \R^{d\times d} \;,
    \end{equation}
the derivatives of the Lagrangian with respect to the components of $\sv$ and $\dot{\sv}$ are given by
\begin{subequations}
    \begin{equation}
        \label{eq:app_EL1}
        \pdv{\dot{\svscalar}_j} \LagrfE = \sum_{l=1}^{d} \left( \dv{\svscalar_l (\time)}{\time}  - f_l (\time,\sv(\time)) \right) \left(\OmkeC^{-1}\right)_{lj} \;,
    \end{equation}
\begin{equation}
    \begin{aligned}
        \pdv{\svscalar_j} \LagrfE = & -  \sum_{l,m=1}^{d} \left( \dv{\svscalar_l (\time)}{\time}  - f_l (\time,\sv(\time)) \right) \left(\OmkeC^{-1}\right)_{lm} \left(\Jacf (\time,\sv(\time)) 
        \right)_{mj}  \\
                &+ \frac{1}{2} \sum_{l,m=1}^{d}  \left( \dv{\svscalar_l (\time)}{\time}  - f_l (\time,\sv(\time)) \right) \left( \pdv{\svscalar_j}  \left(\OmkeC^{-1}\right)_{lm} \right)\left( \dv{\svscalar_m (\time)}{\time}  - f_m (\time,\sv(\time)) \right)   \;,
    \end{aligned}    
\end{equation}
\end{subequations}
where we have used the fact that $\OmkeC$ is symmetric. 

Reverting to matrix notation, the set of Euler-Lagrange equations~\eqref{eq:aa_EL} can then be represented by:
\begin{equation}
    \label{eq:app_EL2}
        \begin{aligned}
       & \dv{\time} \left[ \left( \dv{\sv (\time)}{\time}  - \f (\time,\sv(\time)) \right)^\top \OmkeC^{-1}   \right]  \\
       &  +  \left( \dv{\sv(\time)}{\time}  - f (\time,\sv(\time)) \right)^\top \OmkeC^{-1} \Jacf (\time,\sv(\time))  \\
       &   - \frac{1}{2} \left( \dv{\sv(\time)}{\time}  - \f (\time,\sv(\time)) \right)^\top      \pdv{\OmkeC^{-1}}{\sv} \left( \dv{\sv(\time)}{\time}  - \f (\time,\sv(\time)) \right)^\top  = \zero  \;,
        \end{aligned}        
    \end{equation}
where  $\pdv{\OmkeC^{-1}}{\sv} $ is defined as the tensor with components
\begin{equation}
    \left( \pdv{\OmkeC^{-1}}{\sv} \right)_{j}^{lm} \equiv \pdv{\svscalar_j} \left(\OmkeC^{-1}\right)_{lm} \;.
\end{equation}
Similarly to the covariant formulation of classical electromagnetism and special relativity, we assumed that upper indices are the ones to be contracted on vector-tensor multiplication.  

%%%%%%%%%%%%%%%%%%%
\subsection{Dominant paths}\label{app:highprob}
%%%%%%%%%%%%%%%%%%%
Eqs.~\eqref{eq:app_EL2} constitute an intricate set of second-order differential equations. Nevertheless, we recall that $\svode (\time)$ is assumed to be the unique global solution of the initial value problem:
\begin{equation}
    \label{eq:app_ivp}
    \dv{\sv(\time)}{\time}   = \f (\time, \sv(\time))  \; \; \; \; , \;\;\;\; \svode(\tI)=\sv_0     \;.
\end{equation}
If the terminal condition $\svode(t)=\sv$ is also satisfied, then $\svode (\cdot)$ is also a solution of the Euler-Lagrange equations for all $\time$ in the interval $[\tI, t]$. Moreover, the action would vanish for this solution:
\begin{equation}
        \Svteode =  \int_{\tI}^{t} \dd \time   \;  \LagrfEode = 0 \;.
\end{equation} 

From Eq.~\eqref{eq:app_act_g0}, we know that $\Svte$ is non-negative. Thus, for any solution of the initial value problem~\eqref{eq:app_ivp} with terminal condition $\svode(t) \ne \sv$, the action is greater than zero. Since the $\ProbFI \propto e^{-\frac{1}{\gamma}\Svte}$, paths $\svode (\time)$ not satisfying both initial  $\svode(\tI)=\sv_0$ and terminal conditions  $\svode(t)=\sv$ are exponentially less probable in the limit $\gamma\to 0$. Henceforth, we will refer to the paths $\svode(\time)$ satisfying both initial and terminal conditions,
\begin{equation}
    \label{eq:app_boundary2}
      \begin{cases}
        \svode(\tI) & = \sv_0 \;, \\
        \svode(t) & = \sv \;,
        \end{cases}
\end{equation}
as the {\it dominant paths}.

%%%%%%%%%%%%%%%%%%%
\subsection{Fluctuations around dominant paths}
%%%%%%%%%%%%%%%%%%%
Consider now a small time-dependent stochastic perturbation $\z(\time)$ around a dominant path such that the trajectory $\bar{\sv}(\time)$ satisfying Eq.~\eqref{eq:app_sdege} can be written as
\begin{equation}
\label{eq:app_fluc}
    \bar{\sv}(\time) = \svode (\time) +  \sqrt{\gamma} \z (\time)  \;,
\end{equation}
where $\svode (\time)$ is the high-probability path satisfying the initial and terminal conditions $\svode(\tI)=\sv_0$ and $\svode(t)=\sv$, respectively. The random perturbation $\z(\time)$ is constrained to satisfy the boundary conditions $\z(\tI) = \z(t\to\infty) = \zero$. For intermediate times, there is no constraint on the fluctuations.

In order to study the fluctuations' effects on the probability measure, the key quantity is the covariance matrix: 
\begin{equation}
    \label{eq:covz}
    \C(\time)  \equiv \E \left[ \sv(\time) - \E \left[ \sv(\time) \right]   \right] \E\left[ \sv(\time) - \E\left[ \sv(\time) \right] \right]^\top  \;,
\end{equation}
where $\E[\cdot]$ stands for the expectation over all sources of randomness. For $\sv(\time) = \bar{\sv} (\time)$, the only stochasticity is the perturbation and the covariance reduces to
\begin{equation}
    \label{eq:covz2}
     \C (\time) =  \gamma\;  \left( \E_{\z}\left[ \z(\time) \z(\time)^\top \right] - \E_{\z}\left[ \z(\time) \right] \E_{\z}\left[ \z(\time) \right]^\top  \right)    \;,
\end{equation}
where $\E_{\z}[\cdot]$ stands for the expectation over the perturbation $\z(\cdot)$. Up to now, the probability distribution for $\z$ is unknown. Assuming $\gamma\to0$, we will construct a measure for $\z$ in order to compute the covariance. We first Taylor expand the elements of $ \f \left(\time,\sv(\time)\right) $ and $  \OmkeC^{-1}  $ around $\svode (\time)$:
\begin{subequations}
    \label{eq:expTaylor}
\begin{align}
 f_l \left( \time ,  \bar{\sv}(\time) \right)   &= f_l ( \time,\svode (\time) ) + \sqrt{\gamma} \; \z (\time)^\top  \grad f_l ( \time,  \svode (\time) ) + \O (\gamma)   \;, \\ 
   \left( \Omeps \left( \time ,\bar{\sv}(\time)  \right)^{-1}\right)_{lm}   &= \left( \Omeps \left( \time , \svode (\time)\right)^{-1}\right)_{lm}    +  \sqrt{\gamma} \; \z (\time)^\top  \grad \left( \Omeps \left( \time , \svode (\time)\right)^{-1}\right)_{lm} + \O (\gamma)  \;.
  \end{align}
\end{subequations}
for $l,m=1,\dots,d$. Additionally, since by assumption $\svode(\time)$ is the solution of $\dv{\sv(\time)}{\time} = \f (\time, \sv(\time))$ satisfying the boundary conditions~\eqref{eq:app_boundary2}: 
\begin{equation}                                                
    \dv{\bar{\svscalar}_l (\time)}{\time} - f_l (\time,\sv(\time)) \approx \dv{z_l (\time)}{\time} - \z(\time)^\top \grad f_l (\time,\svode(\time))  \;.
\end{equation}
Plugging it into the Lagrangian together with Eqs.~\eqref{eq:expTaylor} and retaining contributions up to order two on $\z(\cdot)$, we obtain:
\begin{equation}
    \label{eq:app_lagr}
\begin{aligned}
    \Lagr^{\eps} & \left(\time,\bar{\sv}(\time), \dot{\bar{\sv}}(\time)\right) = \\
    & =  \frac{\gamma}{2} \sum_{l,m=1}^d \left( \dv{z_l(\time)}{\time} - \z(\time)^\top \grad f_l (\time,\svode(\time))\right) 
    \left( \Omeps \left( \time , \svode (\time)\right)^{-1}\right)_{lm}
    \left( \dv{z_m (\time)}{\time} - \z(\time)^\top \grad f_m (\time,\svode(\time))\right) \\
            &= \frac{\gamma}{2} \left( \dv{\z(\time)}{\time} - \Jacf(\time,\svode(\time)) \z(\time) \right)^\top
            \Omeps \left( \time , \svode (\time)\right)^{-1}
            \left( \dv{\z (t)}{t} - \Jacf(t,\svode(t)) \z(t) \right) \;,
\end{aligned}
\end{equation}
where in the second line we have used the Jacobian matrix defined in~\eqref{eq:app_jacdef}. Therefore, the measure governing the fluctuations $\z(t)$ in the limit $\gamma\to0$ is given by 
\begin{equation}
    \bar{\Prob} \left( \z , t  | \zero, \tI \right)   \propto \lim_{\eps \to 0 }  \int^{\z(t) = \z}_{\z(\tI) = \zero} \Dzt  \exp\left( -  \int_{\tI}^{t} \dd \time    \;  \bar{\Lagr}^{\eps}\left(\time,\z (\time), \dot{\z}(\time)\right)  \right) \;,
\end{equation}
where 
\begin{equation}
    \label{eq:app_Lz}
    \bar{\Lagr}^{\eps}\left(\time,\z (\time), \dot{\z}(\time)\right)  
    \equiv  \frac{1}{2} \left( \dv{\z (\time)}{\time} - \Jacf(\time,\svode(\time)) \z(\time) \right)^\top
    \Omeps\left(\time,\svode(\time)\right)^{-1}
    \left( \dv{\z (t)}{t} - \Jacf(t,\svode(t)) \z(t) \right)  \;.
\end{equation}

The diffusion matrix appearing on the path measure,
\begin{equation}
    \Dzt  \equiv                          \lim_{N\to\infty}  \prod_{k'=1}^{N-1} \frac{\Delta \z_{k'}}{\left( 2 \pi \Delta \time  \right)^{\nicefrac{d}{2}} \left( \det \Om_{k}^{\eps\text{,ode}} \right)^{\nicefrac{1}{2}}}   \;,
\end{equation}
is now calculated on the discretized dominant trajectory: $ \Om_{k}^{\eps\text{,ode}} = \G (k,\sv_k^\text{ode}) \G (k,\sv_k^\text{ode})^\top  + \eps \; \ID_d $.

For the purpose of simplifying the Lagrangian~\eqref{eq:app_Lz}, we perform the change of variables:
\begin{equation}
    \label{eq:app_ztil}
       \z(\time)  =   \T \left\{e^{\int_{\tI}^{\time} \dd s \; \Jacf(s,\svode(s))} \right\}   \Tilde{\z}(\time) \;,
\end{equation}
where the object inside the brackets is a matrix in $\R^{d\times d}$. The exponential of a matrix is, as usual, defined via Taylor series. The symbol $\T$ indicates time-ordering in the sense that the exponential is understood as a Dyson series~\cite{weinberg_1995}. For the convenience of the reader, we review the time-ordered exponential for time-dependent non-commutative matrices in Appendix~\ref{app:time_ord}.

Under the change of variables, the time derivative of $\z (t)$ is written as 
\begin{equation}
\begin{aligned}
 \dv{\z(\time)}{\time} &= \Jacf(\time,\svode(\time)) \;  \T \left\{   e^{\int_{\tI}^{\time} \dd s \; \Jacf(s,\svode(s))} \right\}  \Tilde{\z}(\time)  +  \T  \left\{  e^{\int_{\tI}^{\time} \dd s \; \Jacf(s,\svode(s))}   \right\}  \dv{\Tilde{\z}(\time)}{\time}   \\ 
  &= \Jacf(\time,\svode(\time)) \z(\time) +   \T  \left\{  e^{\int_{\tI}^{\time} \dd s \; \Jacf(s,\svode(s))}  \right\}  \dv{\Tilde{\z}(\time)}{\time}   \;,
\end{aligned}
\end{equation}
so that the Lagrangian~\eqref{eq:app_Lz}, in terms of $ \Tilde{\z}(\time)$, is written as
\begin{subequations}
    \begin{equation}
        \bar{\Lagr}^{\eps} \left(\time,\Tilde{\z} (\time), \dot{\Tilde{\z}}(\time) \right)  
        \equiv  \frac{1}{2}  \dv{\Tilde{\z}(\time)}{\time}^\top \;  
        \PsiB^\eps (\time,\svode(\time)) \dv{\Tilde{\z}(\time)}{\time}   \;.
    \end{equation}
with
\begin{equation}
    \label{eq:app_psi}
    \PsiB^\eps(\time,\svode(\time))  \equiv  \T  \left\{  e^{\int_{\tI}^{\time} \dd s \; \Jacf(s,\svode(s))}  \right\}^\top  \Omeps\left(\time,\svode(\time)\right)^{-1} \; \T  \left\{  e^{\int_{\tI}^{\time} \dd s \; \Jacf(s,\svode(s))}  \right\} \;.
\end{equation}
\end{subequations}

Let us mention in passing an analogy that can be made with techniques from quantum mechanics at this point. One could interpret~\eqref{eq:app_psi} as a kind of transformation from a mix of Schrödinger and Heisenberg pictures to the interaction picture, with $ \Jacf(s,\svode(s))$ and $\PsiB^\eps ( \time,  \svode (\time)   )$ playing the role of time-dependent Hamiltonian and potential, respectively~\cite{sakurai_2020}. 

With the intention of computing the expectations in~\eqref{eq:covz2}, which are over the original coordinates $\z(\cdot)$, we consider the quantities $ \E_{\Tilde{\z}} \left[ \Tilde{\z}(\cdot)  \right]$ and $\E_{\Tilde{\z}} \left[ \Tilde{\z}(\cdot) \Tilde{\z}(\cdot)^\top \right]$, where $\E_{\Tilde{\z}}$ indicates expectation over the measure $ \propto e^{- \frac{1}{2 } \int_{\tI}^{t} \dd\time \dv{\Tilde{\z}(\time)}{\time}^\top \;  \PsiB^\eps (\time,\svode(\time)) \dv{\Tilde{\z}(\time)}{\time}  }$. At this point, it is convenient to return to the discretized form and to define a new discrete variable:
\begin{equation}
    \Tilde{\z}_k = \sum_{k' = 0}^{k-1} \left(\Tilde{\z}_{k' +1} - \Tilde{\z}_{k'} \right)  \equiv \sum_{k' = 0}^{k-1}  \bm{u}_{k'} \;,
\end{equation}
as the assumption $ \Tilde{\z} (\tI) = \zero $ implies that $\Tilde{\z} (\tI) =  \Tilde{\z}_{0} = \zero$. Similarly, the discretized derivative in terms of the variable $\bm{u}_k$ is given by:
\begin{equation}
    \frac{\Delta \Tilde{\z}_k}{\Delta\time} = \frac{\Tilde{\z}_{k+1} - \Tilde{\z}_k}{\Delta\time} = \frac{\bm{u}_k}{\Delta\time}  \;.
\end{equation}

The discretized measure then reads:
\begin{equation}
    \begin{aligned}
    \exp\left[ - \frac{1}{2} \int_{\tI}^{t} \dd\time   \dv{\Tilde{\z}(\time)}{\time}^\top \;  \PsiB^\eps(\time,\svode(\time))    \dv{\Tilde{\z}(\time)}{\time}  \right] &\to \exp\left[ - \frac{1}{2 } \sum_{k=0}^{N-1} \Delta\time \; \frac{\Delta \Tilde{\z}_k}{\Delta\time}^\top    \PsiB^\eps ( k,  \svode_k   )  \frac{\Delta \Tilde{\z}_k}{\Delta\time}  \right] \\ 
      & \; =   \exp\left[ - \frac{1}{2} \sum_{k=0}^{N-1} \frac{\Delta\time}{(\Delta\time)^2}  \bm{u}_k^\top   \PsiB^\eps ( k,\svode_k   ) \bm{u}_k \right] \\
       &\; = \prod_{k=0}^{N-1}  \exp\left[ - \frac{1}{2 \Delta\time}   \bm{u}_k^\top  \PsiB^\eps(k,\svode_k) \bm{u}_k \right] \;,
    \end{aligned}
\end{equation}
where $\PsiB^\eps (k,\svode_k )$ means the discretization of $\PsiB^\eps (\time,  \svode (\time)   )$ on the discretized dominant path. We then conclude that the first and second moments of $\bm{u}_k$ are given by:
\begin{subequations}
    \begin{align}
    \E \left[ \bm{u}_k \right]  &= \zero \;,   \label{eq:fmom} \\ 
    \E \left[ \bm{u}_k \bm{u}_k^\top \right]  &=  \Delta\time \; \PsiB^\eps (k,\svode_k )^{-1}  \;,
    \end{align}
\end{subequations}
respectively. From Eq.~\eqref{eq:fmom}, we immediately conclude that $\E \left[ \Tilde{\z}(\time) \right] = \E \left[ \z (\time) \right]   = \zero$. With respect to the second moment, we observe that:
\begin{equation}
\begin{aligned}
    \E \left[ \Tilde{\z}_k \Tilde{\z}_{k}^\top  \right]    &=  \sum_{k' , k''  = 0}^{k-1} \E \left[ \bm{u}_{k'} \bm{u}_{k''}^\top  \right]  = \sum_{k'  = 0}^{k-1}  \E \left[ \bm{u}_{k'} \bm{u}_{k'}^\top  \right]  \\ 
  &=  \sum_{k'  = 0}^{k-1}  \Delta\time   \; \PsiB^\eps (k,\svode_k )^{-1}   \;,
\end{aligned}
\end{equation}
where we have used the Markovian property. In the continuous-time limit:
\begin{equation}
    \E \left[  \Tilde{\z}(t) \Tilde{\z}(t)^\top \right] = \int_{\tI}^{t} \dd\time \; \PsiB^\eps(\time,\svode(\time))^{-1}  \;.
\end{equation}
Returning to the original coordinates:
\begin{equation}
    \begin{aligned}
   \E\left[ \z(t) \z(t)^\top \right]   &= \T \left\{   e^{\int_{\tI}^{t} \dd s \;\Jacf(s,\svode(s))}   \right\} \E\left[ \Tilde{\z}(t) \Tilde{\z}(t)^\top\right]    \T \left\{   e^{\int_{\tI}^{t} \dd s \; \Jacf(s,\svode(s))}  \right\}^\top   =  \\ 
      &=  \T\left\{e^{\int_{\tI}^{t} \dd s \; \Jacf(s,\svode(s))}\right\} \int_{\tI}^{t} \dd \time \;  \PsiB^\eps(\time,\svode(\time))^{-1} \; \T \left\{   e^{\int_{\tI}^{t} \dd s \;\Jacf(s,\svode(s))}  \right\}^\top \;.
    \end{aligned}
    \end{equation}
Hitherto, the covariance matrix can then be expressed as 
\begin{equation}
    \label{eq:app_covH}
    \C (t)  = \gamma \;  \lim_{\eps\to 0} \; \T\left\{e^{\int_{\tI}^{t} \dd s \; \Jacf(s,\svode(s))}\right\} \int_{\tI}^{t} \dd \time \;  \PsiB^\eps(\time,\svode(\time))^{-1} \; \T \left\{   e^{\int_{\tI}^{t} \dd s \;\Jacf(s,\svode(s))^\top   }  \right\}  \;,
\end{equation}
with
\begin{equation}
    \PsiB^\eps(\time,\svode(\time))  = \T  \left\{  e^{\int_{\tI}^{\time} \dd s \; \Jacf(s,\svode(s))^\top}  \right\}  \Omeps\left(\time,\svode(\time)\right)^{-1} \; \T  \left\{  e^{\int_{\tI}^{\time} \dd s \; \Jacf(s,\svode(s))}  \right\}  \;,
\end{equation}
and
\begin{equation}
    \Om^\eps ( \time,  \svode (\time)   ) = \G \left(\time,\svode(\time)\right) \G \left(\time,\svode(\time)\right)^\top + \eps \; \ID_d  \;.
\end{equation}
We continue by explicitly inverting $\PsiB^\eps$,
\begin{equation}
    \begin{aligned}
        \PsiB^\eps & (\time,\svode(\time))^{-1}  =  \\
        &=  \left( \T  \left\{  e^{\int_{\tI}^{\time} \dd s \; \Jacf(s,\svode(s))}  \right\}  \right)^{-1} \Omeps\left(\time,\svode(\time)\right) \; \left(  \T  \left\{  e^{\int_{\tI}^{\time} \dd s \; \Jacf(s,\svode(s))^\top}  \right\}  \right)^{-1} \\
        &=  \left( \T  \left\{  e^{\int_{\tI}^{\time} \dd s \; \Jacf(s,\svode(s))}  \right\}  \right)^{-1} \left(   \G \left(\time,\svode(\time)\right) \G \left(\time,\svode(\time)\right)^\top + \eps \; \ID_d    \right) \; \left(  \T  \left\{  e^{\int_{\tI}^{\time} \dd s \; \Jacf(s,\svode(s))^\top}  \right\}  \right)^{-1}
    \end{aligned}
\end{equation}
and replacing it in the expression~\eqref{eq:app_covH} for the covariance matrix, which leads to:
\begin{equation}
    \begin{aligned}
        \C (t) &= \gamma \; \T\left\{e^{\int_{\tI}^{t} \dd s \; \Jacf(s,\svode(s))}\right\} \int_{\tI}^{t} \dd \time \;  \left( \T  \left\{  e^{\int_{\tI}^{\time} \dd s \; \Jacf(s,\svode(s))}  \right\}  \right)^{-1}   \G \left(\time,\svode(\time)\right) \G \left(\time,\svode(\time)\right)^\top  \cdot \\
               & \cdot \left(  \T  \left\{  e^{\int_{\tI}^{\time} \dd s \; \Jacf(s,\svode(s))^\top}  \right\}  \right)^{-1} \; \T \left\{   e^{\int_{\tI}^{t} \dd s \;\Jacf(s,\svode(s))^\top}  \right\}     
    \end{aligned}
\end{equation}
where we have already computed the limit $\eps\to0$. Since, for any $\time$ such that $\tI < \time < t $:
\begin{equation}
    \T\left\{e^{\int_{\tI}^{t} \dd s \; \Jacf(s,\svode(s))}\right\} =  \T\left\{e^{\int_{\time}^{t} \dd s \; \Jacf(s,\svode(s))}\right\} \cdot  \T\left\{e^{\int_{\tI}^{\time} \dd s \; \Jacf(s,\svode(s))}\right\}  \;,
\end{equation}
we have the final expression for the covariance matrix:
\begin{equation}
    \C (t) = \gamma   \int_{\tI}^{t} \dd \time \;   \T\left\{e^{\int_{\time}^{t} \dd s \; \Jacf(s,\svode(s))}\right\} 
    \G \left(\time,\svode(\time)\right) \G \left(\time,\svode(\time)\right)^\top  \T\left\{e^{\int_{\time}^{t} \dd s \; \Jacf(s,\svode(s))^\top}\right\}  \;.
\end{equation}
Setting $ \U^\text{ode} (t, \time)  \equiv  \T\left\{ e^{\int^t_\time \dd s \Jacf (s, \svode(s)) } \right\}$ and observing that $\Om^0(\time,\svode(\time)) = \G \left(\time,\svode(\time)\right) \G \left(\time,\svode(\time)\right)^\top $, we obtain:
\begin{equation}
    \label{eq:app_general-final-cov}
    \C (t) = \gamma   \int_{\tI}^{t} \dd \time \;  \U^\text{ode} (t, \time)  \;
    \Om^0(\time,\svode(\time)) \; \U^\text{ode} (t, \time)^\top  \;,
\end{equation}
which is exactly the same expression as in Eqs.~\eqref{general-final-cov} presented in the main text.

%%%%%%%%%%%%%%%%%%%
\subsection{Special cases}\label{sec:app1_ml0}
%%%%%%%%%%%%%%%%%%%
%%%%%%%%%%%%%%%%%%%
\subsubsection{Conservative drifts}
%%%%%%%%%%%%%%%%%%%
If there exists a scalar function $\Phi(\time,\sv(\time))$ such that the drift $\f(\time,\sv(\time)) \in \R^d $ is given by:
\begin{equation}
    \f(\time,\sv(\time)) = - \grad \Phi(\time,\sv(\time)) \;,
\end{equation}
we say that the drift is \emph{conservative}, in analogy with the conservative forces in classical mechanics~\cite{fetter_2023}. In this case, the Jacobian matrix of $\f$ is nothing but the negative of the Hessian of $\Phi$: 
\begin{equation}
    \Jacf (\time,\sv(\time)) = -  \Hess_{\Phi} (\time,\sv(\time)) \;,
\end{equation}
where
\begin{equation}
    \Hess_{\Phi} (\time,\sv(\time))   \equiv  \grad \otimes \grad\Phi(\time,\sv(t)) =      \begin{bmatrix}
        \pdv{\Phi(\time,\sv(\time))}{\svscalar_1^2} & \pdv{\Phi(\time,\sv(t))}{\svscalar_1 \partial \svscalar_2} & \hdots &
        \pdv{\Phi(\time,\sv(\time))}{\svscalar_1 \partial \svscalar_d} 
        \\
        \vdots & \vdots & \ddots & \vdots \\
        \pdv{\Phi(\time,\sv(\time))}{\svscalar_d \partial \svscalar_1} &
        \pdv{\Phi(\time,\sv(\time))}{\svscalar_d \partial \svscalar_2} &
        \hdots &
        \pdv{\Phi(\time,\sv(\time))}{\svscalar_d^2  }
    \end{bmatrix} \in \R^{d \times d} \;.
 \end{equation}

%%%%%%%%%%%%%%%%%%%
\subsubsection{Cholesky-like diffusions}
%%%%%%%%%%%%%%%%%%%
If the diffusion matrix $\G(\time,\sv(\time)) \in \R^{d\times n} $ is such that there exists a matrix $  \bm{\Xi} (\time,\sv(\time))   \in \R^{d\times d}$ given by
\begin{equation}
    \bm{\Xi} (\time,\sv(\time))   =  \G(\time,\sv(\time)) \G(\time,\sv(\time))^\top \;,
\end{equation}
we say that $\G(\time,\sv(\time))$ is a Cholesky-like diffusion matrix. In this case, we use the convention $ \G(\time,\sv(\time))  \equiv \sqrt{\bm{\Xi} (\time,\sv(\time))}$.

%%%%%%%%%%%%%%%%%%%
\subsection{Machine learning realm}\label{sec:app1_ml}
%%%%%%%%%%%%%%%%%%%
We consider now the particular case of the stochastic gradient flow modeling of SGD:
\begin{equation}
    %\label{eq:sdeg}
    \dd \bbh(t) = - \grad \lossL\bigl(\bbh(t) ;\X,\y\bigl) \dd t  
    + \sqrt{ \lr \;  \bm{\Sigma} \bigl(\bbh(t) ;\X,\y\bigl) } \; \dd \brow(t) \;,
\end{equation}
which is Eq.~\eqref{eq:sdeg} in the main text (where all the quantities are defined). Clearly, the drift is conservative with the Jacobian equal to minus the Hessian of the loss. The diffusion matrix is also Cholesky-like. Additionally, if $\lr=0$, the SDE is reduced to the gradient flow equation, and this is why we term the solution of the ODE in this context as $\bbh^\text{GF}(\time)$.

In summary, for SGF the covariance~\eqref{eq:app_general-final-cov} reduces to
\begin{subequations}
    \begin{equation}
        \C^\text{SGF} (t) = \gamma   \int_{\tI}^{t} \dd \time \;  \U^\text{GF} (t, \time)  \;
        \bm{\Sigma} \bigl(\bbh^\text{GF}(\time) ;\X,\y\bigl)\; \U^\text{GF} (t, \time)^\top  \;, 
    \end{equation}
with
    \begin{equation}
        \U^\text{GF} (t, \time)  \equiv  \T\left\{ e^{ - \int^t_\time \dd s \HessL (\bbh^\text{GF}(s) ;\X,\y) } \right\} \;,
    \end{equation}
\end{subequations}
which are equivalent to Eqs.~\eqref{cov-ML} in the main text.

%%%%%%%%%%%%%%%%%%%
\subsection{Discrete to continuous-time approximation guarantees}
%%%%%%%%%%%%%%%%%%%
If the drift is conservative and the diffusion Cholesky-like, we have:
\begin{equation}
        \dd \sv (\time) =- \grad \Phi(\time,\sv(\time))   \dd \time + \sqrt{\gamma \; \bm{\Xi} (\time,\sv(\time))} \;   \dd \svbrow(\time) \;.   
\end{equation}

From Corollary 10 in~\cite{li_2019}:
\begin{equation}
  \max_{k=0,\dots,N}   \abs{\E \left[ h\left(\sv_k\right) \right]  - \E \left[ h\left(\sv(k \Delta \time)\right) \right]  } \le C \gamma   \;,
\end{equation}
for a suitable set of `smooth' functions $h$ with `polynomial growth'.

Let us also mention that one may look at an `improved' SGF modeling:
\begin{equation}
    \dd \sv (\time) =- \grad \left( \Phi(\time,\sv(\time))  + \frac{1}{4}\gamma \norm{\grad \Phi(\time,\sv(\time)) }^2   \right)  \dd \time + \sqrt{\gamma \; \bm{\Xi} (\time,\sv(\time))} \;   \dd \svbrow(\time) \;,   
\end{equation}
where the convergence is improved to (Theorem 9 in~\cite{li_2019}):
\begin{equation}
    \max_{k=0,\dots,N}   \abs{\E \left[ h\left(\sv_k\right) \right]  - \E \left[ h\left(\sv(k \Delta \time)\right) \right]  } \le C \gamma^2   \;,
  \end{equation}
again for a suitable set of `smooth' functions $h$ with `polynomial growth'.
\section{Time-ordered exponential}
\label{app:time_ord}

The concept of time-ordering integrals was first introduced by Dyson~\cite{dyson_1949} in the context of time-dependent perturbation theory applied to scattering problems in quantum mechanics. Let $\bm{B}(t)$ and $\bm{A}(t) $ be two time-dependent square real matrices of the same dimension. Consider the following first-order differential equation:
\begin{equation}
\label{eq:dy_eq}
    \dv{\bm{B}(t)}{t}   = \bm{A} (t)  \bm{B} (t)  \; 
\end{equation}
with initial condition $\bm{B}(t_0)= \ID $. In quantum mechanics, both matrices could be complex with $\bm{B}(t)$ representing the time-evolution operator and $\bm{A}(t)$ a potential in the interaction picture. 

Observe that Eq.~\eqref{eq:dy_eq} (with the initial condition) can be reformulated as an integral equation:
\begin{equation}
    \bm{B} (t) = \ID - \int_{t_0}^{t} \dd \tau \;  \bm{A} (\tau) \bm{B}(\tau) \;.
\end{equation}
Iterating the expression above:
\begin{equation}
\label{eq:dys_exp}
    \bm{B} (t) = \ID - \int_{t_0}^t \dd\tau_1 \; \bm{A} (\tau_1) +
   \int_{t_0}^t  \dd\tau_1 \int_{t_0}^{\tau_1}  \dd\tau_2 \;  \bm{A} (\tau_1) \bm{A} (\tau_2) - \int_{t_0}^t \dd\tau_1 \int_{t_0}^{\tau_1} \dd\tau_2 \int_{t_0}^{\tau_2}  \dd\tau_3 \;   \bm{A} (\tau_1) \bm{A} (\tau_3) \bm{A} (\tau_3) + \dots   \;.
\end{equation}

The \emph{time-ordered product} $\T \{ \cdot \}$ of any time-dependent matrix is defined as the product with factors arranged in such a way that the factor latest in time takes the leftmost position, the next-latest next to the leftmost, and so on. For example:
\begin{subequations}
\begin{align}
\T \{ \bm{A}(\tau) \}   & = \bm{A}(\tau) \;,   \\ 
\T \{ \bm{A}(\tau_1) \bm{A}(\tau_2) \} &  = \Theta(\tau_1 - \tau_2) \bm{A}(\tau_1) \bm{A}(\tau_2) + \Theta(\tau_2 - \tau_1)  \bm{A}(\tau_2) \bm{A}(\tau_1)      \;,
  \end{align}
\end{subequations}
where $\Theta(\tau)$ is the Heaviside step function equal to $1$ for $\tau>0$ and to zero for $\tau<0$. This definition generalizes to the time-ordered product of $\bm{A}$'s for `$n$ time instants' as a sum over all $n!$ permutations. Each permutation gives the same integral over $\tau_1, \tau_2, \dots, \tau_n$. Eq.~\eqref{eq:dys_exp} can then be written as
\begin{equation}
\label{dys_series}
    \bm{B} (t)  =  \ID_d  
    +  \frac{(-1)^n}{n!} \sum_{n=1}^{\infty} \int_{t_0}^t \dd \tau_1
    \int_{t_0}^{\tau_1} \dd \tau_2  \ldots \int_{t_0}^{\tau_{n-1}} \dd \tau_n \; \T\left\{ \bm{A}(\tau_1) \bm{A}(\tau_2) \ldots \bm{A}(\tau_n) \right\}  \;,
\end{equation}
which in the context of quantum mechanics is referred to as the \emph{Dyson series}.

If the $\bm{A}(\tau)$'s would be commuting matrices for all $\tau$, one could simply represent the series~\eqref{dys_series} by $ e^{\int_{t_0}^{t} \dd \tau \bm{A}(\tau)}$. Nevertheless, since commutation in time is not guaranteed in general, we introduce the notation for the \emph{time-ordered exponential}:
\begin{equation}
\label{eq:exp_tm_ord}
    \T\left\{ e^{\int_{t_0}^{t} \dd \tau \bm{A}(\tau) } \right\} \equiv  \ID  
    +  \frac{(-1)^n}{n!} \sum_{n=1}^{\infty} \int_{t_0}^t \dd \tau_1
    \int_{t_0}^{\tau_1} \dd \tau_2  \ldots \int_{t_0}^{\tau_{n-1}} \dd \tau_n \; \T\left\{   \bm{A}(\tau_1) \bm{A}(\tau_2) \ldots \bm{A}(\tau_n) \right\} \;,
\end{equation}
for any square time-dependent real matrix $\bm{A}(\tau)$.

\section{Application to the weak features model: details of derivations}
\label{app:belkin}

%%%%%%%%%%%%%%%%%%%
\subsection{Preliminaries}
%%%%%%%%%%%%%%%%%%%
For arbitrary $n_1, n_2 \in \NN$, we use notation $\zero_{n_1 \times n_2} $ for all-zero $n_1 \times n_2$ matrix, $\ID_{n_1 \times n_2}  $ for $n_1 \times n_2$ matrix with elements $\left( \ID_{n_1 \times n_2} \right)_{ij} = \delta_{ij} $. $\left( \bm{A} \right)^\dagger  \in \R^{n_2 \times n_1 }$ denotes Moore-Penrose inverse of a matrix $\bm{A} \in \R^{n_1 \times n_2 }$. We use $\Unifd$ to denote uniform (Haar) measure over a group. 

We will find SVDs useful for our derivations. Suppose $r = \rank \X_{\sub} \leq \min(n, p)$.  Now, let 
\begin{subequations}
\label{eq:xsub_svd}
\begin{equation}
    \X_{\sub} = \Usvd \Lsvd \Vsvd^\top, \; \Usvd \in \R^{n \times n}, \Lsvd \in \R^{n \times p}, \Vsvd \in \R^{p \times p}
\end{equation} 
be a `full' singular value decomposition, and
\begin{equation}
    \X_{\sub} = \Usvd_r \Lsvd_r \Vsvd^\top_r, \; \Usvd_r \in \R^{n \times r}, \Lsvd_r \in \R^{r \times r}, \Vsvd_r \in \R^{p \times r}
\end{equation} 
\end{subequations}
be `truncated' the one. We also denote columns of $\Usvd$ as $\{ \usvd_i \}^n_{i = 1}$, of $\Vsvd$ as $\{ \vsvd_i \}^p_{i = 1}$ and positive singular values as $\{ \lambda_i \}^r_{i = 1}$. For convenience, we also utilize the notation $\lambda_i = 0$ for $i > r$.  Besides, we make use of $\Usvd_\rorth  := \left[  \usvd_{r+1 } \mid \dots \mid \usvd_n \right] \in \R^{n \times (n - r)} $ and  $\Vsvd_\rorth  := \left[  \vsvd_{r+1 } \mid \dots \mid \vsvd_p \right] \in \R^{p \times (p - r)} $ -- i.e., $\Usvd_\rorth$ and $\Vsvd_\rorth$ consist of columns of $\Usvd$ and $\Vsvd$ respectively which are not included in $\Usvd_r$ and $\Vsvd_r$.

Let us state some useful properties. We keep in mind that $\Usvd^\top_r \Usvd_r = \Vsvd^\top_r \Vsvd_r  =  \ID_r $, $ \Usvd^\top_\rorth \Usvd_\rorth = \ID_{n - r}  $  and $ \Vsvd^\top_\rorth \Vsvd_\rorth = \ID_{p - r}  $, and  $\Usvd^\top_r \Usvd_\rorth = \zero_{r \times (n - r)} $ and  $\Vsvd^\top_r \Vsvd_\rorth = \zero_{r \times (p - r)} $. Besides, the Moore-Penrose inverse of $\X_{\sub}$ can be written as 
\begin{equation}
    \label{eq:X_moore_penrose}
    \left( \X_{\sub} \right)^\dagger =  \Vsvd_r \Lsvd^{-1}_r \Usvd^\top_r  \;,.
\end{equation} 

Using properties of matrix exponentials, 
\begin{subequations}
    \label{eq:exp_XX_svd}
    \begin{align}
    e^{-\frac{1}{n}  \X_{\sub} \X^\top_{\sub} t } &= 
     \Usvd e^{-\frac{1}{n } \Lsvd \Lsvd^\top t } \Usvd^\top =
     \Usvd_r e^{-\frac{1}{n } \Lsvd^2_r t } \Usvd^\top_r + \Usvd_\rorth \Usvd^\top_\rorth 
     \overset{t \to \infty}{\longrightarrow} 
      \Usvd_\rorth \Usvd^\top_\rorth \;,  \\
    e^{-\frac{1}{n} \X^\top_{\sub} \X_{\sub} t} & =
     \Vsvd e^{ - \frac{1}{n} \Lsvd^\top \Lsvd t} \Vsvd^\top 
     =
     \Vsvd_r e^{-\frac{1}{n } \Lsvd^2_r t } \Vsvd^\top_r + \Vsvd_\rorth \Vsvd^\top_\rorth
    \overset{t \to \infty}{\longrightarrow}  \Vsvd_\rorth \Vsvd^\top_\rorth  \;.
     \end{align}
\end{subequations}  

In the model we look at, elements of $\X_{\sub} \in \R^{n \times p}$ are sampled independently from a standard normal distribution. This has several consequences. First, with probability 1, $\X_\sub$ is a full rank matrix, i.e. $r = \rank \X_\sub = \min(n, p)$. Second, this allows us to characterize some properties of the distribution of SVD components from Eq.~\eqref{eq:xsub_svd} --  we can consider $\Usvd, \Lsvd, \Vsvd$ to be mutually independent; in addition, 
$\Usvd$ and $\Vsvd$ are distributed uniformly on orthogonal groups of unitary matrices $\orthg{n}$ and $\orthg{p}$ of size $n$ and $p$ and respectively.  

%%%%%%%%%%%%%%%%%%%
\subsection{SDE modeling}\label{app:sde_modeling}
%%%%%%%%%%%%%%%%%%%
We apply the SDE modeling of Eq.~\eqref{eq:sdeg} to the simple regression model. To do this, we first find the drift vector 
\begin{equation}
\label{eq:belkin_drift}
    \grad \lossL\bigl(\bbh(t) ;\X,\y; \sub \bigl)= - \frac{1}{n}  \X_{\sub}^\top  \left( \y  - \X_{\sub} \betas(t) \right)
\end{equation}
and then the diffusion matrix. For the latter, we define, according to  Eq.~\eqref{eq:xi},  stochastic perturbation 

\begin{equation}
\begin{aligned}
    \xibAll   & = \grad \loss \bigl( \bbh  ; \x^{k_\i} , y^{k_\i}; \sub  \bigl) - \grad \lossL\bigl(\bbh ;\X,\y; \sub\bigl)  \\
    &= - \left( y^{k_\i}  - \betas^\top \x^{k_\i}_{\sub}   \right) \x^{k_{\i}}_{\sub}   + \frac{1}{n}  \X_{\sub}^\top  ( \y  - \X_{\sub} \betas)  \\
    & = - \left[ \left( y^{k_\i}  - \betas^\top \x^{k_\i}_{\sub}   \right) \x^{k_{\i}}_{\sub}  -  \frac{1}{n} \sum_{k=1}^{n}  \left( y^{k}  - \betas^{\top} \x^{k}_{\sub}   \right) \x^{k}_{\sub} \right]  \;.
\end{aligned}
\end{equation}

Let $\{ \e_j \}_{j=1}^n$ be a set of vectors composing the standard orthonormal basis in $\R^n$: $\e_j^\top \e_l = \delta_{jl}$. The random variable $\xib$ can then be rewritten as  
\begin{equation}
    \xib_{k_\i} \bigl(\bbh ;\X,\y; \sub \bigl)  = 
    - \X^\top_\sub \left[ 
      \left( y^{k_\i}  - \betas^{\top} \x^{k_\i}_{\sub}   \right) \e_{k_{\i}}  
       -  \frac{1}{n} \sum_{k=1}^{n}  \left( y^{k}  - \betas^{\top} \x^{k}_{\sub}   \right) \e_{k}   \right] \;.
 \end{equation}
 Additionally, since the samples are picked uniformly at random, we can replace the sum in the second term with the expectation:
 \begin{equation}
    \xibAll  = -
    \X^\top_\sub \left[ 
      \left( y^{k_\i}  - \betas^\top \x^{k_\i}_{\sub}   \right) \e_{k_{\i}}  
       - \EUnq  \left[   \left( y^{q_\i}  - \betas^\top \x^{q_\i}_{\sub}   \right) \e_{q_{\i}}    \right]  \right] \;.
 \end{equation}
 
%As mentioned before in Eq.~\eqref{eq:xi_mean_zero},  the first moment is zero:
The first moment is zero:
 \begin{equation}
   \EUn  \left[  \xibAll  \right]  = \zero \;.
 \end{equation}

 We now aim to compute the covariance matrix
 \begin{equation}
    \begin{aligned}
        \bm{\Sigma} \bigl(\bbh;\X,\y;\sub\bigl) &= 
    \EUn \left[ \xibAll \xibAll^\top \right] =   \\
    & =  \X^\top_\sub \left[
    \frac{1}{n} \sum_{k_{\i}=1}^n \left( y^{k_{\i}}  - \betas^{\top} \x^{k_{\i}}_{\sub}    \right)^2 \e_{k_{\i}} \e_{k_{\i}}^\top \;  - \frac{1}{n^2}
    \sum_{k_{\i},q_{\i}=1}^n    \left( y^{k_{\i}}  - \betas^{\top} \x^{k_{\i}}_{\sub}    \right) \left( y^{q_{\i}}  - \betas^\top \x^{q_{\i}}_{\sub}    \right)  \e_{k_{\i}} \e_{q_{\i}}^\top \right] \X_\sub  \\
    & =  \X^\top_\sub \left[
        \frac{2}{n} \sum_{k_{\i}=1}^n  \loss \bigl( \bbh  ; \x^{k_\i} , y^{k_\i}; \sub  \bigl)  \e_{k_{\i}} \e_{k_{\i}}^\top \;  - \frac{1}{n^2}
        \sum_{k_{\i},q_{\i}=1}^n    \left( y^{k_{\i}}  - \betas^{\top} \x^{k_{\i}}_{\sub}    \right) \left( y^{q_{\i}}  - \betas^\top \x^{q_{\i}}_{\sub}    \right)  \e_{k_{\i}} \e_{q_{\i}}^\top \right] \X_\sub  \\
      & = \frac{2}{n} \X^\top_\sub \begin{pmatrix}
        l \bigl( \betas ; \x^{1}, y^{1} ; \sub \bigl) & 0 & \dots & 0 \\
        0 &  l \bigl( \betas ; \x^{2}, y^{2}; \sub  \bigl) & \dots & 0 \\
        \vdots & \vdots & \ddots & \vdots \\
        0 & 0 & \dots &  l \bigl( \betas ; \x^{n}, y^{n}; \sub  \bigl)  
      \end{pmatrix}\X_\sub  + \bigO \left(\frac{1}{n^2}\right) \;.
\end{aligned}
\end{equation}
    
We neglect $ \bigO \left(n^{-2}\right) $ contributions and assume that the individual loss function $ l \bigl( \betas ; \x^{k}, y^{k} ; \sub \bigl)$ is equal to its average $ \lossL\bigl(\bbh ;\X,\y; \sub\bigl) $ . Indeed, the result is then
\begin{equation} 
        \bm{\Sigma} \bigl(\bbh(t) ;\X,\y;\sub\bigl) \approx \frac{2}{n} \lossL\bigl(\bbh(t);\X,\y;\sub\bigl) \X^\top_\sub  \X_\sub \;,
\end{equation}
which matches Eq.~\eqref{eq:diff_gen_belkin}. Along with Eq.~\eqref{eq:belkin_drift}, we obtain Eq.~\eqref{eq:belkin_sde}.

\subsection{Solving gradient flow equation}\label{app:belkin:gf}

We assume initial condition $\betas(0) \equiv \bbhzs$. When $\gamma = 0$, SDE~\eqref{eq:belkin_sde} becomes the following ODE:
\begin{equation}
    \label{eq:belkin_ode}
    \dv{\betas}{t} \!= \!\frac{1}{n}  \X_{\sub}^\top  ( \y  - \X_{\sub} \betas)   \;.
\end{equation} 

This is a first-order matrix differential equation. Applying \eqref{eq:xsub_svd}, we can write it as:
    \begin{equation}
 \dv{\betas}{t}  = -\frac{1}{n} \Vsvd \Lsvd^\top \Lsvd \Vsvd^\top \betas + \frac{1}{n} \Vsvd  \Lsvd^\top \Usvd^\top \y  \;.
    \end{equation}

Now, we can transform it to a system of independent linear ODEs by multiplying both sides by $\Vsvd^\top$, which is full-rank:
\begin{equation}
    \dv{(\Vsvd^\top \betas)}{t}  =  - \frac{1}{n}   \Lsvd^\top \Lsvd  (\Vsvd^\top \betas) + \frac{1}{n}   \Lsvd^\top \Usvd^\top \y   \;,
\end{equation}
i.e., we solve:
\begin{equation} 
     \dv{\betat}{t}  =  - \frac{1}{n}   \Lsvd^\top \Lsvd  \betat + \frac{1}{n}   \Lsvd^\top \Usvd^\top \y \;,
\end{equation}
where $\betat \equiv \Vsvd^\top \betas $. Let $\betat (t) = \left( \bbtcoord_1 (t), \dots, \bbtcoord_p (t) \right)$ be its component representation. Then, the system can be written as:
\begin{equation}
    \dv{\bbtcoord_i}{t} =  \begin{cases}
          - \frac{1}{n} \lambda^2_i \bbtcoord_i + \frac{1}{n} \lambda_i \usvd^\top_i \bm{y} &  \text{for }\; i \leq r \;; \\
         0 &  \text{for }\;  i > r \;.
      \end{cases}
  \end{equation}
with initial condition $  \betat (0) = \Vsvd^\top \betas (0) = \Vsvd^\top  \bbhzs $. The solution of the system is then: 
\begin{equation}
     \bbtcoord_i (t) =  \begin{cases} e^{ - \frac{1}{n} \lambda^2_i t} \left( \bbtcoord_i (0) - \frac{\usvd^\top_i \bm{y}}{\lambda_i} \right)
        + \frac{\usvd^\top_i \bm{y}}{\lambda_i} & \text{for }\; i \leq r  \;; \\
        \bbtcoord_i (0)  & \text{for }\;  i > r  \;.
   \end{cases}
  \end{equation}
In vector form, it can be written as:
\begin{equation}
    \betat(t) =  e^{ - \frac{1}{n} \Lsvd^\top \Lsvd t} \betat(0) + \ID_{p\times r} \Lsvd^{-1}_r  \left( \ID_r - e^{-\frac{1}{n}\Lsvd^2_r t} \right) \Usvd^\top_r \y \;.
\end{equation}
Multiplying by $\Vsvd$, we get back to the original basis
\begin{equation}
    \betas(t)  =    
    \Vsvd e^{ - \frac{1}{n} \Lsvd^\top \Lsvd t} \Vsvd^\top  \betas(0) +  \Vsvd_r  \Lsvd^{-1}_r \Usvd^\top_r 
  \left( \IDn  -\Usvd_r  e^{-\frac{1}{n}\Lsvd^2_r t} \Usvd^\top_r  \right) \bm{y} \;,
\end{equation}
or
\begin{equation}
    \betas(t)  =    
    e^{-\frac{1}{n} \X^\top_{\sub} \X_{\sub} t}  \betas(0) +  \X^{\dagger}_{\sub}
  \left( \IDn  -\Usvd_r  e^{-\frac{1}{n}\Lsvd^2_r t} \Usvd^\top_r  \right) \bm{y} \;.
\end{equation}
Note that, using Eq.~\eqref{eq:exp_XX_svd}:
\begin{equation}
\begin{aligned}
    \X^{\dagger}_{\sub}  e^{-\frac{1}{n}  \X_{\sub} \X^\top_{\sub} t }  & =  \Vsvd_r  \Lsvd^{-1}_r \Usvd^\top_r \cdot  \Usvd e^{-\frac{1}{n } \Lsvd \Lsvd^\top t } \Usvd^\top =  \Vsvd_r  \Lsvd^{-1}_r \Usvd^\top_r \left( \Usvd_r e^{-\frac{1}{n } \Lsvd^2_r  t } \Usvd^\top_r + \Usvd_{\rorth} \Usvd^\top_{\rorth} \right) =  \\
    &= \Vsvd_r  \Lsvd^{-1}_r \Usvd^\top_r \Usvd_r e^{-\frac{1}{n } \Lsvd^2_r  t } \Usvd^\top_r =   \X^{\dagger}_{\sub}  \Usvd_r e^{-\frac{1}{n } \Lsvd^2_r  t } \Usvd^\top_r \;,
\end{aligned}
\end{equation}
which leads us to the concluding form of the GF trajectory solution shown in Eq.~\eqref{eq:belkin_ode_sol}:
\begin{equation}
    \label{eq:belkin_ode_sol_appendix}
    \betasode(t)  =    
    e^{-\frac{1}{n} \X^\top_{\sub} \X_{\sub} t}  \betas(0) +  \X^{\dagger}_{\sub}
  \left( \IDn  - e^{-\frac{1}{n}  \X_{\sub} \X^\top_{\sub} t }  \right) \y \;.
\end{equation}
When $t \to \infty$, we can observe using Eq.~\eqref{eq:exp_XX_svd}:
\begin{equation}
    \betasode(t) \overset{t \to \infty}{\longrightarrow}   
    \Vsvd_\rorth \Vsvd^\top_\rorth   \betas(0) +  \X^{\dagger}_{\sub}
  \left( \IDn  -   \Usvd_\rorth \Usvd^\top_\rorth    \right) \y =   \Vsvd_\rorth \Vsvd^\top_\rorth   \betas(0)  +  \X^{\dagger}_{\sub}  \y \;.
\end{equation}
Therefore, GF converges to the minimum norm solution of the least-squares problem $\min_{\betas \in \R^p} \lossL\bigl(\bbh;\X,\y; \sub\bigl)$ with one caveat -- when $r < p$ the problem is overparametrized and therefore the solution includes the projection of $\betas(0)$ onto the subspace orthogonal to row space of $\X_\sub$. In the Gaussian setting, aside from cases with probability $0$, this happens iff $n < p$. 

%%%%%%%%%%%%%%%%%%%
\subsection{Evaluating the generalization error for gradient flow}\label{app:belkin:gf_test}
%%%%%%%%%%%%%%%%%%%

Our goal is to compute: 
\begin{equation}  
\label{eq:etestode_def}
     \Etestode  = \E_{\sub; \X, \y  }  \left[ \E_{ \xnew,  \ynew }\left[ 
        \loss \bigl( \betasode(t); \xnew , \ynew  \bigl)
     \right] \right] =  \frac{1}{2}
    \E_{\sub;  \X, \y; \xnew,  \ynew }  \left[ \left( \ynew -  \left( \xnew_{\sub} \right)^\top  \betasode(t)\right)^2 \right] \;,
\end{equation}

where $ \xnew,  \ynew \sim \Prob ( \xnew, \ynew) $ is a new sample from the distribution described in Section~\ref{sec:simple-model-regression}.

First, we look at the expectation over a new sample; as $\ynew = \bb^\top    \xnew + \mu \epsnew = \bb^\top_\sub \xnew_\sub + \bb^\top_\subc \xnew_\subc  + \mu \epsnew $ and $ \xnew_\sub \sim \N ( \xnew_\sub | \bm{0}, \ID_p ), \xnew_\subc  \sim \N  ( \xnew_\subc | \bm{0}, \ID_{d - p} ), \epsnew \sim \N ( \epsnew |0, 1) $ are mutually independent,
\begin{equation}
\label{eq:avg_xnew_ynew}
\begin{aligned}
    \E_{ \xnew,  \ynew } & \left[ \left( \ynew -   \left( \xnew_{\sub} \right)^\top    \betasode(t)\right)^2 \right] = \\
    % =  \E_{ \xnew,  \epsnew }  \left[ \left( \left( \bb_\sub - \betasode(t) \right)^\top \xnew_\sub + \bb^\top_\subc \xnew_\subc  + \mu \epsnew       \right)^2 \right] = \\
    &= \E_{ \xnew_\sub  }  \left[ \left( \left( \bb_\sub - \betasode(t) \right)^\top \xnew_\sub \right)^2 \right] +  \E_{ \xnew_\subc  }  \left[  \left( \bb^\top_\subc \xnew_\subc \right)^2 \right]  + \mu^2  \E_{ \epsnew     }  \left[ \left(  \epsnew \right)^2 \right] \\
    &=    \left( \bb_\sub - \betasode(t) \right)^\top  \E_{ \xnew_\sub  }  \left[ \xnew_\sub  \left(  \xnew_\sub \right)^\top \right] \left( \bb_\sub - \betasode(t) \right)  +   \bb^\top_\subc  \E_{ \xnew_\subc  }  \left[  \xnew_\subc  \left( \xnew_\subc \right)^\top \right]  \bb_\subc  + \mu^2  \\
    &=  \| \bb_\sub - \betasode(t) \|^2  +  \| \bb_\subc \|^2   + \mu^2 \;.
\end{aligned}  
\end{equation}

We then continue deriving the expression for Eq.~\eqref{eq:etestode_def} by taking the expectation over $\left( \X, \y \right)$.  Only the first term of the resulting sum of Eq.~\eqref{eq:avg_xnew_ynew} depends on it, therefore our goal now is to compute $\E_{\X, \y} \left[ \| \bb_\sub - \betasode(t) \|^2 \right] $. 

We denote $\beps := \left( \epsilon^1, \dots, \epsilon^n \right) \in \R^n$ as a vector of additive Gaussian noises. Then, we can write the target vector as:
\begin{equation}
    \label{eq:y_decomposition}
    \y = \X_{\sub} \bb_\sub + \X_{\subc} \bb_{\subc} + \mu \beps \;.
\end{equation}
The vector of differences between the true value of a parameter and GF estimate from Eq.~\eqref{eq:belkin_ode_sol_appendix}  is:
\begin{multline}
\label{eq:betadiff_ode} 
        \bb_\sub - \betasode(t)
         =
    \bb_\sub -   e^{ - \frac{1}{n} \X^\top_{\sub} \X_{\sub} t}  \betas(0) -  \X^\dagger_{\sub} \left( \IDn  - e^{-\frac{1}{n}  \X_{\sub} \X^\top_{\sub} t }  \right) \y = \\
      =  \left( \ID_p - \X^\dagger_{\sub} \left( \IDn  - e^{-\frac{1}{n}  \X_{\sub} \X^\top_{\sub} t }  \right) \X_{\sub} \right) \bb_\sub -   e^{ - \frac{1}{n} \X^\top_{\sub} \X_{\sub} t}  \betas(0) -  \X^\dagger_{\sub} \left( \ID_n  - e^{-\frac{1}{n}  \X_{\sub} \X^\top_{\sub} t }  \right) \left( \X_{\subc} \bb_{\subc} + \mu \beps \right) \;.
\end{multline}
As $ \X_{\subc} \bb_{\subc} + \mu \beps$ is zero-mean vector independent from $\X_\sub$, we can write:
\begin{equation}
\label{eq:avg_betadiff_split}
\begin{aligned}
        \E_{\X, \y}  \left[ \| \bb_\sub - \betasode(t) \|^2_2 \right] 
        & =
         \E_{\X_\sub}  \left[ \| 
        \left( \ID_p - \X^\dagger_{\sub} \left( \IDn  - e^{-\frac{1}{n}  \X_{\sub} \X^\top_{\sub} t }  \right)  \X_{\sub} \right)
        \bb_\sub -   e^{ - \frac{1}{n} \X^\top_{\sub} \X_{\sub} t}  \betas(0)   
        \|^2 \right]  \\
        & + \E_{\X_\sub, \X_\subc, \beps}  \left[ \|  \X^\dagger_{\sub} \left( \ID_n  - e^{-\frac{1}{n}  \X_{\sub} \X^\top_{\sub} t }  \right) \left( \X_{\subc} \bb_{\subc} + \mu \beps \right)   
        \|^2  \right] \;.
\end{aligned}
\end{equation}

Using Eqs.~\eqref{eq:xsub_svd},~\eqref{eq:X_moore_penrose},~\eqref{eq:exp_XX_svd}, we can simplify the first term, because:
\begin{equation}
\begin{aligned}
   & \left( \ID_p - \X^\dagger_{\sub} \left( \IDn  - e^{-\frac{1}{n}  \X_{\sub} \X^\top_{\sub} t }  \right)  \X_{\sub} \right)
        \bb_\sub 
         = \\
       &= \left( \ID_p - \Vsvd_r \Lsvd^{-1}_r \Usvd^\top_r \left( \IDn  -\Usvd_r e^{-\frac{1}{n } \Lsvd^2_r t } \Usvd^\top_r - \Usvd_\rorth \Usvd^\top_\rorth    \right)  \Usvd_r \Lsvd_r \Vsvd^\top_r  \right) \bb_\sub  = \\
       &=
        \left( \ID_p - \Vsvd_r \Vsvd^\top_r  +
        \Vsvd_r     e^{-\frac{1}{n } \Lsvd^2_r t }         \Vsvd^\top_r  \right) \bb_\sub    =    \left(  
    \Vsvd_r     e^{-\frac{1}{n } \Lsvd^2_r t }         \Vsvd^\top_r + \Vsvd_\rorth \Vsvd^\top_\rorth  \right) \bb_\sub =  \\
      &=  e^{ - \frac{1}{n} \X^\top_{\sub} \X_{\sub} t}  \bb_\sub \;,
\end{aligned}
\end{equation}
and, therefore:
\begin{multline}
  \E_{\X_\sub}  \left[ \| 
    \left( \ID_p - \X^\dagger_{\sub} \left( \IDn  - e^{-\frac{1}{n}  \X_{\sub} \X^\top_{\sub} t }  \right)  \X_{\sub} \right)
    \bb_\sub -   e^{ - \frac{1}{n} \X^\top_{\sub} \X_{\sub} t}  \betas(0)   
    \|^2 \right] = \\
     =  \E_{\X_\sub}  \left[ \| 
    e^{ - \frac{1}{n} \X^\top_{\sub} \X_{\sub} t}  (
    \bb_\sub -   \betas(0) )
    \|^2 \right] = 
    \left(   \bb_\sub -   \betas(0) \right)^\top
    \E_{\X_\sub}  \left[  
    e^{ - \frac{2}{n} \X^\top_{\sub} \X_{\sub} t}  \right]
    \left(   \bb_\sub -   \betas(0) \right) \;.
\end{multline}
Going back to the representation of Eq.~\eqref{eq:exp_XX_svd}, we can rewrite it in terms of the singular values of the matrix $\X_\sub$. Indeed, applying trace trick (again, $\Lsvd$ and $\Vsvd$ are independent):
\begin{equation}
\begin{aligned}
    \E_{\X_\sub}  & \left[ \| 
    \left( \ID_p - \X^\dagger_{\sub} \left( \IDn  - e^{-\frac{1}{n}  \X_{\sub} \X^\top_{\sub} t }  \right)  \X_{\sub} \right)
    \bb_\sub -   e^{ - \frac{1}{n} \X^\top_{\sub} \X_{\sub} t}  \betas(0)   
    \|^2 \right]
     = \\ & =   \Tr{
    \E_{\Lsvd, \Vsvd} \left[
        \left( \Vsvd^\top \left( \bb_{\sub} - \betas(0)\right)\right)^\top  e^{-\frac{2}{n} \Lsvd^\top \Lsvd  t } \Vsvd^\top \left( \bb_{\sub} - \betas(0)\right)  \right]  }
    = \\ & =
    \Tr{ \E_{\Lsvd} \left[e^{-\frac{2}{n} \Lsvd^\top \Lsvd  t }  \right]  \E_{\Vsvd} \left[ \Vsvd^\top \left( \bb_{\sub} - \betas(0)\right)   \left( \bb_{\sub} - \betas(0)\right)^\top   \Vsvd   \right]  }
    \;.
\end{aligned}
\end{equation}

As $\Vsvd \sim \Unif{\orthg{p}}$, if $\avec \in \R^p$ is arbitrary vector of dimensionality $p$, then
\begin{subequations}
\label{eq:V_projection_distrib}
\begin{equation}
    \Vsvd^\top \avec \sim \Unif{\sphere{p}{\|\avec\|}} \;, 
\end{equation}
where $\sphere{d}{r }$ denotes $d$-dimensional sphere  with radius $r$,  and thus
\begin{equation}
    \E_{\Vsvd} \left[ \Vsvd^\top \avec   \right] = \zero_{p}, \: \E_{\Vsvd} \left[  \Vsvd^\top \avec   \left( \Vsvd^\top \avec  \right)^\top  \right] = \frac{1}{p} \|\avec\|^2 \cdot \ID_p \;.
\end{equation}
\end{subequations}

This results in:
\begin{equation}
\label{eq:avg_betadiff_1_final}
\begin{aligned}
    \E_{\X_\sub} & \left[ \| 
    \left( \ID_p - \X^\dagger_{\sub} \left( \IDn  - e^{-\frac{1}{n}  \X_{\sub} \X^\top_{\sub} t }  \right)  \X_{\sub} \right)
    \bb_\sub -   e^{ - \frac{1}{n} \X^\top_{\sub} \X_{\sub} t}  \betas(0)   
    \|^2 \right]
    = \\
    & =
     \frac{1}{p}  \| \bb_{\sub} - \betas(0) \|^2
    \Tr{ \E_{\Lsvd} \left[e^{-\frac{2}{n} \Lsvd^\top \Lsvd  t }  \right] } \\
    &
    =
      \| \bb_{\sub} - \betas(0) \|^2 \cdot
     \frac{1}{p}
    \Tr{ \E_{\Lsvd} \left[e^{-\frac{2}{n} \Lsvd^2_r   t } + (p - r)  \right] }
     =
    \\
    & = 
    \| \bb_{\sub} - \betas(0) \|^2 \cdot \left( 
    \frac{1}{p}
   \Tr{ \E_{\Lsvd} \left[e^{-\frac{2}{n} \Lsvd^2_r   t }   \right] } + \max\left(0, 1 - \frac{n}{p }\right)
   \right)
   \;.     
\end{aligned}
\end{equation}

For the second term of Eq.~\eqref{eq:avg_betadiff_split}, we can apply trace trick again:
\begin{equation}
\label{eq:avg_betadiff_2_trace_trick}
\begin{aligned}
  &  \E_{\X_\sub, \X_\subc, \beps}  \left[ \|  \X^\dagger_{\sub} \left( \ID_n  - e^{-\frac{1}{n}  \X_{\sub} \X^\top_{\sub} t }  \right) \left( \X_{\subc} \bb_{\subc} + \mu \beps \right)   
        \|^2  \right] = \\
& =   
    \E_{\X_\sub, \X_\subc, \beps} \left[ \Tr{
        \left( \X_{\subc} \bb_{\subc} + \mu \beps \right)^\top    
    \left(  \X^\dagger_{\sub} \left( \ID_n  - e^{-\frac{1}{n}  \X_{\sub} \X^\top_{\sub} t }  \right) \right)^\top 
    \X^\dagger_{\sub} \left( \ID_n  - e^{-\frac{1}{n}  \X_{\sub} \X^\top_{\sub} t }  \right)
   \left( \X_{\subc} \bb_{\subc} + \mu \beps \right)
    } \right] = \\
& =  \Tr{ \E_{\X_{\sub}} \left[ \left( \ID_n  - e^{-\frac{1}{n}  \X_{\sub} \X^\top_{\sub} t }  \right) \left(  \X^\dagger_{\sub}  \right)^\top 
\X^\dagger_{\sub} \left( \ID_n  - e^{-\frac{1}{n}  \X_{\sub} \X^\top_{\sub} t }  \right) \right]
\E_{\X_{\subc}, \beps}  \left[
\left( \X_{\subc} \bb_{\subc} + \mu \beps \right)
\left( \X_{\subc} \bb_{\subc} + \mu \beps \right)^\top \right] } \;.
\end{aligned}
\end{equation}
We note that the  vector $\X_{\subc} \bb_{\subc} + \mu \beps$ is distributed as:
\begin{equation}
\label{eq:Xsubc_mueps_distrib}
    \X_{\subc} \bb_{\subc} + \mu \beps \sim \N  \left(\bm{0}, \left(
    \|\bb_{\subc}  \|^2 + \mu^2 
     \right) \ID_n \right) \;,
\end{equation}
and that on the other hand,
\begin{multline}
\label{eq:sol_gf_mult_before_y}
    \X^\dagger_{\sub} \left( \ID_n  - e^{-\frac{1}{n}  \X_{\sub} \X^\top_{\sub} t }  \right)
    = \Vsvd_r \Lsvd^{-1}_r \Usvd^\top_r \left( \ID_n  - \Usvd_r e^{-\frac{1}{n } \Lsvd^2_r t } \Usvd^\top_r - \Usvd_\rorth \Usvd^\top_\rorth   \right) = \\
    = \Vsvd_r \Lsvd^{-1}_r \left(  \Usvd^\top_r   -  e^{-\frac{1}{n } \Lsvd^2_r t } \Usvd^\top_r   \right) =  \Vsvd_r \Lsvd^{-1}_r \left(  \ID_r  -  e^{-\frac{1}{n } \Lsvd^2_r t }    \right)  \Usvd^\top_r \;;
\end{multline}
thus it follows from Eq.~\eqref{eq:avg_betadiff_2_trace_trick}:
\begin{equation}
\label{eq:avg_betadiff_2_final}
\begin{aligned} 
    \E_{\X_\sub, \X_\subc, \beps}  & \left[ \|  \X^\dagger_{\sub} \left( \ID_n  - e^{-\frac{1}{n}  \X_{\sub} \X^\top_{\sub} t }  \right) \left( \X_{\subc} \bb_{\subc} + \mu \beps \right)   
        \|^2  \right] =  \\
        & =
        \left(
            \|\bb_{\subc}  \|^2 + \mu^2 
                \right)
            \Tr{ \E_{\Usvd, \Lsvd, \Vsvd } \left[  \Usvd_r  \left(  \ID_r  -  e^{-\frac{1}{n } \Lsvd^2_r t }    \right)  \Lsvd^{-1}_r \Vsvd^\top_r
                \Vsvd_r \Lsvd^{-1}_r \left(  \ID_r  -  e^{-\frac{1}{n } \Lsvd^2_r t }    \right)  \Usvd^\top_r
            \right]  } = \\
    & = \left(
        \|\bb_{\subc}  \|^2 + \mu^2 
            \right)
        \Tr{ \E_{  \Lsvd } \left[   \Lsvd^{-2}_r \left(  \ID_r  -  e^{-\frac{1}{n } \Lsvd^2_r t }    \right)^2   
        \right]  } \;.
\end{aligned}
\end{equation}
    
Gathering expressions from \eqref{eq:avg_betadiff_1_final} and \eqref{eq:avg_betadiff_2_final} back to \eqref{eq:avg_betadiff_split}, we obtain:
\begin{multline}
    \E_{\X, \y}  \left[ \| \bb_\sub - \betasode(t) \|^2_2 \right]  = 
    \| \bb_{\sub} - \betas(0) \|^2 \cdot \left( 
        \max\left(0, 1 - \frac{n}{p }\right) + 
    \frac{1}{p}
   \Tr{ \E_{\Lsvd} \left[e^{-\frac{2}{n} \Lsvd^2_r   t }   \right] } 
   \right) + \\
   + \left(
    \|\bb_{\subc}  \|^2 + \mu^2 
        \right)
    \Tr{ \E_{  \Lsvd } \left[   \Lsvd^{-2}_r \left(  \ID_r  -  e^{-\frac{1}{n } \Lsvd^2_r t }    \right)^2   
    \right]  }\;,
\end{multline}
which can be used to obtain average test error over $\X, \y$ by taking the expectation over both first and last equal expressions of Eq.~\eqref{eq:avg_xnew_ynew} and substituting one of the terms with this one.

We return to test error Eq.~\eqref{eq:etestode_def} and average over $\sub$. The only terms depending on $\sub$ are $\| \bb_{\sub} - \betas(0) \|^2 $ and $\|\bb_{\subc}  \|^2 $. As $\sub$ in our model is a uniformly random subset of $[d]$ of cardinality $p$, 
%\ar{Do we need proof of this or it is obvious enough?}\nm{obvious enough after 20 something pages}
\begin{equation}
    \label{eq:sub_avg}
    \E_{\sub} \left[ \| \bb_{\sub} - \betas(0) \|^2 \right] =  \frac{p}{d} \cdot \| \bb  - \bbhz \|^2, \; \E_{\sub} \left[ \|  \bb_{\subc}  \|^2 \right] = \left(1 - \frac{p}{d} \right) \| \bb \|^2 \;.
\end{equation}

Finally, we obtain:
\begin{equation}
\label{eq:etest_ode_repeat} 
\begin{aligned}
    \Etestode  &=  \frac{1}{2}
    \E_{\sub }  \left[  \E_{\X, \y}  \left[ \| \bb_\sub - \betasode(t) \|^2_2 \right]   +  \| \bb_\subc \|^2   + \mu^2  \right]  
    = \\ &= 
    \frac{1}{2} \Bigg( \| \bb  - \bbhz  \|^2
     \cdot \left( 
        \max\left(0, \; \frac{p - n}{d} \right) + 
    \frac{1}{d}
   \Tr{ \E_{\Lsvd} \left[e^{-\frac{2}{n} \Lsvd^2_r   t }   \right] } 
   \right) 
   + \\ &+
     \left(
    \left(1 - \frac{p}{d} \right) \| \bb \|^2 + \mu^2 
        \right)
    \left[
        1 + 
    \Tr{ \E_{  \Lsvd } \left[   \Lsvd^{-2}_r \left(  \ID_r  -  e^{-\frac{1}{n } \Lsvd^2_r t }    \right)^2   
    \right]  } \right] \Bigg) \;, 
\end{aligned}
\end{equation}
which matches expression \eqref{eq:etest_ode}.

\subsection{Evaluating train error for gradient flow}\label{app:belkin:gf_train} 

In this section, we compute the average train error for the GF solution:
\begin{equation}
\label{eq:etrainode_def}
    \Etrainode =  \E_{\sub; \X, \y} \left[ \lossL\bigl( \betasode(t) ;\X,\y; \sub\bigl) \right] = \frac{1}{2 n} \E_{\sub; \X, \y} \left[  \| \y - \X_\sub \betasode(t)\|^2 \right] \;.
\end{equation}
While in this paper we do not insist on the properties of the train error, the derivation itself will prove useful in \ref{app:belkin:sgf_test}. This is not surprising, as the train error is involved in the modeling of the diffusion matrix.   

First, we look at the 'vector of train errors'; we substitute the gradient flow estimator with the solution from Eq.~\eqref{eq:belkin_ode_sol_appendix}:
\begin{equation}
    \y - \X_\sub \betasode(t) = \left[ \IDn -  \X_{\sub} \X^{\dagger}_{\sub}
    \left( \IDn  - e^{-\frac{1}{n}  \X_{\sub} \X^{\top}_{\sub} t }  \right) \right] \y - \X_{\sub}  e^{-\frac{1}{n} \X^\top_{\sub} \X_{\sub} t}  \betas(0) \;.
\end{equation}

The multiplier in the first term can be simplified utilizing Eqs.~\eqref{eq:sol_gf_mult_before_y},~\eqref{eq:exp_XX_svd}: 
\begin{multline}
    \IDn -  \X_{\sub} \X^{\dagger}_{\sub}
    \left( \IDn  - e^{-\frac{1}{n}  \X_{\sub} \X^{\top}_{\sub} t }  \right) = 
    \IDn -  \Usvd_r \Lsvd_r \Vsvd^\top_r   \Vsvd_r \Lsvd^{-1}_r \left(  \ID_r  -  e^{-\frac{1}{n } \Lsvd^2_r t }    \right)  \Usvd^\top_r = \\
    =  \Usvd_r  \Usvd^\top_r +  \Usvd_\rorth  \Usvd^\top_\rorth - 
   \Usvd_r  \Usvd^\top_r  + \Usvd_r  e^{-\frac{1}{n } \Lsvd^2_r t }  \Usvd^\top_r
   =  \Usvd_r  e^{-\frac{1}{n } \Lsvd^2_r t }  \Usvd^\top_r +   \Usvd_\rorth  \Usvd^\top_\rorth = e^{-\frac{1}{n}  \X_{\sub} \X^{\top}_{\sub} t } \;;
\end{multline}
with this and the explicit form of $\y$ from Eq.~\eqref{eq:y_decomposition}, we obtain:
\begin{equation}
    \y - \X_\sub \betasode(t) = e^{-\frac{1}{n}  \X_{\sub} \X^{\top}_{\sub} t }  \X_{\sub} \bb_\sub - \X_{\sub}  e^{-\frac{1}{n} \X^\top_{\sub} \X_{\sub} t}  \betas(0)    + e^{-\frac{1}{n}  \X_{\sub} \X^{\top}_{\sub} t }
    \left(   \X_{\subc} \bb_{\subc} + \mu \beps 
    \right) \;.
\end{equation}

Now, using properties of SVD of $\X_{\sub}$, we can see that:
\begin{equation}
\begin{aligned}
    e^{-\frac{1}{n}  \X_{\sub} \X^{\top}_{\sub} t }  \X_{\sub}  &= \left( \Usvd_r  e^{-\frac{1}{n } \Lsvd^2_r t }  \Usvd^\top_r  +   \Usvd_\rorth  \Usvd^\top_\rorth \right) \Usvd_r \Lsvd_r \Vsvd^\top_r  = \Usvd_r \Lsvd_r  e^{-\frac{1}{n } \Lsvd^2_r t }  \Vsvd^\top_r \;; \\
    \X_{\sub}  e^{-\frac{1}{n} \X^\top_{\sub} \X_{\sub} t}  &=  \Usvd_r \Lsvd_r \Vsvd^\top_r 
    \left(   \Vsvd_r e^{-\frac{1}{n } \Lsvd^2_r t } \Vsvd^\top_r + \Vsvd_\rorth \Vsvd^\top_\rorth  \right) = \Usvd_r  \Lsvd_r  e^{-\frac{1}{n } \Lsvd^2_r t }  \Vsvd^\top_r \;.
\end{aligned}
\end{equation}
Then, the vector of train errors can be split into two terms -- one depending on $\bb_\sub -   \betas(0) $, another on $\X_{\subc} \bb_{\subc} + \mu \beps $:
\begin{equation}
\label{eq:etrainode_errorvec_final}
    \y - \X_\sub \betasode(t) =  \Usvd_r  \Lsvd_r  e^{-\frac{1}{n } \Lsvd^2_r t }  \Vsvd^\top_r \left( \bb_\sub -   \betas(0) \right)   + e^{-\frac{1}{n}  \X_{\sub} \X^{\top}_{\sub} t }
    \left(   \X_{\subc} \bb_{\subc} + \mu \beps 
    \right) \;.
\end{equation}

Now, we would like to take expectation of $\| \y - \X_\sub \betasode(t)\|^2$ over $\X, \y$, or, equivalently,  over  $\X_\sub,  \X_{\subc}, \beps $. We start with the latter two. Again $ \X_{\subc} \bb_{\subc} + \mu \beps$ is zero-mean vector independent from $\X_\sub$,
so we can write:
\begin{equation}
\label{eq:etrainode_avg_subc}
    \E_{ \X_{\subc}, \beps } \left[ \| \y - \X_\sub \betasode(t) \|^2 \right] = 
    \|\Usvd_r \Lsvd_r   e^{-\frac{1}{n } \Lsvd^2_r t }  \Vsvd^\top_r \left( \bb_\sub -   \betas(0) \right) \|^2 
    + 
    \E_{ \X_{\subc}, \beps } \left[  \| e^{-\frac{1}{n}  \X_{\sub} \X^{\top}_{\sub} t }
    \left(   \X_{\subc} \bb_{\subc} + \mu \beps 
    \right)  \|^2 \right] \;.
\end{equation}
The first term can be reduced to:
\begin{equation}
\begin{aligned}
    \|\Usvd_r \Lsvd_r   e^{-\frac{1}{n } \Lsvd^2_r t }  \Vsvd^\top_r \left( \bb_\sub -   \betas(0) \right) \|^2 
    &=\left( \bb_\sub -   \betas(0) \right)^\top \Vsvd_r   e^{-\frac{1}{n } \Lsvd^2_r t } \Lsvd_r     \Usvd^\top_r
     \Usvd_r \Lsvd_r   e^{-\frac{1}{n } \Lsvd^2_r t }  \Vsvd^\top_r  \left( \bb_\sub -   \betas(0) \right)   \\ 
     &= \left( \bb_\sub -   \betas(0) \right)^\top \Vsvd_r    \Lsvd^2_r     e^{-\frac{2}{n } \Lsvd^2_r t }  \Vsvd^\top_r  \left( \bb_\sub -   \betas(0) \right) \;,
\end{aligned}
\end{equation}
To take the expectation in the second one, along the lines of Eq.~\eqref{eq:avg_betadiff_2_trace_trick}, we use again use trace trick and  the covariance of vector $  \X_{\subc} \bb_{\subc} + \mu \beps $ (see Eq.~\eqref{eq:Xsubc_mueps_distrib}):
\begin{multline} 
    \E_{ \X_{\subc}, \beps } \left[  \| e^{-\frac{1}{n}  \X_{\sub} \X^{\top}_{\sub} t }
    \left(   \X_{\subc} \bb_{\subc} + \mu \beps 
    \right)  \|^2 \right] = \Tr{
        \E_{ \X_{\subc}, \beps } \left[  
            \left(   \X_{\subc} \bb_{\subc} + \mu \beps 
            \right)^\top    
         e^{-\frac{2}{n}  \X_{\sub} \X^{\top}_{\sub} t }
    \left(   \X_{\subc} \bb_{\subc} + \mu \beps 
    \right)  \right]
    } = \\
    = \Tr{ e^{-\frac{2}{n}  \X_{\sub} \X^{\top}_{\sub} t } \cdot
        \E_{ \X_{\subc}, \beps } \left[  
    \left(   \X_{\subc} \bb_{\subc} + \mu \beps 
    \right)  \left(   \X_{\subc} \bb_{\subc} + \mu \beps 
    \right)^\top      \right]
    } =  \left(
        \|\bb_{\subc}  \|^2 + \mu^2 
         \right)    \Tr{ e^{-\frac{2}{n}  \X_{\sub} \X^{\top}_{\sub} t }  } \;.   
\end{multline}
We write this result in terms of SVDs of matrices (noting that, in fact, the expectation over $\Usvd$ is not necessary here):
\begin{equation}
    \Tr{ e^{-\frac{2}{n}  \X_{\sub} \X^{\top}_{\sub} t }  } =  \Tr{   \Usvd e^{-\frac{2}{n } \Lsvd \Lsvd^\top t } \Usvd^\top   } = \Tr{    e^{-\frac{2}{n } \Lsvd \Lsvd^\top t }   } =  \Tr{    e^{-\frac{2}{n } \Lsvd^2_r t }   } + n - r \;.
\end{equation}
All in all, we collect everything back into Eq.~\eqref{eq:etrainode_avg_subc}:
\begin{multline} 
    \E_{\X_{\subc}, \beps } \left[  \| \y - \X_\sub \betasode(t) \|^2   \right] = \\ 
    =  
        \left( \bb_\sub -   \betas(0) \right)^\top \Vsvd_r    \Lsvd^2_r     e^{-\frac{2}{n } \Lsvd^2_r t }  \Vsvd^\top_r  \left( \bb_\sub -   \betas(0) \right)
      +  \left(
        \|\bb_{\subc}  \|^2 + \mu^2 
         \right)  \left(  \Tr{    e^{-\frac{2}{n } \Lsvd^2_r t }   } + n - r \right)\;.
\end{multline}

Now, to average over $\X_{\sub}$, we start with the expectation over $\Usvd$ and $\Vsvd$ first, because the result would be convenient to employ in section \ref{app:belkin:sgf_test}. $\Usvd$ is not present at all, and $\Vsvd$ is found only in the first term. With an application of trace trick,
\begin{multline}
    \E_{\Vsvd}
    \left[ 
    \left(  \Vsvd^\top_r  \left( \bb_\sub -   \betas(0) \right) \right)^\top \Lsvd^2_r     e^{-\frac{2}{n } \Lsvd^2_r t }  \Vsvd^\top_r  \left( \bb_\sub -   \betas(0) \right)
    \right] = \\
    = \Tr{
        \Lsvd^2_r     e^{-\frac{2}{n } \Lsvd^2_r t }  
        \E_{\Vsvd} \left[    \Vsvd^\top_r  \left( \bb_\sub -   \betas(0) \right) \left(  \Vsvd^\top_r  \left( \bb_\sub -   \betas(0) \right) \right)^\top \right]
    } \;.
\end{multline}
We discussed distribution of the vector $ \Vsvd^\top  \left( \bb_\sub -   \betas(0) \right)$ in Eq.~\eqref{eq:V_projection_distrib}; then, we can see that $\Vsvd^\top_r  \left( \bb_\sub -   \betas(0) \right)$ are simply first $r$ components of it, and, therefore, the new vector is also zero-mean with covariance matrix is $ \frac{1}{p} \| \bb_\sub -   \betas(0) \|^2 \ID_r $. This leads us to:
\begin{equation} 
\label{eq:etrainode_sans_lsvd}
    \E_{\Usvd, \Vsvd; \X_{\subc}, \beps } \left[  \| \y - \X_\sub \betasode(t) \|^2   \right] = 
    \| \bb_\sub -   \betas(0) \|^2 \frac{1}{p} 
    \Tr{
        \Lsvd^2_r     e^{-\frac{2}{n } \Lsvd^2_r t }   
    }
      +  \left(
        \|\bb_{\subc}  \|^2 + \mu^2 
         \right)  \left(  \Tr{    e^{-\frac{2}{n } \Lsvd^2_r t }   } + n - r \right) \;.
\end{equation}

At last, the whole expectation of square of train error over $\X, \y$ is found to be 
\begin{multline}
    \E_{\X, y} \left[  \| \y - \X_\sub \betasode(t) \|^2   \right] = 
    \E_{\X_{\sub}; \X_{\subc}, \beps } \left[  \| \y - \X_\sub \betasode(t) \|^2   \right]
    = \\
    = 
    \| \bb_\sub -   \betas(0) \|^2 \frac{1}{p} 
    \Tr{
      \E_{\Lsvd} \left[  \Lsvd^2_r     e^{-\frac{2}{n } \Lsvd^2_r t }  \right]  
    }
      +  \left(
        \|\bb_{\subc}  \|^2 + \mu^2 
         \right)  \left(  \Tr{ \E_{\Lsvd} \left[   e^{-\frac{2}{n } \Lsvd^2_r t } \right]  } + \max(0, n - p) \right) \;,
\end{multline}
and, by taking average over feature set $\sub$ (sampled as in Eq.~\eqref{eq:sub_avg}) we obtain the final expression for Eq.~\eqref{eq:etrainode_def}:
\begin{multline}
    \Etrainode = \frac{1}{2 } \Bigg(   \frac{1}{d} \| \bb  - \bbhz \|^2  \cdot
    \frac{1}{n} \Tr{
      \E_{\Lsvd} \left[  \Lsvd^2_r     e^{-\frac{2}{n } \Lsvd^2_r t }  \right]  
    } + \\
      +  \left(
        \left(1 - \frac{p}{d} \right) \| \bb \|^2   + \mu^2 
         \right)  \cdot  \left(   \frac{1}{n}  \Tr{ \E_{\Lsvd} \left[   e^{-\frac{2}{n } \Lsvd^2_r t } \right]  } + \max \left(0, 1 - \frac{p}{n} \right) \right) \Bigg) \;.
\end{multline}

%%%%%%%%%%%%%%%%%%%
\subsection{Evaluating generalization error for stochastic gradient flow}\label{app:belkin:sgf_test}
%%%%%%%%%%%%%%%%%%%
Now, we would like to evaluate stochastic gradient flow test error, i.e. error for estimator $\betassde(t) = \betasode(t) +  \sqrt{\lr} \z (t)$, with $\z(0)=\bm{0}$  and  $\z(t)$ being a mean-zero Gaussian process with the covariance matrix $\C(t)$.

First of all, it is easy to see that the derivations in Eq.~\eqref{eq:avg_xnew_ynew} are still valid if we replace $\betasode$ with $\betassde$, and therefore:
\begin{equation} 
\begin{aligned}
        \Etestsde  & =
        \E_{\sub; \X, \y ; \brow    }   \left[ \E_{ \xnew,  \ynew }\left[ 
            \loss \bigl( \betassde(t); \xnew , \ynew  \bigl)
            \right] \right]  \\
            &  =
                \frac{1}{2}
    \E_{\sub;  \X, \y;  \brow  }  \left[  \E_{ \xnew,  \ynew }\left[  \left( \ynew -  \left( \xnew_{\sub} \right)^\top  \betassde(t)\right)^2   \right] \right]   \\
    & =  \frac{1}{2}
    \E_{\sub  }  \left[  \E_{ \X, \y;  \brow  }  \left[ \| \bb_\sub - \betassde(t) \|^2 \right]  +  \| \bb_\subc \|^2   + \mu^2 \right] 
    \;,
\end{aligned}
\end{equation}
so we can easily obtain equality from Eq.~\eqref{eq:etestsde_def}.

Now, as we can decompose:
\begin{equation} 
\begin{aligned}
    \E_{  \brow  }  \left[ \| \bb_\sub - \betassde(t) \|^2 \right] & =
    \E_{  \brow  }  \left[ \left\| \left(  \bb_\sub -  \betasode(t)  \right) -   \sqrt{\lr} \z (t) \right\|^2 \right]
      \\ 
      & =
      \left\|   \bb_\sub -  \betasode(t)   \right\|^2 - 2 \sqrt{\lr} \left(  \bb_\sub -  \betasode(t)  \right) 
      \E_{  \brow  }  \left[   \z (t)  \right] + \lr \:
      \E_{  \brow  }  \left[  \z^\top (t) \z (t)   \right]  \\
      & =    \left\|   \bb_\sub -  \betasode(t)   \right\|^2   + \lr 
      \Tr{ \C (t) } \;,
\end{aligned}
\end{equation}

we can see that $\Etestsde $ indeed can be split into a sum of test error $\Etestode$ of deterministic gradient flow and a term coming from perturbation and proportional to learning rate $\lr$ as in Eq.~\eqref{sum-of-two-terms}:
\begin{equation}  
        \Etestsde  =
    \Etestode + 
        \frac{\lr}{2}
    \E_{\sub;   \X, \y  }  \left[  \Tr{ \C (t) }  \right] \;.
\end{equation} 

Evaluation of the first term was done in Section~\ref{app:belkin:gf_test}. To complete it for the second term, we refer to Eq.~\eqref{cov-ML-1}.  In our setting, $\lossL\bigl(\bbh;\X,\y; \sub\bigl)   = \frac{1}{2 n}  \| \y  - \X_{\sub} \betas \|^2$, and its Hessian is
\begin{equation}
     \HessL \bigl(\bbh;\X,\y; \sub\bigl) = \grad_{\bbh_{\sub}} \otimes \grad_{\bbh_{\sub}} \lossL \bigl(\bbh;\X,\y; \sub\bigl) =   \frac{1}{n} \X^\top_{\sub} \X_{\sub} \in \R^{p \times p} \;.
\end{equation}

Then, the term in Eq.~\eqref{cov-ML-2} in our model becomes: 
\begin{align} 
    \U^\text{GF} \left(t, \time ; \X,\y; \sub    \right)  \equiv 
     \T\left\{ e^{ -  \int_\time^t \dd s \HessL \bigl(  \betasode(t)  ;\X,\y; \sub\bigl) } \right\} 
     =
     \T\left\{ e^{ - \frac{1}{n} \int_\time^t \dd s   \X^\top_{\sub} \X_{\sub}    } \right\} 
     =  e^{ - \frac{1}{n}  \X^\top_{\sub} \X_{\sub}  (t - \time)   } \;.
\end{align}
On the other hand, the diffusion matrix is, as stated in Eq.~\eqref{eq:diff_gen_belkin}:
\begin{equation} 
    \bm{\Sigma} \bigl(\bbh(t) ;\X,\y;\sub\bigl) \approx \frac{1}{n^2}  \X^\top_\sub  \X_\sub \cdot
    \| \y  - \X_{\sub} \betas(t) \|^2
    \;.
\end{equation}
This results in the following covariance matrix of the process:
\begin{equation} 
\begin{aligned} 
    \C(t) & = \int_0^t 
    \dd \time  \;  \U^\text{GF} \left(t, \time ; \X,\y; \sub   \right)  \; \bm{\Sigma} \bigl(\betasode(\time) ;\X,\y;\sub\bigl)   \;  \U^\text{GF} \left(t, \time ; \X,\y; \sub   \right)^\top  \\
    & = \frac{1}{n^2} 
    \int_0^t 
    \dd \time \;  e^{ - \frac{1}{n}  \X^\top_{\sub} \X_{\sub}  (t - \time)   }   \cdot
    \X^\top_\sub  \X_\sub \cdot
    \| \y  - \X_{\sub} \betasode(\time) \|^2 
    \cdot e^{ - \frac{1}{n}  \X^\top_{\sub} \X_{\sub}  (t - \time)   } \\
    & = \frac{1}{n^2} 
    \int_0^t 
    \dd \tau \;      \| \y  - \X_{\sub} \betasode(\time) \|^2  \cdot 
    \X^\top_\sub  \X_\sub \;
     e^{ - \frac{2}{n}  \X^\top_{\sub} \X_{\sub}  (t - \time)   } \;.
\end{aligned}
\end{equation}

As $\| \y  - \X_{\sub} \betasode(\time) \|^2 $ is scalar and with
Eq.~\eqref{eq:exp_XX_svd},
\begin{multline}
  \Tr{ \X^\top_\sub  \X_\sub \;
    e^{ - \frac{2}{n}  \X^\top_{\sub} \X_{\sub}  (t - \time)   } } 
    = 
    \Tr{  \Vsvd \Lsvd^\top \Usvd^\top \Usvd \Lsvd \Vsvd^\top \Vsvd e^{ - \frac{2}{n}  \Lsvd^\top \Lsvd  (t - \time) }  \Vsvd^\top } 
    = \\
    =
    \Tr{  \Lsvd^\top  \Lsvd e^{ - \frac{2}{n}  \Lsvd^\top \Lsvd  (t - \time) }   } 
= \Tr{  \Lsvd^2_r    e^{ - \frac{2}{n}  \Lsvd^2_r (t - \time) }   } \;,
\end{multline}
we obtain the following expression for the trace of the covariance matrix:
\begin{equation}
\Tr{\C(t)} = \frac{1}{n^2}
\int_0^t 
    \dd \tau  \;      \| \y  - \X_{\sub} \betasode(\time) \|^2  \cdot \Tr{  \Lsvd^2_r    e^{ - \frac{2}{n}  \Lsvd^2_r (t - \time) }   } \;.
\end{equation}

Now, we would like to compute $\E_{\sub;  \X, \y  }  \left[  \Tr{ \C (t) }  \right]$. Expectation over $\X, \y$ is the same as expectation over $\Usvd, \Lsvd, \Vsvd, \X_{\subc}, \beps$, and thus, following Eq.~\eqref{eq:etrainode_sans_lsvd}, 
\begin{equation}
\begin{aligned}
    \E_{ \X, \y  }  \left[  \Tr{ \C (t) }  \right] & = 
    \frac{1}{n^2} \E_{\Lsvd} \left[ 
    \int_0^t 
        \dd \time  \;    \E_{\Usvd, \Vsvd; \X_{\subc}, \beps}  
        \left[  \| \y  - \X_{\sub} \betasode(\time) \|^2 \right] \cdot \Tr{  \Lsvd^2_r    e^{ - \frac{2}{n}  \Lsvd^2_r (t - \time) }   }   \right] \\
    & =
    \frac{1}{n^2} \E_{\Lsvd} \Bigg[ 
    \int_0^t 
        \dd \time  \;    
        \Big(  \| \bb_\sub -   \betas(0) \|^2 \frac{1}{p} 
        \Tr{
            \Lsvd^2_r     e^{-\frac{2}{n } \Lsvd^2_r \time }   
        } + \\ 
          & +  \left(
            \|\bb_{\subc}  \|^2 + \mu^2 
             \right)  \left(  \Tr{    e^{-\frac{2}{n } \Lsvd^2_r  \time  }   } + n - r \right)  \Big) \cdot \Tr{  \Lsvd^2_r    e^{ - \frac{2}{n}  \Lsvd^2_r (t - \time) }   }   \Bigg] \\ 
    & = 
    \frac{1}{n^2}   
        \int_0^t 
        \dd \time   \;    
         \Bigg[  \| \bb_\sub -   \betas(0) \|^2 \frac{1}{p} 
         \E_{\Lsvd} \left[ \Tr{
            \Lsvd^2_r     e^{-\frac{2}{n } \Lsvd^2_r  \time  } }  \Tr{  \Lsvd^2_r    e^{ - \frac{2}{n}  \Lsvd^2_r (t - \time) }    
        }
        \right]   +  \left(
    \|\bb_{\subc}  \|^2 + \mu^2 
        \right)  \cdot \\
        & \cdot \left( 
            \E_{\Lsvd} \left[
                 \Tr{    e^{-\frac{2}{n } \Lsvd^2_r  \time  }   } 
        \Tr{  \Lsvd^2_r    e^{ - \frac{2}{n}  \Lsvd^2_r (t - \time) }    
        } \right] 
        + \max \left(0, n - p \right)  \E_{\Lsvd} \left[
   \Tr{  \Lsvd^2_r    e^{ - \frac{2}{n}  \Lsvd^2_r (t - \time) }    
   } \right]  \right)  \Bigg] \;.
\end{aligned}
\end{equation}

We avoid taking the integral of time over a product of traces right now as it makes further approximation complicated; however, the term below is easy to simplify: 
\begin{multline}
    \int_0^t 
    \dd \time    \;\E_{\Lsvd} \left[
\Tr{ \frac{1}{n}  \Lsvd^2_r    e^{ - \frac{2}{n}  \Lsvd^2_r (t - \time) }    
} \right] = 
  \E_{\Lsvd} \left[ \sum^r_{i = 1}
  \frac{1}{n}  \lambda^2_i   \int_0^t 
  \dd \time  \; e^{   \frac{2}{n}   \lambda^2_i  ( \time - t) }    
  \right] = \\
  = \E_{\Lsvd} \left[ \sum^r_{i = 1}
  \frac{1}{n}  \lambda^2_i  \cdot \frac{n}{2 \lambda^2_i}  
   \left( 1 - e^{ - \frac{2}{n}   \lambda^2_i  t  }  \right)  
  \right] 
  = \frac{1}{2} \E_{\Lsvd} \left[ \sum^r_{i = 1}  
   \left( 1 - e^{ - \frac{2}{n}   \lambda^2_i  t  }  \right)  
  \right] = \frac{1}{2}\E_{\Lsvd}  \left[  \Tr{\ID_r  - e^{ - \frac{2}{n}   \Lsvd^2_r  t  }  }  
 \right] \;.
\end{multline}

With averaging over feature set $\sub$ sampling as in Eq.~\eqref{eq:sub_avg} and substituting as above, we get the expression:
\begin{equation}
\label{eq:etest_sde_repeat}
\begin{aligned}
\E_{\sub; \X, \y  }  \left[  \Tr{ \C (t) }  \right] & =   
    \frac{1}{d} \| \bb  - \bbhz \|^2  \cdot
        \int_0^t 
    \dd \time   \;      \E_{\Lsvd} \left[  \Tr{
        \frac{1}{n}  \Lsvd^2_r  \cdot   e^{-\frac{2}{n } \Lsvd^2_r \time } } 
         \Tr{   \frac{1}{n}  \Lsvd^2_r  \cdot    e^{ - \frac{2}{n}  \Lsvd^2_r (t - \time) }    
    }
    \right]   + 
    \\& + \left(
    \left(1 - \frac{p}{d} \right) \| \bb \|^2 + \mu^2 
    \right)  \cdot \Bigg( \frac{1}{n } 
        \int_0^t 
        \dd \time   \;      \E_{\Lsvd} \left[ 
                \Tr{ e^{-\frac{2}{n } \Lsvd^2_r \time }   } 
    \Tr{ \frac{1}{n }  \Lsvd^2_r  \cdot  e^{ - \frac{2}{n}  \Lsvd^2_r (t - \time) }    
    } \right] 
    + \\
    & +  \frac{1}{2} \max \left(0, 1 - \frac{p}{n} \right)  \E_{\Lsvd}  \left[  \Tr{\ID_r  - e^{ - \frac{2}{n}   \Lsvd^2_r  t  }  } 
    \right]   \Bigg)  \;,
\end{aligned}
\end{equation}
which matches Eq.~\eqref{SGF-finite-size}.
\section{Asymptotic generalization error}
In Appendix~\ref{app:belkin}, we derived precise formulas for generalization errors of GF and SGF solutions. However, to analyze these results, we need to find expectations for several functions of $\Lsvd$.

%%%%%%%%%%%%%%%%%%%
\subsection{Reformulation with empirical distribution of eigenvalues}
%%%%%%%%%%%%%%%%%%%
We consider $p \times p$ matrix $\frac{1}{n} \X^\top_\sub \X_\sub$. As elements of $\X_\sub$ are i.i.d.\:standard normal, the former matrix follows the Wishart distribution $W_p \left( n, 
\: \frac{1}{n} \ID_p \right) $ with $n$ degrees of freedom.
Let $\Wishpn$ be empirical distribution of eigenvalues of  $\frac{1}{n} \X^\top_\sub \X_\sub$.  As $\X_\sub = \Usvd \Lsvd \Vsvd^\top$, the spectral decomposition is
\begin{equation}
    \frac{1}{n} \X^\top_\sub \X_\sub = \Vsvd \left( \frac{1}{n} \Lsvd^\top \Lsvd \right) \Vsvd^\top \;,
\end{equation} 
i.e., eigenvalues are values on the diagonal of $\Lsvd^\top \Lsvd $. Following that, the measure $\Wishpn$ can be defined as:
\begin{equation}
    \Wishpn \left( A \right) = \frac{1}{p} \# \left\{ \frac{1}{n} \lambda^2_i \in A \mid i \in \overline{1, p} \right\} \; , \;  A \subset \R   \;.
\end{equation} 
If $p \leq n$, with probability 1, all eigenvalues are positive; $\{ 0 \}$ is a set of measure 0, and $\Wishpn$ is an absolutely continuous distribution. Otherwise, it has measure $1 - \nicefrac{n}{p}$.  

Let $\varphi : \R \mapsto \R$ be a pointwise function. For arbitrary diagonal $m \times m$ matrix $\D = \diag(d_{11}, \dots, d_{mm})$, we define $\varphi(\D) := \diag\left(\varphi(d_{11}), \dots, \varphi(d_{mm})\right) \in \R^{m \times m}$. Then,
\begin{multline}
    \E_{\sigma \sim \Wishpn} \left[ \varphi(\sigma) \right] = \frac{1}{p} \sum^p_{i = 1} \E_{\lambda_i } \left[  \varphi\left(\frac{1}{n} \lambda^2_i \right) \right]
    =
    \frac{1}{p} \E_{\Lsvd } \left[  \Tr{\varphi \left( \frac{1}{n} \Lsvd^\top \Lsvd \right) } \right] = \\
    =  \frac{1}{p} \: \E_{\Lsvd } \left[  \Tr{\varphi \left( \frac{1}{n} \Lsvd^2_r \right) } + (p - r) \; \varphi(0) \right] =
     \max \left(0, 1 - \frac{n}{p} \right) \varphi(0) + \frac{1}{p} \: \E_{\Lsvd } \left[  \Tr{\varphi \left( \frac{1}{n} \Lsvd^2_r \right) }  \right] \;.
\end{multline} 
Thus, when $p \leq n$, we can express $\frac{1}{p} \: \E_{\Lsvd } \left[  \Tr{\varphi \left( \frac{1}{n} \Lsvd^2_r \right) }  \right]$ in terms of expected value of absolutely continuous measure. When $p > n$, we can repeat our line of thought for  $\Wishnp$, which is an empirical distribution of eigenvalues of $n \times n$ matrix $\frac{1}{p}  \X_\sub \X^\top_\sub$. $\frac{1}{p}  \X_\sub \X^\top_\sub$ also has Wishart distribution -- $W_n\left(p, \: \frac{1}{p} \ID_n\right)$, -- and spectral decomposition
\begin{equation}
    \frac{1}{p}  \X_\sub \X^\top_\sub = \Usvd \left( \frac{1}{p} \Lsvd \Lsvd^\top  \right) \Usvd^\top \;,
\end{equation} 
and obtain:
\begin{equation}
    \E_{\sigma \sim \Wishnp} \left[ \varphi(\sigma) \right] =
     \max \left(0, 1 - \frac{p}{n} \right) \varphi(0) + \frac{1}{n} \: \E_{\Lsvd } \left[  \Tr{\varphi \left( \frac{1}{p} \Lsvd^2_r \right) }  \right] \;.
\end{equation} 

Now, let $\varphi': \R \mapsto \R$ be such function that $\forall x \in \R: \varphi'(x) = \varphi \left( \frac{n}{p } x \right)  $. Then, $\varphi' \left( \frac{1}{n} \Lsvd^2_r \right) = \varphi \left( \frac{1}{p} \Lsvd^2_r \right)$, and we can write:
\begin{equation}
\begin{aligned}
    \frac{1}{p} \: \E_{\Lsvd } \left[  \Tr{\varphi' \left( \frac{1}{n} \Lsvd^2_r \right) }  \right] & =  \frac{n}{p} \cdot \frac{1}{n} \: \E_{\Lsvd } \left[  \Tr{\varphi \left( \frac{1}{p} \Lsvd^2_r \right) }  \right] \\
    & = \frac{n}{p} \left( \E_{\sigma \sim \Wishnp} \left[ \varphi(\sigma) \right] -
    \max \left(0, 1 - \frac{p}{n} \right) \varphi(0)   \right)  \\ 
    & = \frac{n}{p} \: \E_{\sigma \sim \Wishnp} \left[ \varphi'\left( \frac{p}{n} \sigma \right) \right] -
    \max \left(0, \frac{n}{p} - 1 \right) \varphi(0) \;.
\end{aligned}
\end{equation} 

To sum up, for $\varphi: \R \mapsto \R$, we can use the following representations:
\begin{equation}
    \label{eq:lsvd_wishart_distrib}
    \frac{1}{p} \: \E_{\Lsvd } \left[  \Tr{\varphi \left( \frac{1}{n} \Lsvd^2_r \right) }  \right]
    =
    \begin{cases}
        \E_{\sigma \sim \Wishpn} \left[ \varphi(\sigma) \right]  & \text{for }\; p \leq n \;; \\
        \frac{n}{p} \: \E_{\sigma \sim \Wishnp} \left[ \varphi\left( \frac{p}{n} \sigma \right) \right] & \text{for }\; p > n \;.
    \end{cases}
\end{equation} 

%%%%%%%%%%%%%%%%%%%
\subsection{Asymptotic approximation with Marchenko-Pastur distribution}
%%%%%%%%%%%%%%%%%%%
Unfortunately, to our knowledge, there is no explicit expression for probability density of $\Wishnp$ which would be convenient to use for analysis of generalization error of solutions for finite sizes. However, for high enough values of $n, d, p$ we can approximate the results asymptotically. 

More precisely, as mentioned in Section \ref{sec:test-GF-SGF},  we consider $n, d, p \to \infty$ such that $\nicefrac{p}{n} \to \alpha$,  $\nicefrac{d}{n} \to \psi$, where $\alpha$ and $\psi$ are fixed positive values. As $p$ must be less or equal than $d$, $\alpha \leq \psi$. Under these conditions, as $ \X_\sub $ has  i.i.d.\:standard normal elements, empirical distribution $\Wishpn$ of eigenvalues of $\frac{1}{n} \X^\top_\sub \X_\sub$ weakly converges to well-known Marchenko-Pastur law $\MPalphaorig$. Its measure can be written as
\begin{equation}
    \label{eq:MPalphaorig_def}
    \MPalphaorig(A) = \begin{cases}
    \left(1 - \frac{1}{\alpha} \right) \Ind \left(0 \in A \right) + \MPalpha(A) \;, & \alpha > 1  \\
    \MPalpha(A) \;, & 0 \leq \alpha \leq 1
\end{cases}; \; \; A \subset \R,
\end{equation}
where $\MPalpha$ is the following measure:
\begin{equation}
    \MPalpha(\dd  \sigma) =   
    \frac{\sqrt{(\aplus - \sigma)(\sigma - \aminus)}}{2\pi\alpha\sigma} \Ind(\sigma\in [\alpha_-, \alpha_+]) \: \dd \sigma\;, \; \; \text{where} \; \alpha_{\pm}=(1\pm \sqrt\alpha)^2 \;.
\end{equation}
We remark that $\MPalpha$ is not a probability measure when $\alpha > 1$ as $\MPalpha(\R) \neq 1$. Likewise to $\Wishpn$, $\Wishnp$ converges to $\MPalphaoriginv$.

Then, we derive from Eq.~\eqref{eq:lsvd_wishart_distrib} that if $\varphi: \R \mapsto \R$ is a pointwise function,
\begin{equation}
    \label{eq:lsvd_MP_distrib_orig}
    \frac{1}{p} \: \E_{\Lsvd } \left[  \Tr{\varphi \left( \frac{1}{n} \Lsvd^2_r \right) }  \right] \MPconverge 
    \begin{cases}
        \E_{\sigma \sim  \MPalphaorig} \left[ \varphi(\sigma) \right]\;,& \alpha \leq 1; \\
        \frac{1}{\alpha} \: \E_{\sigma \sim \MPalphaoriginv} \left[ \varphi\left( \alpha \sigma \right) \right]\;, & \alpha > 1.
    \end{cases}
\end{equation}

Now, when $\alpha \leq 1$,
\begin{equation}
    \label{eq:MP_distrib_orig_alpha_leq_1}
    \E_{\sigma \sim  \MPalphaorig} \left[ \varphi(\sigma) \right] = \int  \varphi(\sigma) \: \MPalpha(\dd \sigma),
\end{equation}
and when $\alpha > 1$,
\begin{equation}
    \frac{1}{\alpha} \: \E_{\sigma \sim \MPalphaoriginv} \left[ \varphi\left( \alpha \sigma \right) \right] = \frac{1}{\alpha}  \int  \varphi(\alpha \sigma) \: \MPalphainv(\dd  \sigma) =
    \frac{1}{\alpha} \int^{\left(\nicefrac{1}{\alpha} \right)_+}_{\left(\nicefrac{1}{\alpha} \right)_-}    \varphi(\alpha \sigma)   \frac{\sqrt{\left( \left(\nicefrac{1}{\alpha} \right)_+ - \sigma \right) \left( \sigma - \left(\nicefrac{1}{\alpha} \right)_- \right) }}{2\pi\sigma / \alpha} \dd  \sigma \;.
\end{equation}
We can transform the latter expression; as
\begin{equation}
    \left(\frac{1}{\alpha} \right)_{\pm} = \left(1 \pm \frac{1}{\sqrt{\alpha}} \right)^2 = \frac{1}{\alpha}  (1 \pm \sqrt{\alpha})^2  = \frac{1}{\alpha} \cdot \alpha_{\pm} \;,
\end{equation} 
we continue
\begin{multline}
    \label{eq:MP_distrib_orig_alpha_g_1}
    \frac{1}{\alpha} \: \E_{\sigma \sim \MPalphaoriginv} \left[ \varphi\left( \alpha \sigma \right) \right] =  
    \int^{ \nicefrac{\alpha_{+}}{\alpha}  }_{ \nicefrac{\alpha_{-}}{\alpha} }    \varphi(\alpha \sigma)   \frac{\sqrt{\left( \alpha_+ - \alpha \sigma \right) \left( \alpha \sigma - \alpha_- \right) }}{2\pi \alpha (\alpha \sigma)  }\: \alpha \,\dd  \sigma = \left| \begin{array}{l}
        \zeta := \alpha \sigma; \\
        \dd \zeta= \alpha \dd  \sigma
        \end{array} \right| = \\
    = \int^{  \alpha_{+}  }_{  \alpha_{-} }    \varphi(  \zeta)   \frac{\sqrt{\left( \alpha_+ - \zeta \right) \left( \zeta - \alpha_- \right) }}{2 \pi 
    \alpha \zeta }\:  \dd \zeta = \int  \varphi(\zeta) \: \MPalpha(\dd \zeta) \;. 
\end{multline}

Combining results from Eqs.~\eqref{eq:MP_distrib_orig_alpha_leq_1} and~\eqref{eq:MP_distrib_orig_alpha_g_1} with Eq.~\eqref{eq:lsvd_MP_distrib_orig}, we can state for any $\alpha \in (0, \psi]$ that
\begin{equation}
    \label{eq:lsvd_MP_distrib_final}
    \frac{1}{p} \: \E_{\Lsvd } \left[  \Tr{\varphi \left( \frac{1}{n} \Lsvd^2_r \right) }  \right] \MPconverge 
    \int  \varphi(\sigma) \: \MPalpha(\dd \sigma) \;.
\end{equation}

%%%%%%%%%%%%%%%%%%%
\subsubsection{Gradient flow test error}
%%%%%%%%%%%%%%%%%%%
With the findings above, we can give an approximate estimate for GF and SGF generalization error. Starting with the former, i.e., Eq.~\eqref{eq:etest_ode_repeat}, we can provide asymptotic limits:
\begin{equation}
    \frac{1}{d}
    \Tr{ \E_{\Lsvd} \left[e^{-\frac{2}{n} \Lsvd^2_r   t }   \right] }  =
     \Big| \varphi(\sigma) := e^{-2 \sigma t} \Big| = \frac{p / n}{d / n} \cdot \frac{1}{p} \E_{\Lsvd } \left[  \Tr{\varphi \left( \frac{\Lsvd^2_r}{n}  \right) }  \right] \;\MPconvergeshort\; \frac{\alpha}{\psi} \int  e^{-2 \sigma t} \: \MPalpha(\dd \sigma)
\end{equation}
and
\begin{multline}
    \Tr{ \E_{  \Lsvd } \left[   \Lsvd^{-2}_r \left(  \ID_r  -  e^{-\frac{1}{n } \Lsvd^2_r t }    \right)^2    \right] } = 
    \Big| \varphi(\sigma) := \frac{\left(1 - e^{-\sigma t} \right)^2}{\sigma}  \Big| 
    = \\ =
    \frac{p}{n} \cdot \frac{1}{p} \E_{\Lsvd } \left[  \Tr{\varphi \left( \frac{\Lsvd^2_r}{n}  \right) }  \right] \MPconvergeshort\; \alpha \int  \frac{\left(1 - e^{-\sigma t} \right)^2}{\sigma} \: \MPalpha(\dd \sigma) \;,
\end{multline}
which results in
\begin{equation}
\label{eq:gf_test_asymp_full}
\begin{aligned}
 \limasymp \Etestode  &=   
    \frac{1}{2} \Bigg(  \frac{1}{\psi} \| \bb  - \bbhz  \|^2
         \left( 
        \max\left(0, \; \alpha - 1 \right) + 
         \alpha \int  e^{-2 \sigma t} \: \MPalpha(\dd \sigma) \right)
    + \\ &+
        \left(
    \left(1 - \frac{\alpha}{\psi} \right) \| \bb \|^2 + \mu^2 
        \right)
    \left[
        1 + 
        \alpha \int  \frac{\left(1 - e^{-\sigma t} \right)^2}{\sigma} \: \MPalpha(\dd \sigma) \right] \Bigg) \;;
\end{aligned}
\end{equation}
this coincides with Eq.~\eqref{assympt-1} when $\Vert\bb\Vert=1$ and $\Vert\bb - \bbhz\Vert^2 \to 2$.

To obtain an infinite-time limit for the test error, we can see that for $\alpha \neq 1$,
\begin{equation}
\label{eq:tinf_exp_int_argument}
\begin{aligned}
    \int  e^{-2 \sigma t} \: \MPalpha(\dd \sigma) & \leq e^{-2 \aminus t} \left( \int \MPalpha(\dd \sigma) \right) \xrightarrow{t \to  +\infty} 0 \;;
    \\
    \int  \frac{e^{-k \sigma t}}{\sigma} \: \MPalpha(\dd \sigma) & \leq e^{-k \aminus t} \left( \int \frac{\MPalpha(\dd \sigma)}{\sigma} \right) \xrightarrow{t \to  +\infty} 0 \;, \; k \in \mathbb{N} \;.
\end{aligned}
\end{equation}

Therefore, the first integral term of Eq.~\eqref{eq:gf_test_asymp_full} will go away, and in the integral second term, we can set aside a part independent of $t$:
\begin{equation}
    \alpha \int  \frac{\left(1 - e^{-\sigma t} \right)^2}{\sigma} \: \MPalpha(\dd \sigma) =   \alpha \int  \frac{1}{\sigma} \: \MPalpha(\dd \sigma)  +  \alpha \int  \frac{e^{-2\sigma t} - 2 e^{- \sigma t}}{\sigma} \: \MPalpha(\dd \sigma) \;,
\end{equation} 
which will be the only one remaining when $t \to  +\infty$ and can be simplified as 
\begin{equation}
    \alpha \int  \frac{1}{\sigma} \: \MPalpha(\dd \sigma) = \alpha \cdot  \int_{\aminus}^{\aplus} \frac{\sqrt{(\aplus - \sigma)(\sigma - \aminus)}}{2 \pi \alpha \sigma^2} \dd \sigma 
  = \alpha \cdot \begin{cases}
        \frac{1}{1 - \alpha} \;, & \alpha \leq 1\;, \\
        \frac{1}{\alpha - 1 } - \frac{1}{\alpha}\;, & \alpha > 1 \;.
    \end{cases} =  
     \begin{cases}
        \frac{1}{1 - \alpha} - 1 \;, & \alpha \leq 1 \;, \\
        \frac{1}{\alpha - 1 } \;, & \alpha > 1 \;.
    \end{cases} 
\end{equation} 
Now, as
\begin{equation}
   1 + \alpha \int  \frac{1}{\sigma} \: \MPalpha(\dd \sigma) = 
     \begin{cases}
        \frac{1}{1 - \alpha} \;,   & \alpha \leq 1 \;, \\
        \frac{\alpha}{\alpha - 1 } \;, & \alpha > 1
    \end{cases} =  \frac{1}{1  - \min\left(\alpha, \nicefrac{1}{\alpha} \right)} \;,
\end{equation} 
we retrieve a GF test error limit when $t \to +\infty$:
\begin{equation}
    \label{eq:gf_test_asymp_tinf}
    \begin{aligned}
   \lim_{t \to +\infty}  \limasymp \Etestode  &=   
        \frac{1}{2} \Bigg(  \frac{1}{\psi} \| \bb  - \bbhz  \|^2
            \max\left(0, \; \alpha - 1 \right)  + 
            \left(
        \left(1 - \frac{\alpha}{\psi} \right) \| \bb \|^2 + \mu^2 
            \right) 
            \frac{1}{1  - \min\left(\alpha, \nicefrac{1}{\alpha} \right)} \Bigg) \;.
    \end{aligned}
    \end{equation}

%%%%%%%%%%%%%%%%%%%
\subsubsection{Stochastic gradient flow test error}\label{app:sgf_testrk}
%%%%%%%%%%%%%%%%%%%
Now, we are going to SGF, i.e., the stochastic component additive to GF error from Eq.~\eqref{eq:etest_sde_repeat}. There, however, we have to deal with expressions of form $\E_{\Lsvd} \left[  \Tr{
    \varphi_1 \left(   \frac{1}{n}  \Lsvd^2_r \right) } 
    \Tr{\varphi_2 \left(   \frac{1}{n}  \Lsvd^2_r \right) } 
\right] $ for $\varphi_1, \varphi_2 : \R \mapsto \R$. However, for large $n, p$, we can assume that the (normalized) traces concentrate so that:
\begin{equation}
    \E_{\Lsvd} \left[  \Tr{
        \varphi_1 \left(   \frac{1}{n}  \Lsvd^2_r \right) } 
        \Tr{\varphi_2 \left(   \frac{1}{n}  \Lsvd^2_r \right) } 
    \right] \approx     \E_{\Lsvd} \left[  \Tr{
        \varphi_1 \left(   \frac{1}{n}  \Lsvd^2_r \right) } 
    \right]    \E_{\Lsvd} \left[  \Tr{
        \varphi_2 \left(   \frac{1}{n}  \Lsvd^2_r \right)  }
    \right] \;.
\end{equation}
Then
\begin{equation}
\begin{aligned}
    \frac{1}{p^2} \E_{\Lsvd} & \left[  \Tr{
        \frac{1}{n}  \Lsvd^2_r  \cdot   e^{-\frac{2}{n } \Lsvd^2_r \time } } 
         \Tr{   \frac{1}{n}  \Lsvd^2_r  \cdot    e^{ - \frac{2}{n}  \Lsvd^2_r (t - \time) }    
    }
    \right] \\
    & =   \left| 
        \varphi_1 (\sigma) := \sigma e^{-2 \sigma \time},
        \varphi_2 (\sigma) := \sigma e^{-2 \sigma (t - \time) }  \right|  
     \\ & \approx
      \frac{1}{p}  \E_{\Lsvd} \left[  \Tr{
        \varphi_1 \left(   \frac{1}{n}  \Lsvd^2_r \right) } 
    \right]  \cdot \frac{1}{p}  \E_{\Lsvd} \left[  \Tr{
        \varphi_2 \left(   \frac{1}{n}  \Lsvd^2_r \right)  }
    \right] \\
    & \MPconvergeshort \int  \sigma_1 e^{-2 \sigma_1 \time}  \: \MPalpha(\dd \sigma_1) \cdot \int  \sigma_2 e^{-2 \sigma_2 (t - \time)}  \: \MPalpha(\dd \sigma_2) \;.
\end{aligned}
\end{equation}
Similarly, by taking $\varphi_1 (\sigma) :=  e^{-2 \sigma \time},
\varphi_2 (\sigma) := \sigma e^{-2 \sigma (t - \time) } $, we get:
\begin{equation}
\begin{aligned}
   \frac{1}{p^2} \E_{\Lsvd} & \left[ 
        \Tr{    e^{-\frac{2}{n } \Lsvd^2_r \time }   } 
\Tr{ \frac{1}{n }  \Lsvd^2_r  \cdot  e^{ - \frac{2}{n}  \Lsvd^2_r (t - \time) }    
} \right] 
\MPconvergeshort  \int e^{-2 \sigma_1 \time}  \MPalpha(\dd \sigma_1) \cdot \int  \sigma_2 e^{-2 \sigma_2 (t - \time)}  \MPalpha(\dd \sigma_2) \;.
\end{aligned}
\end{equation}

We can now integrate over $\time$ by changing the order of integration:
\begin{equation}
\begin{aligned}
    \frac{1}{p^2}  & \int^t_0 \dd \time \;  \E_{\Lsvd}  \left[ 
        \Tr{    e^{-\frac{2}{n } \Lsvd^2_r \time }   } 
\Tr{ \frac{1}{n }  \Lsvd^2_r  \cdot  e^{ - \frac{2}{n}  \Lsvd^2_r (t - \time) }    
} \right] \\
     & \MPconvergeshort 
    \int^t_0 \dd \time \int  \sigma_1 e^{-2 \sigma_1 \time} \MPalpha(\dd \sigma_1)  \int  \sigma_2 e^{-2 \sigma_2 (t - \time)}  \MPalpha(\dd \sigma_2)  \\
    &= \int  \sigma_1   \: \MPalpha(\dd \sigma_1)  \int  e^{-2 \sigma_2 t} \sigma_2  \MPalpha(\dd \sigma_2) \int^t_0 \dd \time   \: e^{-2 ( \sigma_1 - \sigma_2) \time}  \\
&=     \int  \sigma_1   \: \MPalpha(\dd \sigma_1)  \int  e^{-2 \sigma_2 t} \sigma_2  \MPalpha(\dd \sigma_2) \;
\frac{e^{-2 ( \sigma_1 - \sigma_2) t}  - 1}{-2(\sigma_1-\sigma_2)} 
\\
&=  \int   \int   \MPalpha(\dd \sigma_1) \: \MPalpha(\dd \sigma_2) \: \sigma_1 \sigma_2  \frac{e^{-2 \sigma_1 t}  -  e^{-2 \sigma_2 t} }{ 2(\sigma_2-\sigma_1)} 
\;, 
\end{aligned}
\end{equation}

In the same way:
\begin{equation}
\begin{aligned}
    \frac{1}{p^2}  & \int^t_0 \dd \time \; \E_{\Lsvd}  \left[  \Tr{
        \frac{1}{n}  \Lsvd^2_r  \cdot   e^{-\frac{2}{n } \Lsvd^2_r \time } } 
            \Tr{   \frac{1}{n}  \Lsvd^2_r  \cdot    e^{ - \frac{2}{n}  \Lsvd^2_r (t - \time) }    
    }
    \right] \\
        & \MPconvergeshort 
    \int^t_0 \dd \time 
    \int e^{-2 \sigma_1 \time}  \MPalpha(\dd \sigma_1) \cdot \int  \sigma_2 e^{-2 \sigma_2 (t - \time)}  \MPalpha(\dd \sigma_2)
    \\
    &= \int    \: \MPalpha(\dd \sigma_1)  \int  e^{-2 \sigma_2 t} \sigma_2  \MPalpha(\dd \sigma_2) \int^t_0 \dd \time   \: e^{-2 ( \sigma_1 - \sigma_2) \time}  \\
&=     \int  \: \MPalpha(\dd \sigma_1)  \int  e^{-2 \sigma_2 t} \sigma_2  \MPalpha(\dd \sigma_2) \; 
\frac{e^{-2 ( \sigma_1 - \sigma_2) t}  - 1}{-2(\sigma_1-\sigma_2)}   \\
&=  \int   \int   \MPalpha(\dd \sigma_1) \: \MPalpha(\dd \sigma_2) \: \sigma_2  \frac{e^{-2 \sigma_1 t}  -  e^{-2 \sigma_2 t} }{ 2(\sigma_2-\sigma_1)} \;.
\end{aligned}
\end{equation}

Finally, $\E_{\Lsvd}  \left[  \Tr{\ID_r  - e^{ - \frac{2}{n}   \Lsvd^2_r  t  }  } \right]$ can be treated as seen before in Eq.~\eqref{eq:lsvd_MP_distrib_final}:
\begin{equation}
    \frac{1}{p}
    \E_{\Lsvd}  \left[  \Tr{\ID_r  - e^{ - \frac{2}{n}   \Lsvd^2_r  t  }  } \right]   =
     \Big| \varphi(\sigma) := 1 - e^{-2 \sigma t} \Big| =   \frac{1}{p} \E_{\Lsvd } \left[  \Tr{\varphi \left( \frac{\Lsvd^2_r}{n}  \right) }  \right] \;\MPconvergeshort\; \int  \left(1 -  e^{-2 \sigma t} \right) \: \MPalpha(\dd \sigma) \;.
\end{equation}

Then, if we denote
\begin{align} 
    F_1(\alpha, t)&\equiv \int\int\rho_\alpha(\dd\sigma_1)\rho_\alpha(\dd \sigma_2) \sigma_1\sigma_2 K(t,\sigma_1, \sigma_2) \;,\\ 
    F_2(\alpha, t)&\equiv    \int\int\rho_\alpha(\dd\sigma_1)\rho_\alpha(\dd \sigma_2)  \sigma_2 K(t,\sigma_1, \sigma_2) \nonumber \\
    &=  \frac{1}{2} \int\int\rho_\alpha(\dd\sigma_1)\rho_\alpha(\dd \sigma_2) (\sigma_1+\sigma_2) K(t,\sigma_1, \sigma_2) \;,
    \\
    K(t, \sigma_1,\sigma_2) &\equiv \frac{e^{-2\sigma_1 t} - e^{-2\sigma_2 t}}{2(\sigma_2 - \sigma_1)}
    \;,
\end{align}
we obtain an asymptotic result for the stochastic component of SGF error:  
\begin{equation}
    \begin{aligned}
   \limasymp   & \frac{1}{n} \E_{\sub; \X, \y  }  \left[  \Tr{ \C (t) }  \right] 
    =   \limasymp \Bigg\{
          \frac{p^2}{d n }   \cdot \| \bb  - \bbhz \|^2  \cdot
        F_1(\alpha, t)   + \\
       & + \left(
        \left(1 - \frac{p}{d} \right) \| \bb \|^2 + \mu^2 
        \right)    \cdot \left( 
            \frac{p^2}{n^2 }   F_2(\alpha, t)
        +    \frac{1}{2} \max \left(0, 1 - \frac{p}{n} \right) \cdot \frac{p}{n}   
        \int  \left(1 -  e^{-2 \sigma t} \right) \: \MPalpha(\dd \sigma)
        \right) \Bigg\} \\
    &= \alpha \Bigg[ 
        \frac{\alpha}{\psi }   \cdot \| \bb  - \bbhz \|^2  \cdot
      F_1(\alpha, t)   +   \left(
      \left(1 - \frac{\alpha}{\psi} \right) \| \bb \|^2 + \mu^2 
      \right)    \cdot \\ &\qquad \cdot
         \left( 
          \alpha   \cdot F_2(\alpha, t)
      +     \max \left(0, 1 - \alpha \right) \cdot    
      \int  \frac{\left(1 -  e^{-2 \sigma t} \right)}{2} \: \MPalpha(\dd \sigma)
      \right)  \Bigg] \;,
    \end{aligned}
    \end{equation}
or, with scaling
\begin{equation}
    \frac{\gamma}{2} =  \frac{\gamma'}{2 d} = \frac{\gamma'}{2 \psi n} \;,
\end{equation} 
the expression is
\begin{equation}
    \begin{aligned}
   \limasymp     \frac{\gamma}{2}  \E_{\sub; \X, \y  }  \left[  \Tr{ \C (t) }  \right] 
   & =  \frac{\gamma'}{2} \cdot \frac{\alpha}{\psi } \Bigg[ 
        \frac{\alpha}{\psi }   \cdot \| \bb  - \bbhz \|^2  \cdot
      F_1(\alpha, t)   +   \left(
      \left(1 - \frac{\alpha}{\psi} \right) \| \bb \|^2 + \mu^2 
      \right)    \cdot \\ &  \cdot
         \left( 
          \alpha   \cdot F_2(\alpha, t)
      +     \max \left(0, 1 - \alpha \right) \cdot    
      \int  \frac{\left(1 -  e^{-2 \sigma t} \right)}{2} \: \MPalpha(\dd \sigma)
      \right)  \Bigg] \;,
    \end{aligned}
    \end{equation}
which matches Eq.~\eqref{assympt-2} when   $\Vert\bb\Vert=1$ and $\Vert\bb - \bbhz\Vert^2 \to 2$.

Now, we obtain an infinite time limit. With argument similar to Eq.~\eqref{eq:tinf_exp_int_argument}, we can write for $\alpha \neq 1$:
\begin{equation}
    \lim_{t \to +\infty} F_1(\alpha, t) = 0, \lim_{t \to +\infty} F_2(\alpha, t) = 0 \;;
\end{equation}
besides, from Eq.~\eqref{eq:MPalphaorig_def} we can see:
\begin{equation}
    \int \: \MPalpha(\dd \sigma) = \min\left(1, \frac{1}{\alpha}\right) \;,
\end{equation}
and as
\begin{equation}
    \max \left(0, 1 - \alpha \right) \cdot \min\left(1, \frac{1}{\alpha}\right) = \max \left(0, 1 - \alpha \right), 
\end{equation}
the infinite time limit of difference between test errors of SGF and GF is:
\begin{equation}
    \begin{aligned}
 \lim_{t \to +\infty}  \limasymp    \frac{\gamma}{2} \E_{\sub; \X, \y  }  \left[  \Tr{ \C (t) }  \right] 
   & =  \frac{\gamma'}{4} \cdot \frac{\alpha}{\psi } \left(
      \left(1 - \frac{\alpha}{\psi} \right) \| \bb \|^2 + \mu^2 
      \right)    \cdot \max \left(0, 1 - \alpha \right) \;.  
    \end{aligned}
\end{equation}

\section{Additional numerics for the weak features model}
\label{app:add_numerics}

\begin{figure}[ht!]
  \centering    
  \subfigure[]{\label{fig:etestode_a}\includegraphics[width=0.45\textwidth]{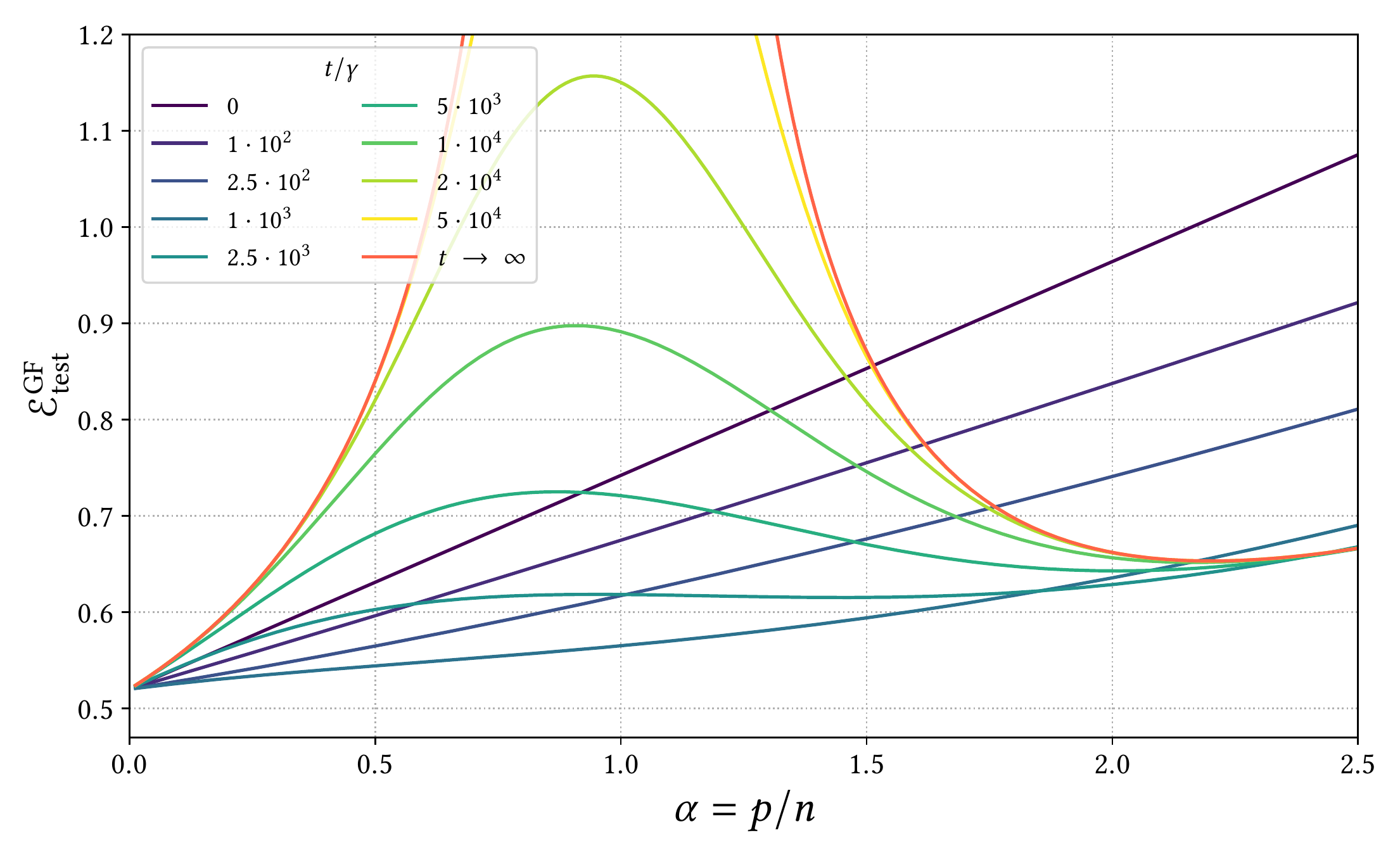}}
  \subfigure[]{\label{fig:etestode_b}\includegraphics[width=0.45\textwidth]{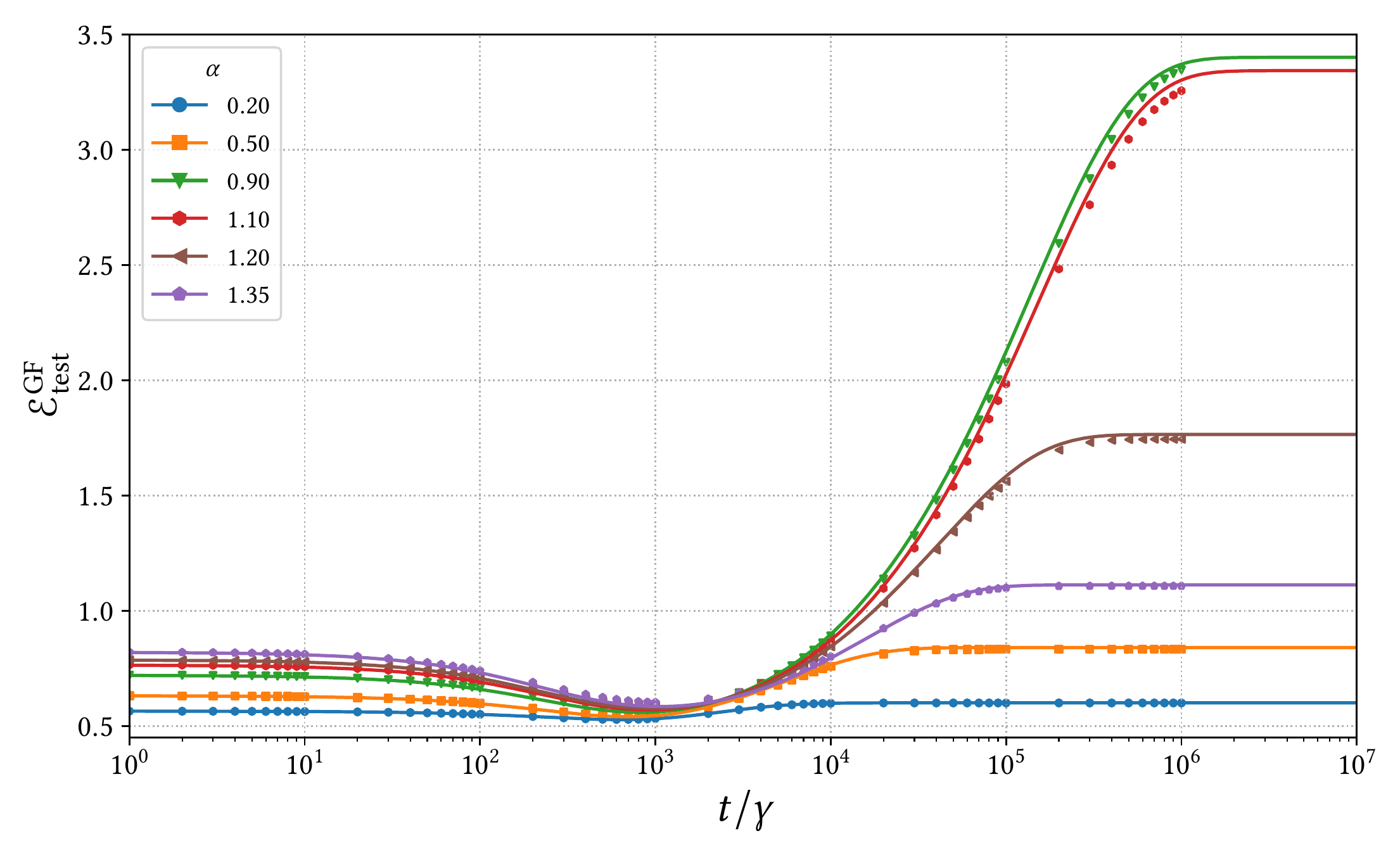}}
  \caption{Plots depicting $\Etestode$ for $\psi = 2.5$. Solid lines represent theoretical asymptotic estimation. Markers on \ref{fig:etestode_b} represent the results of numerical simulations of GD (averaged over $1000$ different random subsets $\sub$). Simulation parameters: $d=1000$, $\lr= 10^{-3}$; vectors $\bm{\beta}$, $\bm{\beta}_0$ are taken at random on the unit $1000$-dimensional sphere and here $\norm{\bm{\beta}-\bm{\beta}_0}^2 \approx 2.11$ }
  \label{fig:etestode}
  \end{figure}

%%%%%%%%%%%%%%%%%%%
\paragraph{Gradient flow -- }
%%%%%%%%%%%%%%%%%%%
Plots for the test risk of the deterministic gradient flow estimator are found in Fig.~\ref{fig:etestode}.
Fig.~\ref{fig:etestode_a}  focuses on the risk dependence on $\alpha$. One can see a limiting double descent picture (with degenerate first descent) described in~\cite{belkin_2020}. Besides, our derivations, particularly Eq.~\eqref{assympt-1}, also allow to see the emergence of the double descent with time. This is illustrated by 
Fig.~\ref{fig:etestode_b} which depicts the time evolution of the test risk for the GF estimator. We observe that under our choice of model parameters early stopping is beneficial for most values of $\alpha$. The gradient descent simulations come very close to our prediction. The discrepancy between the theoretical and empirical predictions is slightly more pronounced for values of $\alpha$ close to $1$, which we hypothesize to be caused by a lower rate of convergence of the empirical distribution of eigenvalues of $\Lsvd^2_r / n$  to Marchenko-Pastur distribution for such $\alpha$'s.

\begin{figure}[ht!]
  \centering    
  \subfigure[Simulations with $d=100$]{\label{fig:sgf_gf_a}\includegraphics[width=0.45\textwidth]{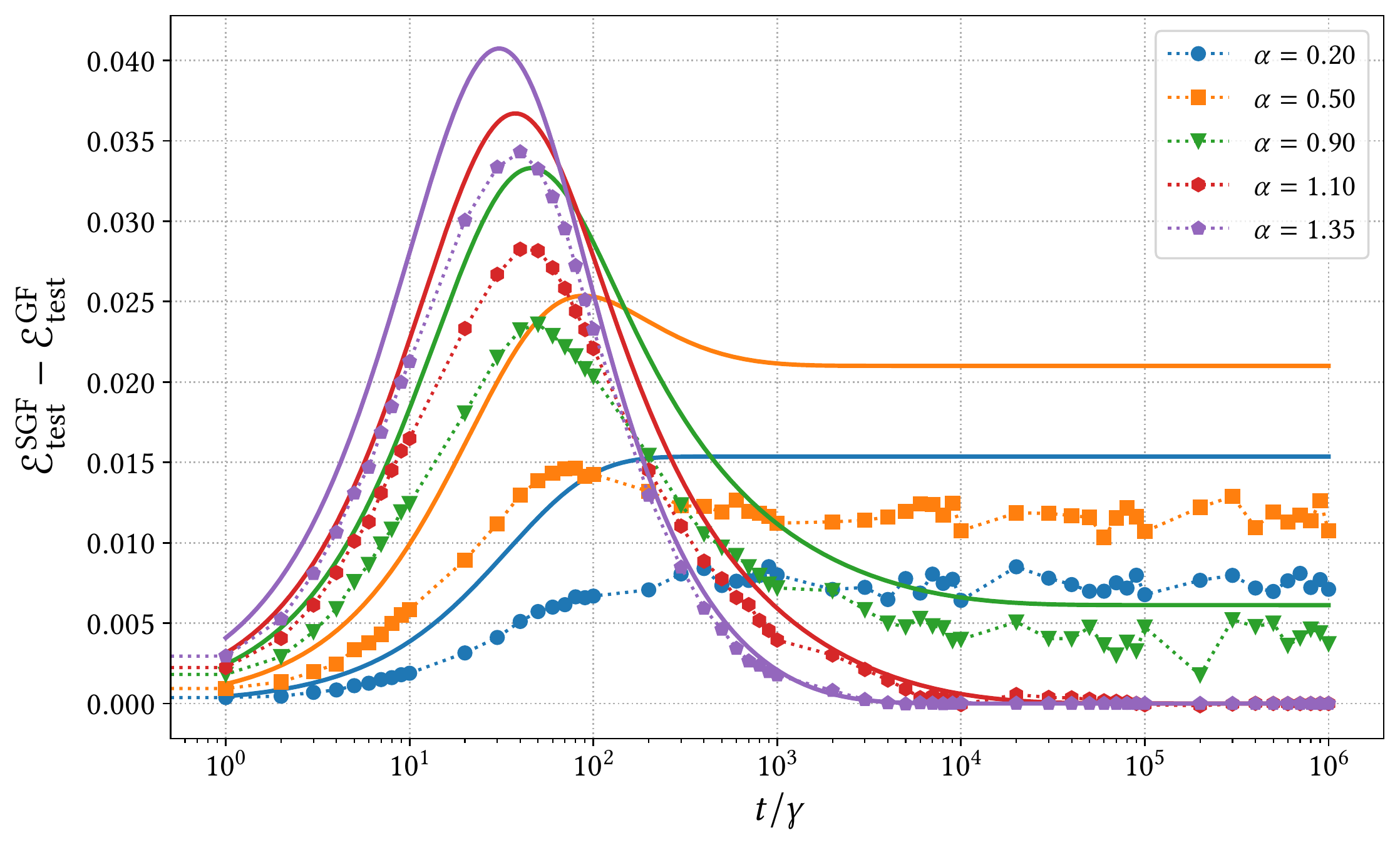}}
  \subfigure[Simulations with $d=1000$]{\label{fig:sgf_gf_b}\includegraphics[width=0.45\textwidth]{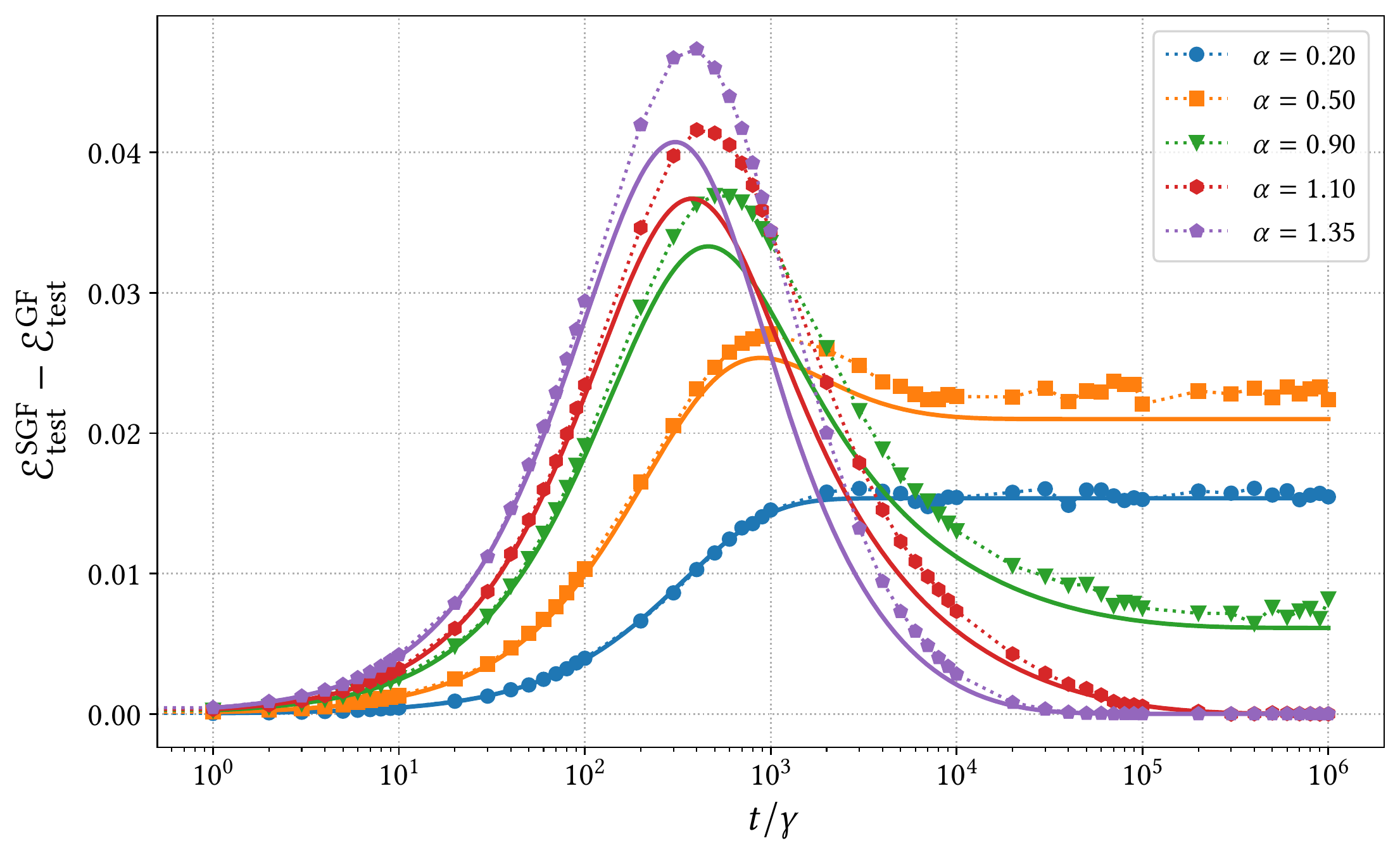}}
  \caption{Time-evolution of $ \Etest^{\text{SGF}} - \Etest^{\text{GF}}$ with $\psi = 2.5$ for different values of $\alpha$. Markers represent simulations while lines represent the theoretical predictions. The learning rate is scaled as $\gamma = \nicefrac{1}{d}$. Vectors $\bm{\beta}$, $\bm{\beta}_0$ are taken at random on the unit $d$-dimensional sphere. Here $\norm{\bm{\beta}-\bm{\beta}_0}^2 \approx 1.78$ for $d=100$ and $\norm{\bm{\beta}-\bm{\beta}_0}^2 \approx 2.11$ for $d=1000$.}
  \label{fig:sgf_gf}
  \end{figure}

%%%%%%%%%%%%%%%%%%%
\paragraph{Difference between SGF and GF --}
%%%%%%%%%%%%%%%%%%%
We illustrate the time-evolution of the difference $\Etest^{\text{SGF}}-\Etest^{\text{GF}}$ and its numerical simulation counterpart 
$\Etest^{\text{SGD}}-\Etest^{\text{GD}}$, 
as well as finite-size effects. These differences are presented in Fig.~\ref{fig:sgf_gf} as a function of $\nicefrac{t}{\gamma}$ for different values of $\alpha$, below and above the interpolation threshold. Note that $t/\gamma$ can be interpreted as the number of iterations in gradient descent. Fig.~\ref{fig:sgf_gf_a} depicts the comparison with simulations for $d=100$ and Fig.~\ref{fig:sgf_gf_b} with simulations for $d=1000$. 
  
  We note pronounced finite-size effects when comparing Figs.~\ref{fig:sgf_gf_a} (with $d=100$) and Figs.~\ref{fig:sgf_gf_b} (with $d=1000$), particularly below the interpolation threshold and high values of $\nicefrac{t}{\gamma}$. 

\begin{figure}[ht!]
    \vskip 0.2in
    \begin{center}
\centerline{\includegraphics[width=\columnwidth]{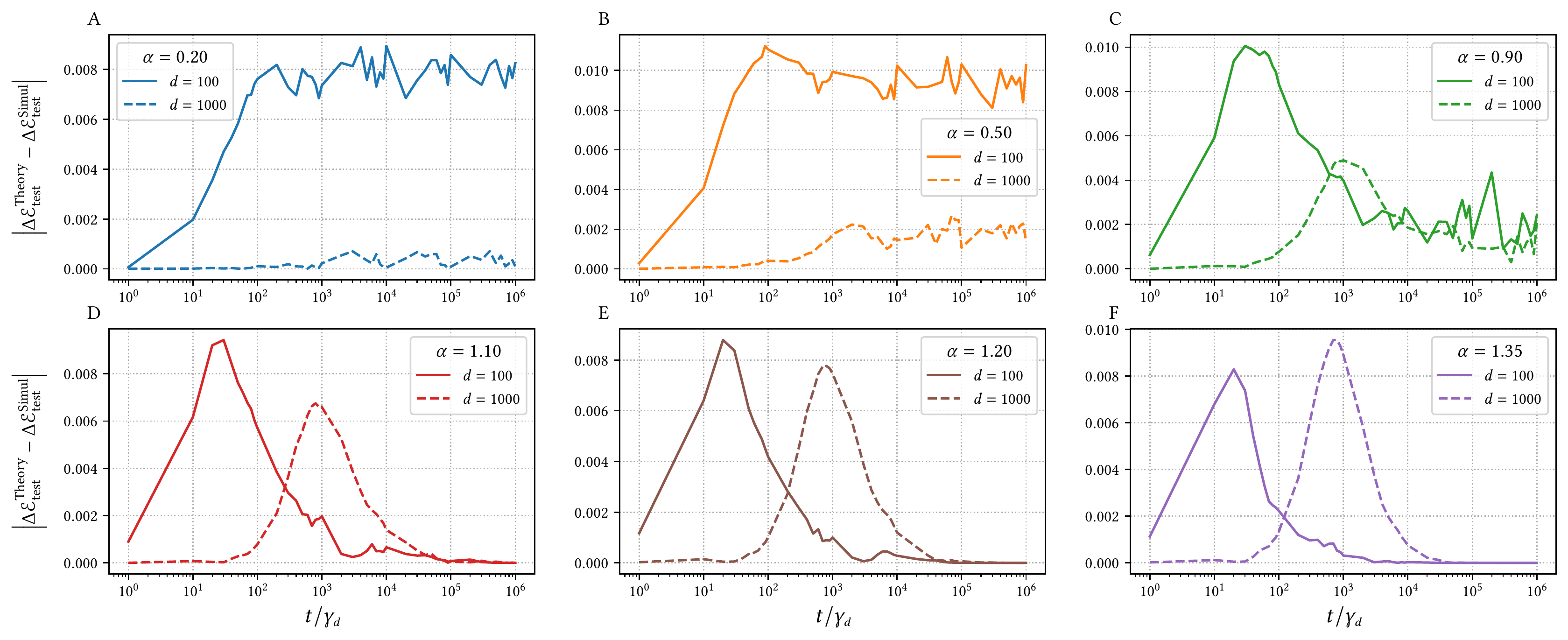}}
    \caption{Time-evolution of the difference between the theory and simulation ($d=100,1000$ and $\psi=2.5$) results for $\Delta\Etest^\text{Theory}-\Delta\Etest^\text{Simul}$ (defined in the text) for different values of $\alpha$, below and above the interpolation threshold. The time scaling is adjusted according to the fine size simulation: $\gamma_ d = \nicefrac{1}{d}$.}
    \label{fig:app_figDeltaE_th_simul}
    \end{center}
    \vskip -0.2in
\end{figure}

  Additionally, for finite times, we observe that the theoretical results are less accurate for times of order $t \sim 1$ (corresponding to roughly $1/\gamma=d$ GD iterations). That is already apparent from the `red curve' in Fig.~\ref{fig:cov_fintime} in the main text. To further investigate this point, we define $\Delta  \Etest^{\text{Theory}}  \equiv \Etest^{\text{SGF}} - \Etest^{\text{GF}}$ and $\Delta  \Etest^{\text{Simul}}  \equiv \Etest^{\text{SGD}} - \Etest^{\text{GD}}$, and compare the difference between $\Delta\Etest^\text{Theory}-\Delta\Etest^\text{Simul}$ (for the same setting as that of Fig.~\ref{fig:sgf_gf}). These differences are presented in Fig.~\ref{fig:app_figDeltaE_th_simul} for different values of $\alpha$, and with  $d=100$, $d=1000$. 
  
For values of $\alpha$ well below the interpolation threshold, the finite size effects are clearly visible in the whole time evolution and go away as $d$ increases. Above the threshold, the discrepancy is more pronounced around $t\sim 1$ and there is no clear evidence that this is a finite size effect. It remains to be seen if better continuous time modeling (for example second-order SDE modeling) would be tractable and make the discrepancy disappear.

\newpage
%%%%%%%%%%%%%%%%%%%%%%%%%%%%%%%
\addcontentsline{toc}{section}{References}
\bibliographystyle{IEEEtran}
\typeout{}\bibliography{paper}
%%%%%%%%%%%%%%%%%%%%%%%%%%%%%%%

\end{document}